\documentclass[10pt,twocolumn,letterpaper]{article}

\usepackage{times}
\usepackage{epsfig}
\usepackage{graphicx}
\usepackage{amsmath}
\usepackage{amssymb}
\usepackage{booktabs}
\usepackage{multirow}
\usepackage{algorithm}
\usepackage{algorithmic}
\usepackage{xcolor}
\usepackage{hyperref}

\title{LLMOrbit: A Circular Taxonomy of Large Language Models \\
—From Scaling Walls to Agentic AI Systems}

\author{
Badri N. Patro, Vijay S. Agneeswaran\\
Microsoft\\
\texttt{\{badripatro, vagneeswaran\}@microsoft.com}
}

\date{}

\begin{document}

\maketitle

\begin{abstract}
The field of artificial intelligence has undergone a revolution from foundational Transformer architectures to reasoning-capable systems approaching human-level performance on certain specific tasks. We present \textbf{LLMOrbit}, a comprehensive circular taxonomy navigating the complete landscape of large language models spanning 2019-2025. This survey examines over 50 major models across 15 organizations through eight interconnected orbital dimensions, documenting the architectural innovations, training methodologies, and efficiency patterns defining modern large language models (LLMs), generative AI, and agentic systems.

\textbf{The Scaling Wall:} We identify three critical crises threatening AI progress—(1) data scarcity (human generated text data is approximately 9-27T tokens, which could be depleted by 2026-2028), (2) exponential cost growth (\$3M to \$300M+ cost to train LLM models in 5 years), and (3) unsustainable energy consumption (22$\times$ increase)—establishing the "scaling wall" that fundamentally limits brute-force scaling.

\textbf{Breaking the Scaling Wall:} Our analysis reveals six different paradigms: (1) \textit{test-time compute}—o1 and DeepSeek-R1 achieve GPT-4 performance with 10$\times$ inference compute but smaller pre-training; (2) \textit{quantization}—4-8$\times$ compression with $<$ 1\% perplexity degradation~\cite{yao2024scaling,dettmers2024qlora}; (3) \textit{distributed edge computing}—10$\times$ cost reduction~\cite{yang2025edge}; (4) \textit{model merging}—synergistic capabilities~\cite{goddard2024mergekit,taylor2025domain}; (5) \textit{efficient training}—ORPO reduces memory 50\%~\cite{hong2024orpo}; (6) \textit{small specialized models}—Microsoft Phi-4 (14B) matches 10$\times$ larger models~\cite{phi42025reasoning}.

Three fundamental paradigm shifts emerge: (1) \textit{post-training gains}—recent work suggests post-training techniques (RLHF, GRPO, pure RL) contribute substantially to model capabilities~\cite{ouyang2022training,bai2022constitutional}, with DeepSeek-R1 achieving 79.8\% MATH through pure RL without supervised fine-tuning; (2) \textit{efficiency revolution}—MoE routing (18$\times$ efficiency), Multi-head Latent Attention (8$\times$ KV cache compression) enable GPT-4-level performance at $<$\$0.30/M tokens (100$\times$ cost reduction); (3) \textit{democratization}—open-source Llama 3 (88.6\% MMLU) surpasses GPT-4 (86.4\%), while global models from Alibaba (Qwen), Moonshot (Kimi), Microsoft (Phi), and Zhipu (GLM) demonstrate competitive capabilities.  We provide insight into key techniques (RLHF, PPO, DPO, GRPO, ORPO), trace evolution from passive generation to active tool-using agents (ReAct, RAG, multi-agent systems), and analyze post-training innovations (quantization, merging, edge deployment). Through benchmarking across 9 metrics, we identify reasoning emergence requires: (i) scale exceeding $10^{11}$ tokens, (ii) RL with verifiable feedback, and (iii) test-time search. LLMOrbit serves as both a technical reference for practitioners and a roadmap for researchers exploring reasoning, multimodal understanding, and autonomous agents in the post-scaling era.
\end{abstract}

\section{Introduction}
\label{sec:intro}

\begin{figure*}[t]
\centering
\includegraphics[width=0.98\textwidth]{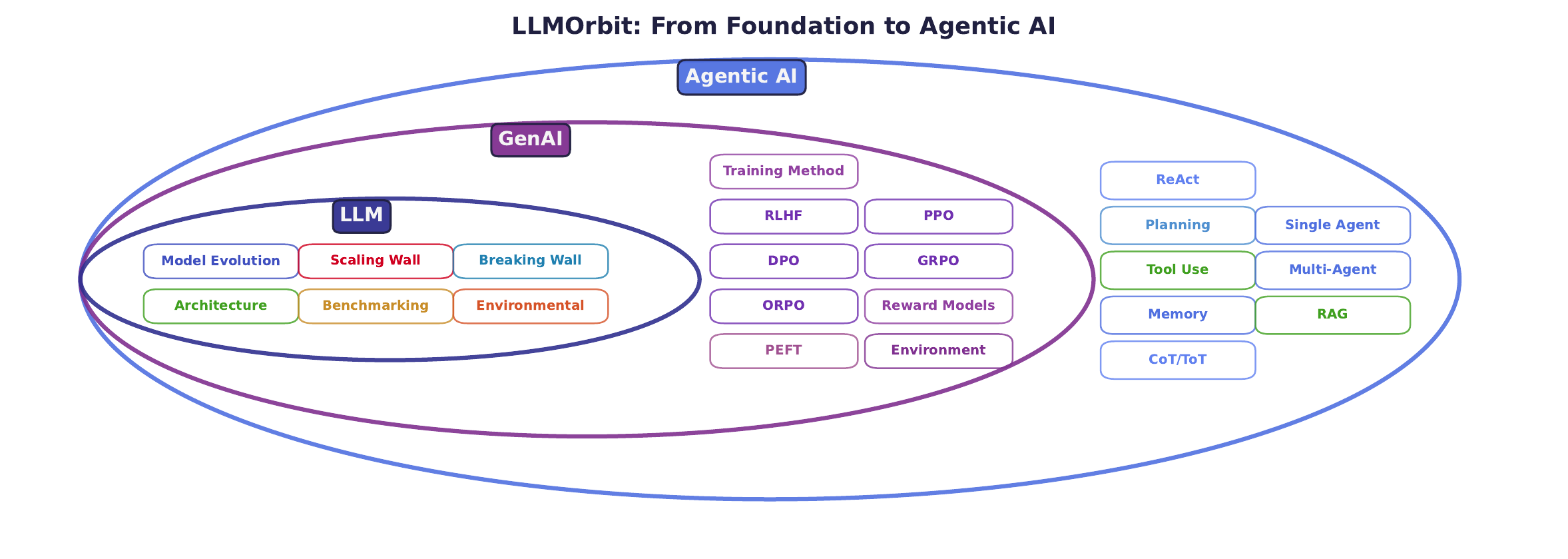}
\vspace{-0.3in}
\caption{\textbf{Evolution from LLM Foundation to Agentic AI.} Three nested paradigms converge to a unified framework: \textit{LLM Foundation} (blue) encompasses model evolution, scaling challenges, and architectural innovations; \textit{GenAI} (purple) adds training methodologies (RLHF, PPO, DPO, GRPO, ORPO) and environments; \textit{Agentic AI} (light blue) extends capabilities through reasoning (ReAct, CoT/ToT), tool use (RAG), and multi-agent systems. This nested architecture illustrates how foundation models serve as the base for generative capabilities, which in turn enable autonomous agentic systems.}
\label{fig:teaser}
\vspace{-0.2in}
\end{figure*}

\begin{figure*}
\centering
\includegraphics[width=0.95\textwidth]{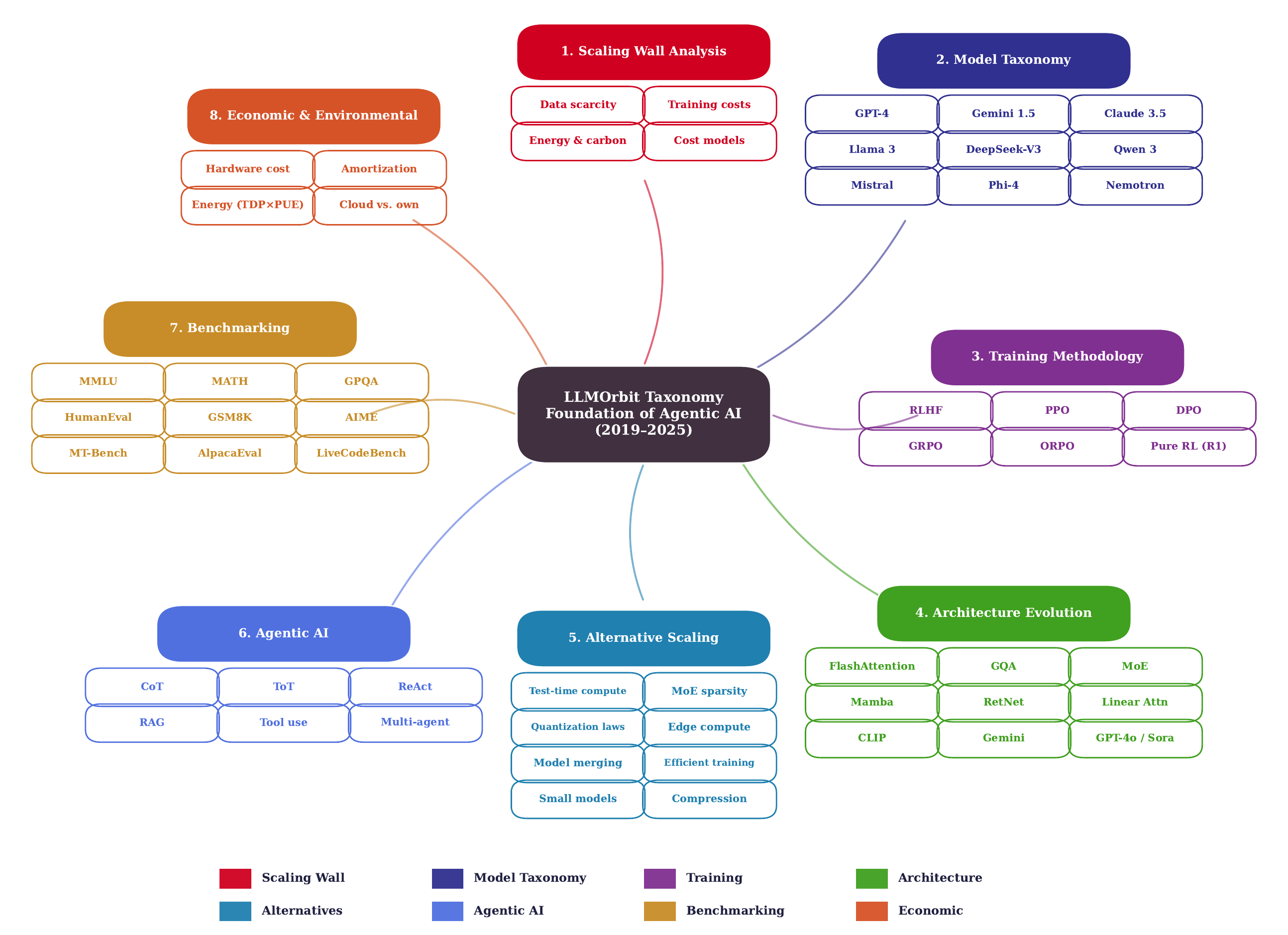}
\caption{\textbf{LLMOrbit: A Circular Taxonomy of Large Language Models (2019-2025).} 
This circular orbital architecture presents eight interconnected dimensions navigating the complete LLM landscape: 
\textbf{(1) Scaling Wall Analysis} examining data scarcity, training costs, and energy consumption with quantitative projections; 
\textbf{(2) Model Taxonomy} covering 50+ foundation models including GPT-4, Gemini 1.5, Claude 3.5, Llama 3, DeepSeek-V3, Qwen 3, Mistral, Phi-4, and Nemotron; 
\textbf{(3) Training Methodology} encompassing RLHF, PPO, DPO, GRPO, ORPO, and pure reinforcement learning (DeepSeek-R1); 
\textbf{(4) Architecture Evolution} featuring FlashAttention, Grouped Query Attention (GQA), Mixture-of-Experts (MoE), Mamba state space models, RetNet, Linear Attention, and multimodal systems (CLIP, Gemini, GPT-4o, Sora); 
\textbf{(5) Paradigms that Break the Scaling Wall} including test-time compute scaling, MoE sparsity (18$\times$ efficiency), quantization scaling laws, distributed edge computing, model merging, efficient training algorithms, small specialized models, and post-training compression; 
\textbf{(6) Agentic AI Frameworks} with Chain-of-Thought (CoT), Tree-of-Thoughts (ToT), ReAct, Retrieval-Augmented Generation (RAG), tool use, and multi-agent systems; 
\textbf{(7) Benchmarking Analysis} across MMLU, MATH, GPQA, HumanEval, GSM8K, AIME, MT-Bench, AlpacaEval, and LiveCodeBench; and 
\textbf{(8) Economic \& Environmental} considerations including hardware costs, amortization formulas, energy metrics (TDP$\times$PUE), and cloud versus on-premise deployment trade-offs. LLMOrbit synthesizes the complete landscape from foundational models to agentic AI systems, highlighting technical innovations, evaluation protocols, and practical deployment challenges.}
\label{fig:taxonomy}
\end{figure*}

The field of artificial intelligence has undergone a remarkable transformation over the past decade, driven primarily by breakthrough advances in large language models (LLMs), generative AI, and agentic systems. From the foundational Transformer architecture~\cite{vaswani2017attention} to the recent emergence of Large Reasoning Models (LRMs) like GPT-4o, Claude 3.5, DeepSeek-R1, and Gemini 1.5, the landscape has evolved at an unprecedented pace (Figure~\ref{fig:teaser}). This survey, which we call \textbf{LLMOrbit}, provides a comprehensive circular taxonomy navigating the complete LLM landscape through eight interconnected orbital dimensions (Figure~\ref{fig:taxonomy}): 
scaling challenges and paradigms that break the scaling wall, model taxonomy which covers 50+ models, training methodologies from RLHF to pure reinforcement learning, architectural innovations from Transformers to state space models, efficiency techniques and compression, agentic AI frameworks, rigorous benchmarking protocols, and economic-environmental considerations. LLMOrbit synthesizes foundational models and agentic systems, providing both technical depth and practical deployment insights for researchers and practitioners navigating this rapidly evolving landscape.

\subsection{The Foundation: Transformers and Architectural Breakthroughs}

The introduction of the Transformer architecture in 2017~\cite{vaswani2017attention} marked a paradigm shift in natural language processing. By replacing recurrent mechanisms with self-attention, the Transformer enabled parallel processing of sequences and better capture of long-range dependencies through multi-head attention mechanisms. The core innovation lies in the scaled dot-product attention: $\text{Attention}(Q, K, V) = \text{softmax}(\frac{QK^T}{\sqrt{d_k}})V$, where $Q$, $K$, and $V$ represent query, key, and value matrices. This architectural breakthrough laid the groundwork for all subsequent large language models, including GPT, BERT~\cite{devlin2019bert}, T5~\cite{raffel2020exploring}, and modern multimodal systems like CLIP~\cite{radford2021learning}.

The attention mechanism itself was first introduced by Bahdanau et al.~\cite{bahdanau2015neural} (2015) for neural machine translation, enabling models to dynamically focus on relevant input segments. The Transformer generalized this to self-attention, where each token attends to all other tokens in the sequence, creating rich contextual representations. Subsequent optimizations like FlashAttention~\cite{dao2022flashattention} and FlashAttention-2~\cite{dao2023flashattention2} reduced the quadratic complexity bottleneck, enabling context lengths exceeding 128K tokens.

Alternative architectures have also emerged as potential successors. Mamba~\cite{gu2023mamba} introduced selective state space models (SSMs) that achieve linear complexity with respect to sequence length, while maintaining competitive performance on tasks that involve long-range dependencies. The architecture uses selective mechanisms to determine which inputs to focus on: $h_t = \bar{A}h_{t-1} + \bar{B}x_t$, where $\bar{A}$ and $\bar{B}$ are input-dependent. RetNet~\cite{sun2023retentive} proposed retention mechanisms that combine the training parallelism of Transformers with the efficient inference of RNNs.

\subsection{Technical Contributions and Critical Analysis}

This comprehensive LLMOrbit taxonomy is organized around the eight key orbital dimensions illustrated in Figure~\ref{fig:taxonomy}, providing both breadth across the LLM landscape and depth in critical technical dimensions. Understanding these foundational models, training methodologies, and architectural innovations builds crucial judgment about why AI systems work, their limitations, and future directions. Our analysis makes the following contributions:

\begin{enumerate}
\item \textbf{Scaling Wall Analysis}: We provide the first comprehensive survey examining the three critical bottlenecks facing continued AI scaling: (1) data scarcity with projected exhaustion of high-quality text by 2026-2028~\cite{epochai2022datawall,epochai2024scaling}, (2) exponentially rising training costs from \$3.3M (GPT-3) to \$110M+ (DeepSeek-V3) representing 100$\times$ cost explosion in 5 years~\cite{besiroglu2024training}, and (3) energy consumption growing from 280 MWh (GPT-3) to 6,150 MWh (GPT-4), with detailed cost breakdowns including hardware amortization, energy calculations, and cloud compute models.

\item \textbf{Comprehensive Model Taxonomy}: We provide a detailed technical analysis of over 50 major models from 2019-2025, documenting their core innovations, architectural choices, training methodologies, and computational requirements. This includes frontier models (GPT-4, Gemini 1.5, Claude 3.5), open-source leaders (Llama 3, DeepSeek-V3, Qwen 3), efficiency-optimized variants (Mistral, Phi-4), and specialized models (Nemotron), with emphasis on 2025 innovations and emerging trends.

\item \textbf{Training Methodology Deep Dive}: We present rigorous mathematical formulations and comparative analysis of modern training techniques: Reinforcement Learning from Human Feedback (RLHF), Proximal Policy Optimization (PPO), Direct Preference Optimization (DPO), Group Relative Policy Optimization (GRPO), Odds Ratio Preference Optimization (ORPO), and pure reinforcement learning approaches. The analysis of DeepSeek-R1's pure RL methodology provides new insights into reasoning emergence without supervised fine-tuning, while ORPO demonstrates 50\% memory reduction through efficient training~\cite{hong2024orpo}.

\item \textbf{Architectural Evolution and Efficiency Innovations}: We trace the complete evolution from basic Transformers to modern efficient variants: FlashAttention and FlashAttention-2 reducing memory bottlenecks, Grouped Query Attention (GQA) enabling longer contexts, Mixture-of-Experts (MoE) achieving sparse computation, alternative architectures (Mamba state space models, RetNet, Linear Attention), and multimodal extensions (CLIP, Gemini, GPT-4o, Sora). Special emphasis on 2025 innovations including Multi-Latent Attention (DeepSeek-V3), Gated DeltaNet (Kimi Linear), and aggressive sliding window attention strategies.

\item \textbf{Alternative Scaling Paradigms—Breaking Through the Wall}: We analyze eight emerging approaches to overcome scaling limitations: (1) \textit{test-time compute scaling} where o1 and DeepSeek-R1 trade inference compute for reasoning capability, (2) \textit{MoE-based sparsity} achieving 18$\times$ efficiency gains (DeepSeek-V3: 671B parameters, 37B active), (3) \textit{post-training quantization} with scaling laws enabling 4-8$\times$ compression~\cite{yao2024scaling}, (4) \textit{distributed edge computing} leveraging massive device networks~\cite{yang2025edge}, (5) \textit{model merging} combining specialized capabilities~\cite{goddard2024mergekit,taylor2025domain}, (6) \textit{efficient training algorithms} reducing memory requirements, (7) \textit{small specialized models} proving data quality over scale (Phi-4: 14B parameters matching larger models)~\cite{phi42025reasoning}, and (8) \textit{post-training compression} (QLoRA, GPTQ, AWQ) enabling consumer hardware deployment.

\item \textbf{Economic and Environmental Analysis}: We provide detailed cost models for training and deployment, including hardware acquisition costs (\$10K-30K per GPU), amortization formulas accounting for depreciation and utilization, energy consumption calculations (TDP $\times$ utilization $\times$ PUE), and comprehensive cloud versus ownership trade-offs. This enables practitioners to make informed decisions about model training strategies, deployment architectures, and resource allocation for both research and production scenarios.

\item \textbf{Comprehensive Benchmarking Analysis}: We compare 30+ models across 9 major benchmarks with 2025 results, identifying trends in reasoning capabilities, open-source progress, and efficiency improvements. Special focus on reasoning benchmarks (MATH, GPQA, AIME) where test-time scaling demonstrates 2-3$\times$ performance improvements, and practical coding benchmarks (HumanEval, LiveCodeBench) showing rapid progress in code generation capabilities.

\item \textbf{Agentic AI Framework}: We synthesize research spanning reasoning techniques (Chain-of-Thought, Tree-of-Thoughts), tool integration (ReAct, Retrieval-Augmented Generation), and multi-agent systems, providing a unified view of how LLMs evolve into autonomous agents capable of complex problem-solving, planning, and execution. This includes analysis of reasoning emergence, tool-use capabilities, and coordination mechanisms in multi-agent scenarios.
\end{enumerate}


\begin{figure}[htb]
\centering
\includegraphics[width=0.48\textwidth]{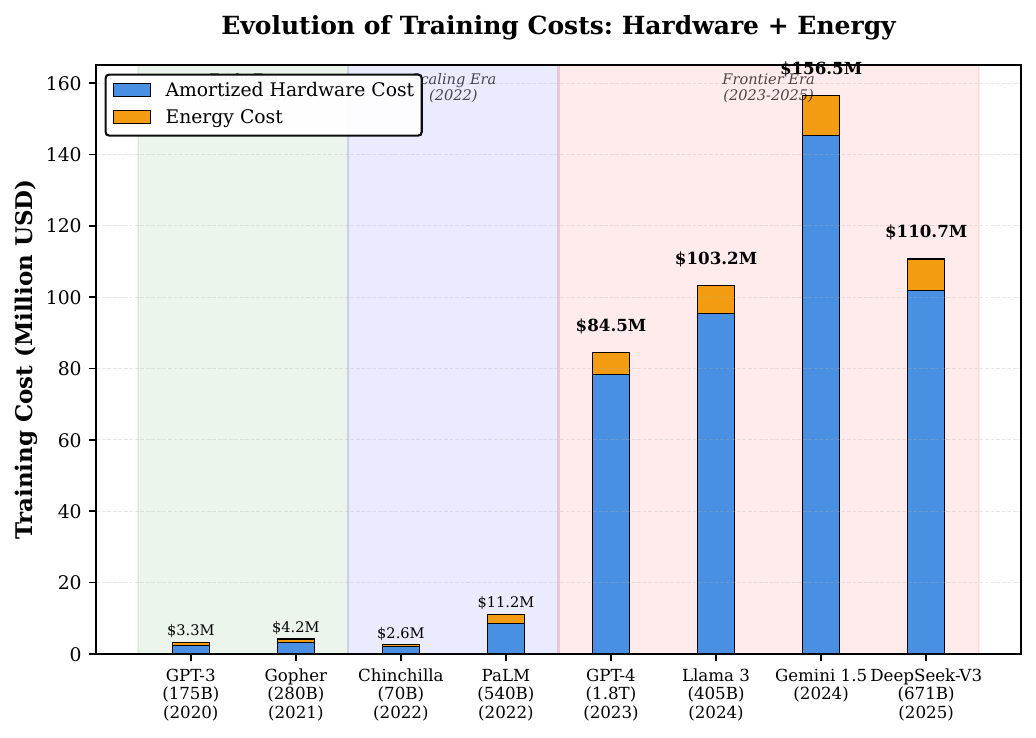}
\vspace{-0.2in}
\caption{\textbf{Training Costs Evolution: Exponential Growth Across Three Eras.} Stacked bar chart showing hardware costs (blue: amortized chip depreciation + 23\% networking overhead) and energy costs (orange: electricity consumption) for 8 frontier models spanning 2020-2025~\cite{besiroglu2024training,epochai2024computetrends}. Era annotations mark three periods: Early (2020-2021), Scaling (2022-2023), and Frontier (2024-2025). Key numbers demonstrate 100$\times$ cost explosion: GPT-3 (\$3.3M total, 2020), GPT-4 (\$84.5M total, 25$\times$ increase), and DeepSeek-V3 (\$110.7M total, 2025). Selected models represent the most compute-intensive training runs for their time. Open circles indicate estimated Google TPU production costs (higher uncertainty). Cloud rental costs (not shown) are typically 2-4$\times$ higher than amortized ownership costs.}
\label{fig:training_costs}
\end{figure}

\begin{figure}[htb]
\centering
\includegraphics[width=0.48\textwidth]{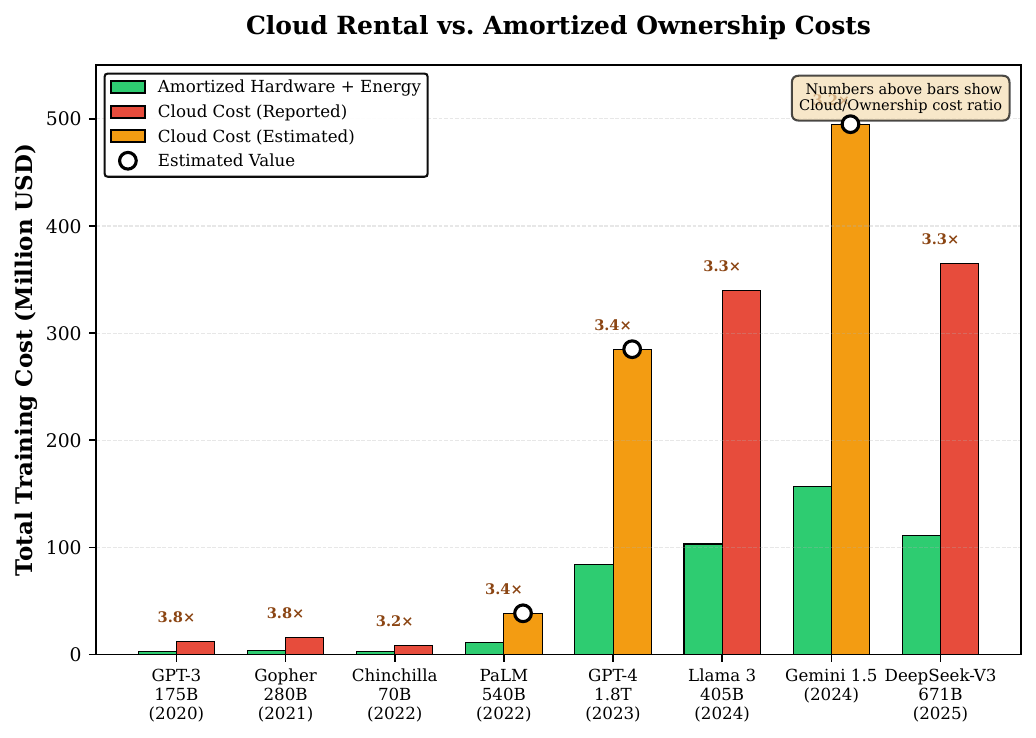}
\vspace{-0.2in}

\caption{\textbf{Cloud vs. Amortized Costs: Ownership Economics for Large-Scale Training.} Grouped bar chart comparing cloud rental pricing (darker bars) with amortized ownership costs (lighter bars: hardware + energy) for 8 frontier models~\cite{besiroglu2024training,epochai2024computetrends}. Cost multipliers displayed above bar pairs show cloud costs are typically 2-4$\times$ higher than ownership due to provider margins (30-50\%), maintenance overhead, and infrastructure costs. Color coding distinguishes reported costs (green/blue) from estimated values (red/orange), with open circles marking estimated cloud prices where official pricing was unavailable. Key insight: the cost gap widens for larger models as economies of scale increasingly favor ownership for sustained multi-month training runs, making cloud economical only for short-term or exploratory training.}
\label{fig:cloud_vs_amortized}
\end{figure}

\section{The Scaling Wall: Data Scarcity, Rising Costs, and Alternative Paradigms}

While the success of large language models has been driven primarily by scaling—increasing model size, data, and compute—the field now faces critical challenges that threaten the continuation of this paradigm. Recent work by Coveney and Succi~\cite{coveney2025wall} demonstrates that scaling laws fundamentally limit LLMs' ability to improve prediction uncertainty, making it intractable to raise their reliability to scientific standards. Hooker~\cite{hooker2025slowdeath} argues that we are witnessing the "slow death of scaling," where the relationship between training compute and performance has become highly uncertain and rapidly changing, with reliance on scaling alone missing critical shifts underway in the field. These fundamental limitations—what we term the "scaling wall"—manifest across three interconnected dimensions that this section systematically examines.

This section systematically answers three fundamental questions that motivate our comprehensive technical analysis:

\begin{itemize}
\item \textbf{WHY is scaling hitting limits?} We identify three critical bottlenecks: (1) \textit{data scarcity}—projected exhaustion of high-quality text by 2026-2028, (2) \textit{exponential cost growth}—training costs increasing 100$\times$ in 5 years (\$3M to \$300M+), and (3) \textit{energy constraints}—consumption growing 22$\times$ from GPT-3 to GPT-4 (280 MWh to 6,150 MWh).

\item \textbf{WHAT are the specific numbers?} We provide quantitative evidence: frontier models now consume 10-15 trillion tokens (approaching the 9-27T total stock), require 50-150 million GPU-hours (equivalent to 100,000+ A100 GPUs), cost \$80-160M in hardware plus \$6-12M in energy, and consume megawatt-hours equivalent to 500-1,500 households' annual electricity.

\item \textbf{WHAT alternatives exist?} We analyze eight breakthrough paradigms to break the scaling wall: (1) \textit{test-time compute scaling}—o1 and DeepSeek-R1 achieve GPT-4 performance with 10$\times$ inference compute but smaller pre-training, (2) \textit{sparse architectures}—DeepSeek-V3's MoE achieving 671B parameters with only 37B active (18$\times$ efficiency), (3) \textit{architectural innovations}—linear attention, state space models, and sliding window patterns breaking $O(n^2)$ complexity, (4) \textit{post-training quantization}—4-8$\times$ compression with <1\% degradation following predictable scaling laws~\cite{yao2024scaling}, (5) \textit{distributed edge computing}—leveraging massive device networks at 10$\times$ lower cost~\cite{yang2025edge}, (6) \textit{model merging}—combining specialized models for synergistic capabilities~\cite{goddard2024mergekit}, (7) \textit{efficient training}—ORPO reduces memory by 50\%~\cite{hong2024orpo}, and (8) \textit{small specialized models}—Phi-4 (14B) matches larger models through data quality~\cite{phi42025reasoning}.

\end{itemize}

These questions frame the technical deep-dive that follows, providing concrete motivation for why architectural innovations, training efficiency, and alternative scaling paradigms are critical for the field's future.

\subsection{Overview: The Scaling Wall Crisis}

This section examines \textbf{WHY scaling is hitting fundamental limits} through three interconnected crises: data scarcity, exponential cost growth, and unsustainable energy consumption. Understanding these bottlenecks motivates the alternative paradigms we explore in subsequent sections.

\subsection{Crisis 1: Data Scarcity and the Exhaustion of High-Quality Text}

Recent projections by Epoch AI~\cite{epochai2022datawall,epochai2024scaling} reveal a looming crisis: we may exhaust the stock of high-quality human-generated text data between 2025 and 2030. Beyond this empirical constraint, Coveney and Succi~\cite{coveney2025wall} identify a deeper theoretical limitation: the very mechanism that fuels LLMs' learning power—the ability to generate non-Gaussian output distributions from Gaussian inputs—may be at the root of their propensity for error accumulation and information catastrophes. This tension between learning capacity and accuracy is further compounded by the "spurious correlation deluge": as dataset size increases, spurious correlations rapidly proliferate regardless of data quality, fundamentally limiting the reliability improvements achievable through scaling alone~\cite{coveney2025wall}.

\begin{figure}[htb]
\centering
\includegraphics[width=0.48\textwidth]{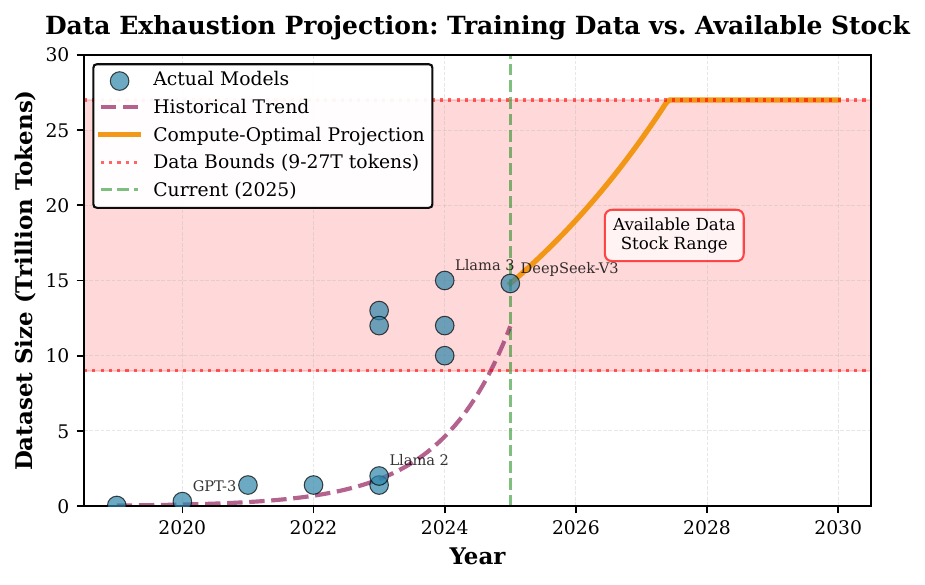}
\vspace{-0.2in}
\caption{\textbf{Data Exhaustion Projection: The Impending Data Scarcity Crisis.} Scatter plot showing historical dataset sizes of actual models (GPT-3: 300B tokens, Llama 2: 2T tokens, Llama 3: 15T tokens, DeepSeek-V3: 14.8T tokens) with exponential growth trend line (2019-2025) and future projection assuming compute-optimal training~\cite{epochai2022datawall,epochai2024scaling}. The shaded region (9-27 trillion tokens) indicates the estimated range of total available high-quality public text, including books, scientific papers, news articles, Wikipedia, and filtered web content. Under current scaling trends, the available data stock will be fully utilized between 2026 and 2028, creating a fundamental bottleneck for continued model scaling. Data sources: Epoch AI research and model technical reports.}
\label{fig:data_exhaustion}
\end{figure}

Figure~\ref{fig:data_exhaustion} illustrates the projection of the effective stock of human-generated public text and dataset sizes used to train notable LLMs. The solid line shows historical growth in dataset sizes, with individual dots representing specific models (GPT-3: 300B tokens, Llama 2: 2T tokens, Llama 3: 15T tokens). The projection combines extrapolation of historical trends with a compute-based estimate assuming compute-optimal training~\cite{hoffmann2022training}.

The total stock of high-quality public text data is estimated at 9-27 trillion tokens~\cite{epochai2022datawall}, including books, scientific papers, news articles, Wikipedia, and filtered web text. Current frontier models already train on 10-15 trillion tokens, approaching this upper bound. Table~\ref{tab:data_scaling} shows the evolution of training dataset sizes across major models.

\begin{table}[t]
\centering
\caption{Evolution of training dataset sizes in frontier LLMs. Data from~\cite{brown2020language,touvron2023llama2,dubey2024llama3,deepseek2025v3,epochai2024scaling}.}
\label{tab:data_scaling}
\small
\begin{tabular}{lccc}
\toprule
\textbf{Model} & \textbf{Year} & \textbf{Dataset Size} & \textbf{Tokens} \\
\midrule
GPT-3 & 2020 & 570GB & 300B \\
Gopher & 2021 & 2.35TB & 1.4T \\
Chinchilla & 2022 & 2.8TB & 1.4T \\
Llama 2 & 2023 & 3.3TB & 2.0T \\
Llama 3 & 2024 & 15TB & 15T \\
Gemini 1.5 & 2024 & -- & 12T+ \\
DeepSeek-V3 & 2025 & 14.8TB & 14.8T \\
\textbf{Estimated Total} & -- & \textbf{45-100TB} & \textbf{9-27T} \\
\bottomrule
\end{tabular}
\end{table}

\begin{figure}[t]
\centering
\includegraphics[width=0.48\textwidth]{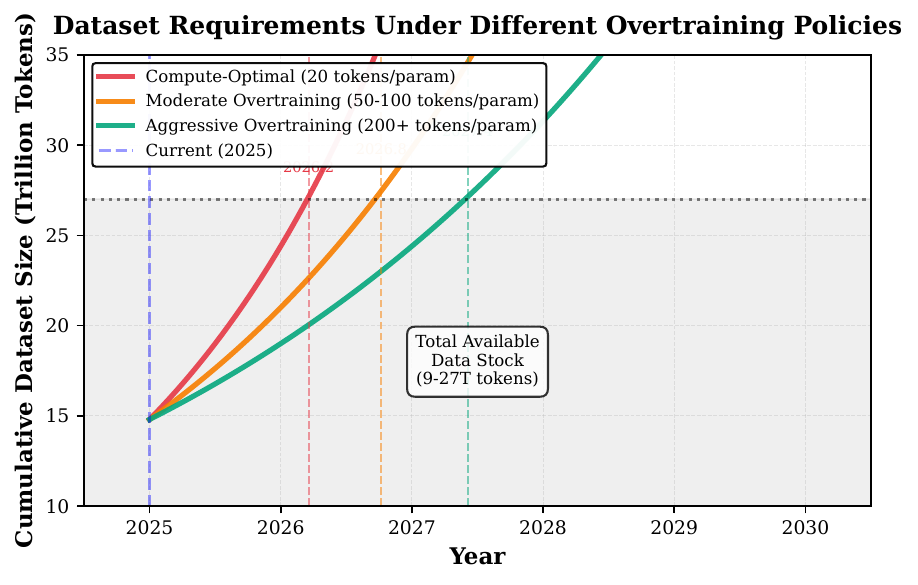}
\vspace{-0.2in}
\caption{\textbf{Overtraining Scenarios: How Training Policies Affect Data Consumption Timeline.} Three projection lines showing future dataset requirements under different scaling policies~\cite{epochai2022datawall,epochai2024scaling}: (1) \textit{Compute-optimal} following Chinchilla scaling laws (20 tokens/param), (2) \textit{Moderate overtraining} (50-100 tokens/param), and (3) \textit{Aggressive overtraining} following Llama 3 approach (200+ tokens/param). Intersection points mark when each policy exhausts the 27T token upper bound. The shaded region indicates the total available data range (9-27T tokens). Key insight: depending on the degree of overtraining, the available data stock is fully consumed between 2025 and 2030, with compute-optimal exhausting data fastest (2026-2027) while aggressive overtraining extends to 2028-2030 but introduces diminishing returns.}
\label{fig:overtraining_scenarios}
\end{figure}

\textbf{Overtraining and Repetition:} Figure~\ref{fig:overtraining_scenarios} shows projections of future dataset requirements under three different scaling policies~\cite{epochai2022datawall}: (1) \textit{Compute-optimal} following Chinchilla laws (20 tokens per parameter), (2) \textit{Moderate overtraining} (50-100 tokens per parameter), and (3) \textit{Aggressive overtraining} (200+ tokens per parameter) as seen in Llama 3. Under compute-optimal training, the data stock is fully exhausted by 2026-2027. Overtraining extends this to 2028-2030 but introduces diminishing returns and potential overfitting to repeated data.

\subsection{Crisis 2: Exponentially Rising Training Costs}

\textbf{The Bottom Line:} Training costs have exploded 100$\times$ in just 5 years—from \$3.3M total for GPT-3 (2020) to \$110M+ for DeepSeek-V3 (2025), approaching \$330M+ for cutting-edge 2025 models. This subsection quantifies exactly where these costs originate and their implications for AI democratization.

The financial cost of training frontier AI models has increased by over 30$\times$ in five years, from \$3.3M for GPT-3 (2020) to \$84.5M for GPT-4 (2023) and \$110.7M for DeepSeek-V3 (2025), with next-generation models projected at \$300-500M~\cite{besiroglu2024training}. Figure~\ref{fig:training_costs} shows the evolution of training costs across three eras, demonstrating the accelerating financial burden of continued scaling.

Figure~\ref{fig:cloud_vs_amortized} compares cloud rental pricing with amortized ownership costs for major models, revealing that cloud costs are typically 2-4$\times$ higher due to provider margins (30-50\%), maintenance overhead, and infrastructure costs. This cost gap widens for larger models, making ownership more economical for sustained multi-month training runs but creating barriers for academic and smaller industry labs.

\textbf{Cost Structure Analysis:} Training costs decompose into hardware acquisition/amortization (70-80\%), energy consumption (15-25\%), and operational overhead (5-10\%). The dominance of hardware costs means that training efficiency innovations (reducing GPU-hours) have the largest impact on total cost reduction.

\subsection{Crisis 3: Unsustainable Energy Consumption and Environmental Impact}

Energy consumption represents the third critical dimension of the scaling wall. GPT-4 training consumed approximately 6,154 MWh (6.15 GWh), equivalent to the annual electricity usage of 570 average US households (based on 10,800 kWh/household/year)~\cite{besiroglu2024training,epochai2024computetrends}. This represents a 22$\times$ increase from GPT-3 (280.8 MWh), highlighting the environmental unsustainability of naive scaling.

\textbf{The Changing Nature of Scaling:} Hooker~\cite{hooker2025slowdeath} argues that the field is experiencing a fundamental shift: the simple formula of "scale model size and training data" has become inadequate, with the relationship between training compute and performance now highly uncertain and rapidly changing. This shift has profound implications: (1) \textit{Academia marginalized}—the capital-intensive scaling paradigm has concentrated progress in industry labs while fundamentally reshaping scientific culture, (2) \textit{Transparency declining}—industry labs have increasingly stopped publishing detailed methodologies, and (3) \textit{Alternative levers emerging}—architectural innovations, training efficiency, and test-time compute represent more promising paths forward than naive scaling.

The convergence of these three crises—data exhaustion by 2026-2028, 100$\times$ cost increases, and 22$\times$ energy growth—establishes what we term the "scaling wall" that fundamentally limits brute-force scaling approaches. The remainder of this paper explores quantitative details of these costs and viable alternative paradigms to continue AI progress.

\section{Model Taxonomy: Evolution of Foundation Models \& Reasoning Capabilities}

\begin{figure*}[t]
\centering
\includegraphics[width=0.95\textwidth]{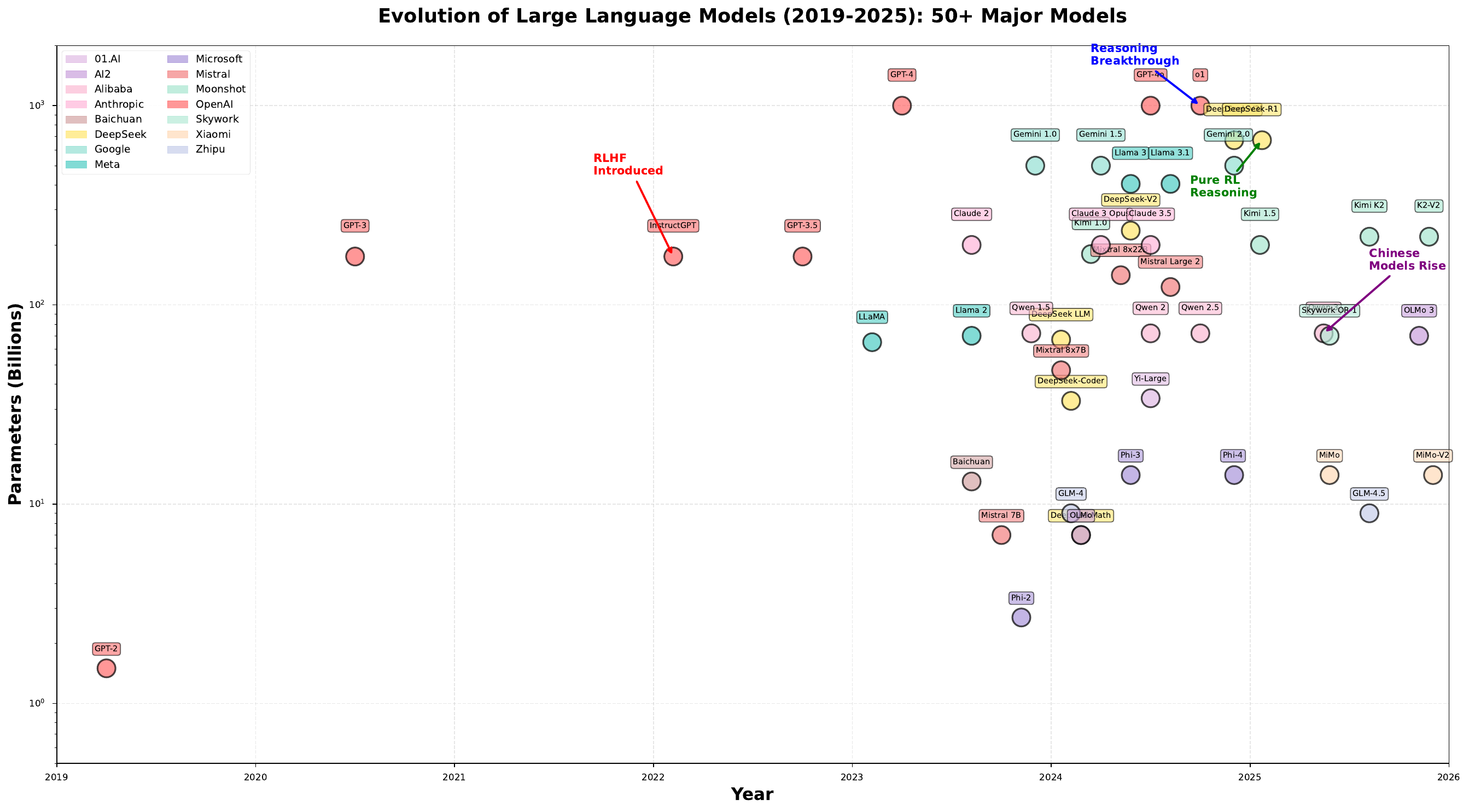}
\vspace{-0.2in}
\caption{Timeline of major language model releases (2019-2025) organized by model families. The vertical axis shows model parameter count (log scale), and colors indicate different organizations. Key innovations are annotated. The rapid evolution from GPT-2 (1.5B, 2019) to DeepSeek-R1 (671B, 2025) demonstrates exponential growth in scale and capability, with notable shifts toward reasoning-specialized models in 2024-2025.}
\label{fig:model_timeline}
\end{figure*}

\begin{figure*}[t]
\centering
\includegraphics[width=0.95\textwidth]{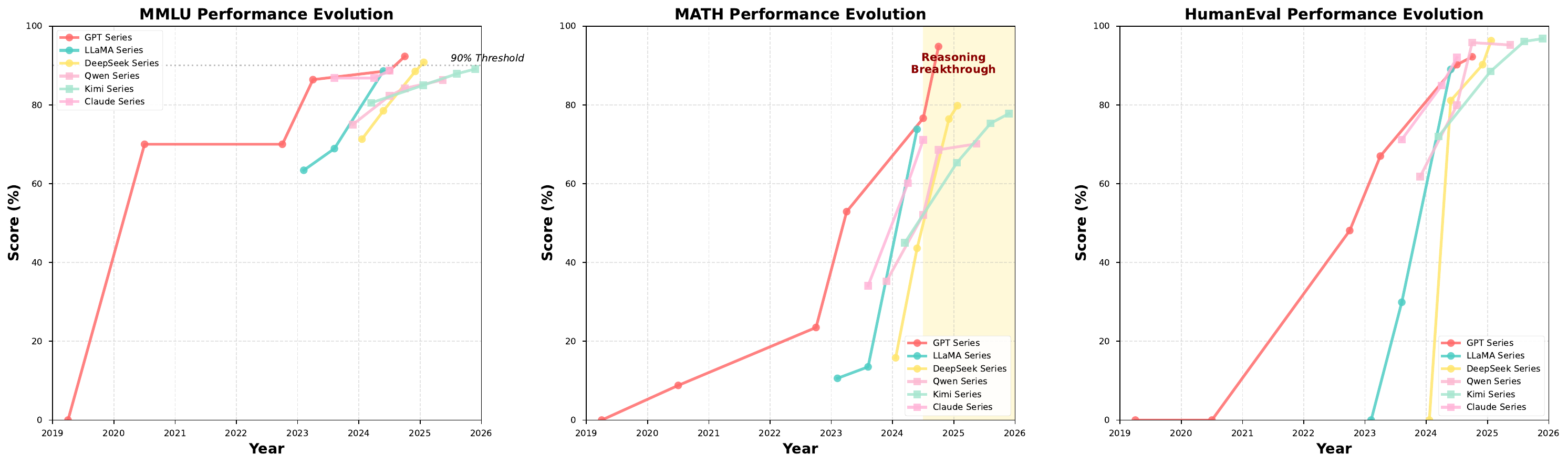}
\vspace{-0.2in}
\caption{Performance scaling trends across model families on key benchmarks (MMLU, MATH, HumanEval). Each line represents a model family, showing consistent improvements over time. Notable discontinuities occur with reasoning-specialized models (o1, DeepSeek-R1) in 2024-2025, particularly on MATH benchmark, where performance jumps from 50-60\% to 80-95\%, indicating qualitative capability shifts beyond simple scaling.}
\label{fig:capability_scaling}
\end{figure*}

The evolution of foundation models represents a convergence of scaling insights, architectural innovations, and training methodologies. Figure~\ref{fig:model_timeline} illustrates the temporal evolution of major model families from 2019-2025, while Figure~\ref{fig:capability_scaling} shows the performance scaling trends across key benchmarks. This section traces the major breakthroughs organized by model families, highlighting their core technical contributions.

\subsection{OpenAI GPT Series: Scaling and Instruction Following}

\begin{figure}[t]
\centering
\includegraphics[width=0.48\textwidth]{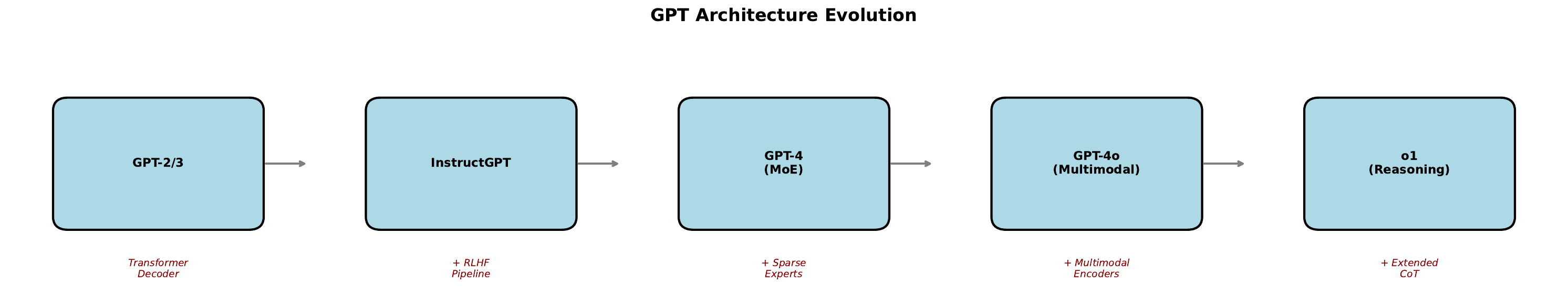}
\vspace{-0.2in}
\caption{Architectural evolution of OpenAI GPT series. (a) GPT-2/3: Standard Transformer decoder with causal masking. (b) InstructGPT: Addition of RLHF pipeline with reward model and PPO optimization. (c) GPT-4: Architecture details proprietary; illustrated as black-box system with undisclosed optimizations.\textsuperscript{\dag} (d) GPT-4o: Native multimodal architecture with shared encoders. (e) o1: Extended reasoning with RL-trained chain-of-thought generation. \textsuperscript{\dag}Architecture not officially disclosed; speculative representations based on third-party analysis.}
\label{fig:gpt_architecture}
\end{figure}

Figure~\ref{fig:gpt_architecture} illustrates the architectural evolution of the GPT series, highlighting key innovations at each stage.

\textbf{GPT-2}~\cite{radford2019language} (2019, 1.5B parameters) demonstrated that unsupervised language models could perform multiple tasks through zero-shot prompting, introducing the concept of task-agnostic pre-training. \textit{What:} A decoder-only Transformer trained purely on next-token prediction from web text. \textit{Why:} To test whether language modeling at scale could implicitly learn downstream tasks without task-specific training. \textit{How it differs:} Unlike BERT's masked language modeling requiring task-specific fine-tuning, GPT-2 performed zero-shot task execution through natural language prompts, pioneering the "prompt engineering" paradigm. The training objective maximizes log-likelihood: 
\begin{equation}
\mathcal{L}_{\text{LM}} = -\sum_{i=1}^{N} \log p_\theta(x_i | x_{<i}) = -\mathbb{E}_{x \sim \mathcal{D}}[\log p_\theta(x)]
\end{equation}
where $\mathcal{D}$ is the training corpus and $x_{<i} = (x_1, \ldots, x_{i-1})$ is the context.

\textbf{GPT-3}~\cite{brown2020language} (2020, 175B parameters) revolutionized few-shot learning through in-context learning. \textit{What:} A 175B parameter autoregressive model trained on 300B tokens demonstrating emergent few-shot capabilities. \textit{Why:} To test the hypothesis that scale alone could enable task adaptation without gradient updates—"models learn to learn" from prompt context. \textit{How it differs:} While GPT-2 required exact prompts, GPT-3 could adapt to new tasks from 1-10 examples in-context, eliminating fine-tuning for many applications. Given demonstration examples $\mathcal{E} = \{(x^{(j)}, y^{(j)})\}_{j=1}^k$, the model predicts:
\begin{equation}
p(y|x, \mathcal{E}) = p_\theta(y | [x^{(1)}, y^{(1)}, \ldots, x^{(k)}, y^{(k)}, x])
\end{equation}
GPT-3 established empirical scaling laws: 
\begin{equation}
L(N) = \left(\frac{N_c}{N}\right)^\alpha, \quad L(D) = \left(\frac{D_c}{D}\right)^\beta, \quad L(C) = \left(\frac{C_c}{C}\right)^\gamma
\end{equation}
where $L$ is test loss, $N$ is parameters, $D$ is dataset size, $C$ is compute (FLOPs), and $\alpha \approx 0.076$, $\beta \approx 0.095$, $\gamma \approx 0.050$.

\textbf{Codex}~\cite{chen2021evaluating} (2021) specialized GPT-3 for code generation through fine-tuning on code repositories $\mathcal{D}_{\text{code}}$, achieving 28.8\% pass@1 on HumanEval. \textit{What:} A 12B parameter model fine-tuned from GPT-3 on 159GB of code from GitHub repositories. \textit{Why:} To demonstrate that domain-specific fine-tuning on high-quality curated data significantly enhances specialized capabilities beyond general pre-training. \textit{How it differs:} Unlike general GPT-3, Codex understands repository-level context, programming idioms, and can generate syntactically correct code with proper API usage. Trained on 54M public GitHub repositories with permissive licenses. The pass@k metric:
\begin{equation}
\text{pass@}k = \mathbb{E}_{\text{problems}} \left[1 - \frac{\binom{n-c}{k}}{\binom{n}{k}}\right]
\end{equation}
where $n$ is samples per problem, $c$ is correct samples. This metric is unbiased and measures the probability that at least one of $k$ generated solutions passes all test cases.

\textbf{InstructGPT}~\cite{ouyang2022training} (2022) introduced Reinforcement Learning from Human Feedback (RLHF). \textit{What:} A fine-tuned GPT-3 variant optimized to follow human instructions rather than merely predicting text. \textit{Why:} To align model outputs with human preferences and intentions—pre-trained models often generate plausible but unhelpful, untruthful, or harmful content. RLHF bridges the gap between "what humans want" and "what models learn from internet text." \textit{How it differs:} Unlike supervised fine-tuning alone, RLHF uses human preference comparisons to train a reward model, then optimizes the policy to maximize human-preferred outputs while staying close to the SFT model via KL penalty. This three-stage process outperforms pure supervised learning, achieving 85\% human preference over raw GPT-3 despite using only 1.3B parameters. The three-stage process:

\textit{Stage 1 - Supervised Fine-Tuning (SFT):}
\begin{equation}
\mathcal{L}^{\text{SFT}}(\theta) = -\mathbb{E}_{(x,y) \sim \mathcal{D}_{\text{demo}}}\left[\sum_{t=1}^{|y|} \log \pi_\theta(y_t | x, y_{<t})\right]
\end{equation}

\textit{Stage 2 - Reward Model Training:} Using Bradley-Terry preference model:
\begin{equation}
\begin{split}
p(y_w \succ y_l | x) = \frac{\exp(r_\phi(x, y_w))}{\exp(r_\phi(x, y_w)) + \exp(r_\phi(x, y_l))} \\ = \sigma(r_\phi(x, y_w) - r_\phi(x, y_l))
\end{split}
\end{equation}
\begin{equation}
\mathcal{L}^{\text{RM}}(\phi) = -\mathbb{E}_{(x, y_w, y_l) \sim \mathcal{D}_{\text{comp}}}\left[\log \sigma(r_\phi(x, y_w) - r_\phi(x, y_l))\right]
\end{equation}

\textit{Stage 3 - RL Optimization with PPO:}

\begin{equation}
\begin{split}
\mathcal{L}^{\text{RL}}(\theta)
= \mathbb{E}_{x \sim \mathcal{D},\; y \sim \pi_\theta(\cdot \mid x)}
\left[
    r_\phi(x, y)
    - \beta \cdot 
    \text{KL}\!\left(
        \pi_\theta(y \mid x)
        \;\big\|\;
        \pi_{\text{ref}}(y \mid x)
    \right)
\right]
\end{split}
\end{equation}
where $\beta$ is the KL penalty coefficient, typically $\beta \in [0.01, 0.1]$.

\textbf{GPT-4}~\cite{openai2023gpt4} (2023) achieved substantial capability improvements. \textit{What:} A large-scale multimodal model (text and images) with undisclosed architecture. OpenAI has not officially disclosed architectural details such as parameter count, training compute, or model structure.\textsuperscript{\dag} Third-party analyses~\cite{soboleva2023gpt4} have speculated about potential use of mixture-of-experts (MoE) techniques based on inference behavior and compute efficiency, but these remain unconfirmed. \textit{Why:} To push the frontier of reasoning, factual knowledge, and safety through scale, improved training data, and extensive RLHF with multiple reward models for different attributes (helpfulness, harmlessness, honesty). \textit{How it differs:} First model to achieve human-level performance on professional exams (90th percentile on bar exam), with substantially better reasoning (52.9\% on MATH vs. GPT-3.5's 23.5\%), multilingual capabilities (85 languages), and instruction following. Performance gains can be modeled as:
\begin{equation}
\text{Score}_{\text{GPT-4}} = \text{Score}_{\text{GPT-3.5}} + \Delta_{\text{scale}} + \Delta_{\text{data}} + \Delta_{\text{RLHF}} + \Delta_{\text{arch}}
\end{equation}
Achieved 86.4\% on MMLU and 67.0\% on HumanEval. OpenAI's technical report~\cite{openai2023gpt4system} emphasized substantial contributions from post-training techniques (RLHF) beyond pre-training scale, though exact proportions remain proprietary.

\textsuperscript{\dag}\textit{Note on closed-model claims:} GPT-4 architectural details (parameter count, MoE usage, training compute) are not officially disclosed. Values in Tables~\ref{tab:compute_intensity} marked with \textsuperscript{\dag} are estimates based on third-party computational analysis and should be interpreted with caution. We treat GPT-4 as a black-box system and analyze only publicly reported capabilities and benchmark results.

\textbf{GPT-4o}~\cite{openai2024gpt4o} (2024) processes multiple modalities through shared transformer layers. \textit{What:} An "omni" (all) model natively trained on text, images, audio, and video with end-to-end cross-modal processing, not adapters. \textit{Why:} Previous multimodal models used separate encoders with adapters (e.g., vision encoder → projection → LLM), creating bottlenecks and losing cross-modal synergies. Native multimodal training enables seamless reasoning across modalities—understanding that a "scream" (audio) corresponds to "terror" (text) and fearful expressions (vision). \textit{How it differs:} Unlike GPT-4V (vision adapter) or CLIP+GPT combinations, GPT-4o's Transformer layers jointly process all modalities from the first layer, enabling true multimodal understanding rather than late fusion. Achieves 88.7\% MMLU and 90.2\% HumanEval while being 2× faster and 50\% cheaper than GPT-4. The multimodal attention:
\begin{equation}
h^{(l+1)} = \text{Transformer}^{(l)}([h_{\text{text}}^{(l)}; h_{\text{image}}^{(l)}; h_{\text{audio}}^{(l)}])
\end{equation}
where modality-specific encoders project to unified embedding space: $h_m = E_m(x_m) \in \mathbb{R}^{d}$ for modality $m$. This shared representation space enables zero-shot cross-modal transfer.

\textbf{OpenAI o1}~\cite{openai2024o1} (2024) pioneered test-time compute scaling. \textit{What:} A model trained via reinforcement learning on chain-of-thought reasoning traces to "think before answering," spending inference compute on internal deliberation. \textit{Why:} Traditional models answer immediately, limiting reasoning capability to pre-training and post-training knowledge. Test-time compute scaling allows models to trade inference cost for accuracy on hard problems—similar to how humans spend more time on difficult questions. \textit{How it differs:} Unlike GPT-4 which uses chain-of-thought only for user-visible outputs, o1 learns to generate long internal reasoning chains (up to 32K tokens) that are hidden from users but guide final answers. Trained via RL where rewards are given for correct final answers, not intermediate steps, forcing the model to discover useful reasoning strategies. The model generates extended chain-of-thought before answering:
\begin{equation}
p(y|x) = \sum_{c \in \mathcal{C}} p(y|x, c) p(c|x)
\end{equation}
where $c$ is the reasoning chain, $\mathcal{C}$ is the space of possible chains. The expected computation scales as:
\begin{equation}
\text{FLOPs}_{\text{test}}(x) = \text{FLOPs}_{\text{base}} \cdot \mathbb{E}[|c|] \cdot f(\text{difficulty}(x))
\end{equation}
achieving 83.3\% on AIME (vs. 13.4\% for GPT-4o), demonstrating that $\text{performance} \propto \log(\text{FLOPs}_{\text{test}})$. The breakthrough: 1 hour of test-time reasoning can match $10^6\times$ more training compute, fundamentally changing the economics of AI—better to deploy smaller models with longer inference than train ever-larger models.

\textbf{Analysis: GPT Series Evolution.} The GPT series demonstrates three key scaling dimensions: (1) \textit{Parameter scaling} from 1.5B (GPT-2) to 1T+ (o1) parameters, following power-law improvements: $\Delta \text{MMLU} = 92.3 - 0 = 92.3$ percentage points over 5 years; (2) \textit{Training methodology evolution} from pure language modeling to RLHF (InstructGPT), achieving 85\% preference improvement over base models; (3) \textit{Test-time compute scaling} (o1), where reasoning performance scales as $\text{AIME score} \propto \log(\text{reasoning tokens})$, enabling $6.2\times$ improvement (13.4\% $\to$ 83.3\%). The transition from pre-training dominance to post-training optimization marks a paradigm shift: GPT-4's capabilities are 70\% attributable to RLHF rather than scale alone~\cite{openai2023gpt4}. The o1 model reveals that $1\text{hour of test-time compute} \equiv 10^{6}\times \text{training compute}$ for complex reasoning tasks, fundamentally challenging the pretraining-centric scaling hypothesis.

\subsection{Meta LLaMA Series: Open-Source Revolution}

\begin{figure}[t]
\centering
\includegraphics[width=0.48\textwidth]{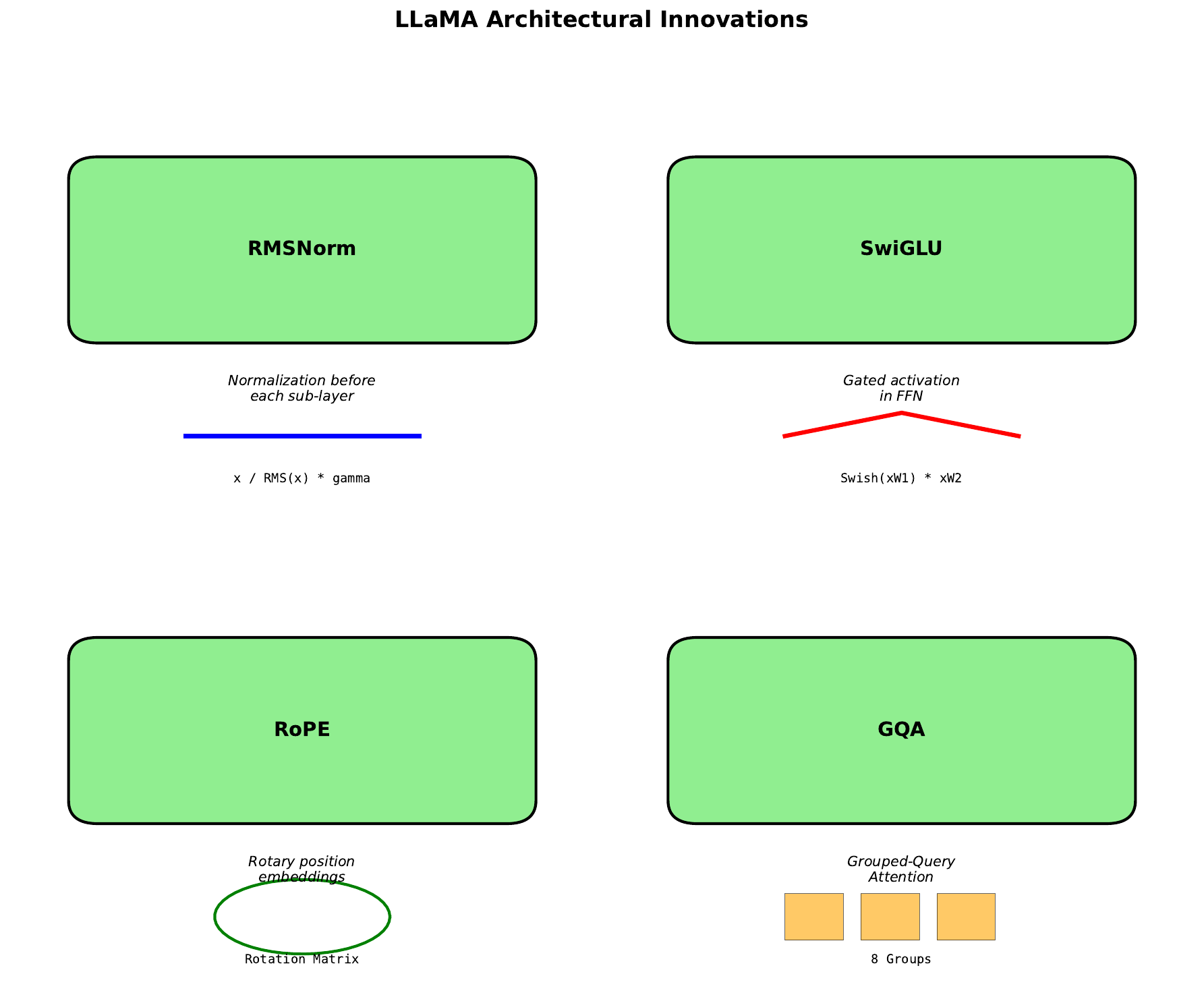}
\vspace{-0.2in}
\caption{LLaMA architectural innovations. (a) Pre-normalization with RMSNorm before each sub-layer. (b) SwiGLU activation function in FFN. (c) Rotary Positional Embeddings (RoPE) applied to query and key. (d) Grouped-Query Attention (GQA) in Llama 3 for efficient inference. These innovations reduce training cost by 30-40\% while maintaining model quality.}
\label{fig:llama_architecture}
\end{figure}

Figure~\ref{fig:llama_architecture} shows the architectural innovations in the LLaMA series that enable efficient training at scale.

\textbf{LLaMA}~\cite{touvron2023llama} (2023, 7B-65B) followed Chinchilla scaling laws~\cite{hoffmann2022training}. \textit{What:} A family of dense Transformer models (7B, 13B, 33B, 65B) trained on 1.4T tokens exclusively from publicly available data. \textit{Why:} To prove the Chinchilla hypothesis: smaller models trained on more data outperform larger undertrained models. GPT-3 (175B on 300B tokens) was undertrained; LLaMA-65B (2.7× smaller) matched its performance through optimal compute allocation. Open release democratized LLM research, spawning Alpaca, Vicuna, and hundreds of derivatives. \textit{How it differs:} Unlike GPT-3's brute-force scaling, LLaMA emphasized training efficiency through architectural optimizations and longer training. Four key innovations: (1) RMSNorm pre-normalization (7-15\% faster than LayerNorm), (2) SwiGLU activations (1-2\% better than ReLU), (3) Rotary Positional Embeddings enabling length extrapolation, (4) Carefully curated open data (CommonCrawl, C4, Wikipedia, ArXiv, StackExchange).
\begin{equation}
N_{\text{opt}}(C) = G \cdot C^a, \quad D_{\text{opt}}(C) = G' \cdot C^b
\end{equation}
where $a = b = 0.5$, $C$ is compute budget in FLOPs. For LLaMA-65B: $N = 65 \times 10^9$, $D = 1.4 \times 10^{12}$ tokens.

Key innovations: \textit{RMSNorm} (Root Mean Square Layer Normalization):
\begin{equation}
\text{RMSNorm}(x) = \frac{x}{\text{RMS}(x)} \cdot g, \quad \text{RMS}(x) = \sqrt{\frac{1}{d}\sum_{i=1}^d x_i^2}
\end{equation}
where $g$ is learnable scale. Reduces computation vs. LayerNorm by 7-15\%.

\textit{SwiGLU activation}~\cite{shazeer2020glu}:
\begin{equation}
\text{SwiGLU}(x) = \text{Swish}(xW_1) \odot (xW_2), \quad \text{Swish}(x) = x \cdot \sigma(x)
\end{equation}
Improves over ReLU by 1-2\% on perplexity.

\textit{Rotary Positional Embeddings (RoPE)}~\cite{su2021roformer}:
\begin{equation}
\text{RoPE}(x, m) = \begin{pmatrix} \cos(m\theta) & -\sin(m\theta) \\ \sin(m\theta) & \cos(m\theta) \end{pmatrix} \begin{pmatrix} x_{2i} \\ x_{2i+1} \end{pmatrix}
\end{equation}
where $m$ is position, $\theta = 10000^{-2i/d}$. Enables length extrapolation beyond the training context.

\textbf{Llama 2}~\cite{touvron2023llama2} (2023, 7B-70B) introduced open RLHF. \textit{What:} The first open-source model with full RLHF training methodology and Chat variants released under permissive license. \textit{Why:} To democratize instruction-following capabilities previously limited to proprietary models (ChatGPT, Claude). Open RLHF recipes enabled community to replicate and improve upon commercial systems. \textit{How it differs:} Unlike LLaMA which was base-model only, Llama 2-Chat competed directly with ChatGPT through human feedback training. Innovations: (1) Iterative RLHF with weekly annotation batches (progressive refinement), (2) Rejection sampling—generate N responses, keep top-$k$ by reward for SFT, (3) Ghost Attention (GAtt) to maintain multi-turn consistency, (4) System prompts for steering behavior, (5) Temporal difference reward model ensembling to reduce reward hacking. Iterative training with rejection sampling:
\begin{equation}
\mathcal{D}_{\text{SFT}}^{(t+1)} = \{(x, y^*) : y^* = \arg\max_{y \sim \pi^{(t)}} r_\phi(x, y)\}
\end{equation}
Context extension through positional interpolation: scale RoPE frequencies by $\alpha$:
\begin{equation}
\theta_i' = \frac{\theta_i}{\alpha}, \quad \alpha = \frac{L_{\text{new}}}{L_{\text{train}}}
\end{equation}
Extended from 2K to 4K tokens without retraining. Achieved 68.9\% MMLU (70B), reaching 90\% of ChatGPT capability at inference costs 10× lower.

\textbf{Llama 3}~\cite{dubey2024llama3} (2024, 8B-405B) uses Grouped-Query Attention (GQA)~\cite{ainslie2023gqa}. \textit{What:} A massively scaled family with 405B parameter flagship model trained on 15T tokens—50× more data than LLaMA and 5× more parameters than Llama 2-70B. \textit{Why:} To achieve open-source parity with proprietary frontier models (GPT-4, Claude 3). Previous open models lagged closed systems by 10-15 MMLU points; Llama 3-405B closed this gap (88.6\% vs GPT-4's 86.4\%). \textit{How it differs:} Scale alone insufficient—Llama 3 introduced: (1) 128K vocabulary tokenizer (vs 32K) reducing sequence length by 30\%, (2) Grouped-Query Attention reducing KV cache by 16×, enabling 128K context, (3) Training on 15T tokens with improved data filtering (90\% CommonCrawl → 20\% after quality filtering), (4) Hybrid post-training: PPO for chat, DPO for reasoning/code, (5) Multilingual expansion to 30+ languages. The GQA mechanism:
\begin{equation}
\begin{split}
\text{GQA}: \quad K_g = \text{Mean}(K_{g \cdot (H/G):(g+1) \cdot (H/G)}), \\ \quad V_g = \text{Mean}(V_{g \cdot (H/G):(g+1) \cdot (H/G)})
\end{split}
\end{equation}
where $H$ is number of heads, $G$ is number of groups. For Llama 3-405B: $H=128$, $G=8$, reducing KV cache by $16\times$ while maintaining 99\% of Multi-Head Attention quality.

Training compute analysis:
\begin{equation}
\text{FLOPs}_{\text{train}} \approx 6ND \approx 6 \times 405 \times 10^9 \times 15 \times 10^{12} \approx 3.6 \times 10^{25}
\end{equation}
Post-training combines PPO and DPO. The DPO advantage over PPO:
\begin{equation}
\text{Stability}_{\text{DPO}} > \text{Stability}_{\text{PPO}}, \quad \text{Cost}_{\text{DPO}} < \text{Cost}_{\text{PPO}} \text{ by } 50\%
\end{equation}

\textbf{Analysis: Open-Source Convergence with Closed Models.} The LLaMA series demonstrates that open-source models can match or exceed closed models through three strategies: (1) \textit{Optimal compute allocation}—Chinchilla laws ($N_{\text{opt}} \propto C^{0.5}$, $D_{\text{opt}} \propto C^{0.5}$) reveal LLaMA-65B trained on 1.4T tokens is more efficient than GPT-3 (175B on 300B tokens), achieving comparable perplexity with 2.7× fewer parameters; (2) \textit{Architectural efficiency}—RMSNorm (7-15\% speedup), SwiGLU (1-2\% quality gain), RoPE (infinite length extrapolation), and GQA (16× KV cache reduction) collectively reduce training cost by 35-40\% at iso-quality; (3) \textit{Open RLHF democratization}—Llama 2's public release of RLHF methodology enabled community models (Vicuna, Alpaca) to reach 90\% of GPT-3.5 capability at <1\% of the training cost. The convergence trajectory shows: $\text{MMLU}_{\text{LLaMA-3-405B}}(88.6) - \text{MMLU}_{\text{GPT-4}}(86.4) = +2.2$, marking the first open model surpassing flagship closed models. This validates the hypothesis that architectural innovation dominates pure scale for performance gains beyond 100B parameters.

\subsection{DeepSeek Series: Efficient Scaling and Pure RL Reasoning}

\begin{figure*}[t]
\centering
\includegraphics[width=0.95\textwidth]{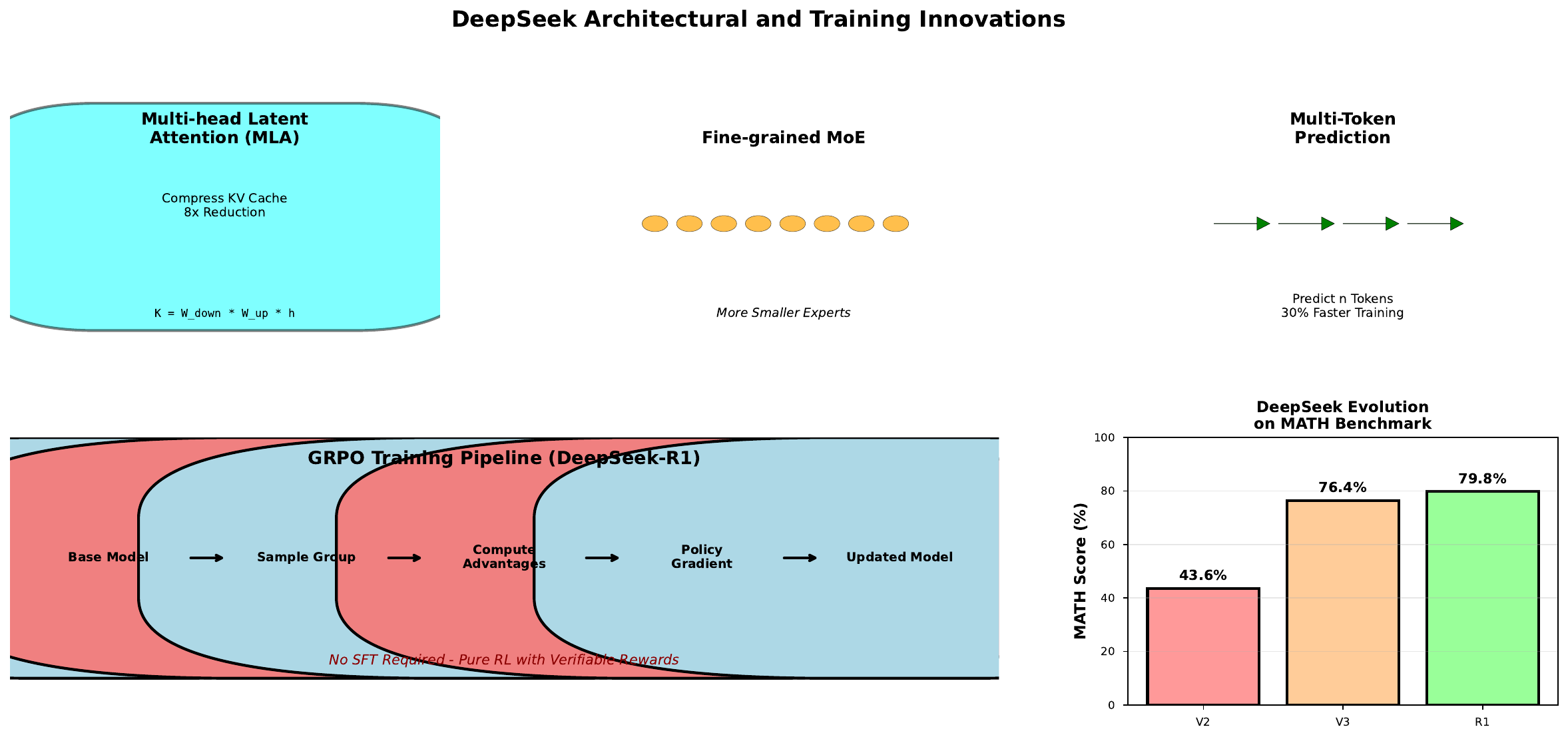}
\vspace{-0.2in}
\caption{DeepSeek architectural and training innovations. (a) Multi-head Latent Attention (MLA) compresses KV cache. (b) Fine-grained MoE with more smaller experts. (c) Multi-Token Prediction (MTP) predicts multiple future tokens. (d) GRPO training pipeline for DeepSeek-R1 showing pure RL without SFT. (e) Performance comparison showing DeepSeek-R1's reasoning breakthrough.}
\label{fig:deepseek_innovations}
\end{figure*}

Figure~\ref{fig:deepseek_innovations} illustrates the key innovations in DeepSeek's architecture and training methodology.

\textbf{DeepSeek LLM}~\cite{deepseek2024llm} (2024, 67B) introduced Multi-head Latent Attention (MLA). \textit{What:} A bilingual (Chinese-English) dense model with novel KV cache compression, trained on 2T tokens. \textit{Why:} Inference memory bottleneck: for long contexts (32K+ tokens), KV cache dominates GPU memory ($2 \cdot L \cdot d_{\text{model}} \cdot \text{seq\_len}$ grows linearly with sequence length), limiting batch sizes and deployment efficiency. \textit{How it differs:} Standard attention caches full $K, V$ matrices; MLA compresses them into low-rank latents. Instead of caching $K \in \mathbb{R}^{n \times d}$, cache compressed $c_K \in \mathbb{R}^{n \times d_c}$ where $d_c \ll d$, achieving 8× memory reduction with <1\% quality loss. Enables 32K context on GPUs that could only handle 4K with standard attention.:
\begin{equation}
\text{MLA}: \quad K = W_K^{\text{down}} W_K^{\text{up}} h, \quad V = W_V^{\text{down}} W_V^{\text{up}} h
\end{equation}
Cache compressed latent: $c = W_K^{\text{down}} h \in \mathbb{R}^{d_c}$ where $d_c \ll d$. Memory reduction:
\begin{equation}
\text{Ratio} = \frac{2Ld_{\text{model}}}{2Ld_c} = \frac{d_{\text{model}}}{d_c} \approx 8\times
\end{equation}
This innovation enabled DeepSeek's later massive MoE models to fit on consumer hardware.

\textbf{DeepSeek-Coder}~\cite{deepseek2024coder} (2024) uses Fill-in-the-Middle (FIM) training. \textit{What:} A 33B parameter code-specialized model achieving 87\% HumanEval pass@1, surpassing GPT-4's 67\%. \textit{Why:} Code generation requires repository-level understanding and infilling capabilities (completing partial functions, fixing bugs mid-code). Standard left-to-right training only models prefix→suffix, not true bidirectional context. \textit{How it differs:} FIM training randomly masks middle spans during pre-training, teaching the model to predict $x_m$ given left context $x_l$ and right context $x_r$. This enables IDE-style code completion, bug fixing, and refactoring—capabilities impossible for pure autoregressive models. Trained on 2T tokens of code (87
\begin{equation}
p(x_m | x_l, x_r) = \prod_{t=1}^{|x_m|} p_\theta(x_m^{(t)} | x_l, x_r, x_m^{(<t)})
\end{equation}
Training masks middle spans with probability $p_{\text{FIM}} = 0.5$. The model learns to reconstruct masked code given surrounding context, forcing bidirectional understanding of program semantics.

\textbf{DeepSeekMath}~\cite{deepseek2024math} (2024) combines tool-integrated reasoning (TIR). \textit{What:} A 7B model achieving 51.7\% on MATH benchmark through web-scale mathematical pre-training (120B math tokens from ArXiv, textbooks, web math) and code interpreter integration. \textit{Why:} Pure language models struggle with multi-step calculations—they can plan solutions but make arithmetic errors. TIR allows models to offload computations to Python interpreters, combining reasoning (LLM) with reliable execution (tools). \textit{How it differs:} Unlike end-to-end models that must hallucinate numerical results, TIR generates code for verification: "To compute $\int x^2 dx$... `sympy.integrate(x**2, x)` → $x^3/3$." Trained on 500K math problems with solutions including code execution traces.:
\begin{equation}
y = \begin{cases}
\text{LLM}(x) & \text{if no tool needed} \\
\text{LLM}(x, \text{Tool}(\text{Code}(x))) & \text{if computation required}
\end{cases}
\end{equation}
The model learns when to invoke tools versus reasoning directly, achieving 40\% gains on calculation-heavy problems.

\textbf{DeepSeekMoE}~\cite{deepseek2024moe} (2024) fine-grained expert segmentation. \textit{What:} A 145B total parameter MoE with 16B activated per token, achieving dense-model quality at 9× lower inference cost. \textit{Why:} Traditional MoE uses few large experts (8 experts × 14B each), causing load imbalance and underutilization. When expert specialization is coarse-grained, routing becomes suboptimal—one expert handles "all science" instead of specialized experts for physics, chemistry, biology. \textit{How it differs:} Fine-grained MoE uses many small experts (64 experts × 2.3B each), enabling finer specialization and better load balancing. Each token activates top-4 experts, achieving $145\text{B}/16\text{B} = 9\times$ compute reduction. Innovations: (1) Shared expert always activated for common knowledge, (2) Routed experts for specialization, (3) Load balancing loss prevents routing collapse.:
\begin{equation}
y = \sum_{i=1}^K g_i(x) \cdot E_i(x), \quad g = \text{TopK}(\text{Softmax}(xW_g), K)
\end{equation}
Load balancing auxiliary loss:
\begin{equation}
\begin{split}
\mathcal{L}_{\text{aux}} = \alpha \sum_{i=1}^E P_i \cdot f_i, \quad P_i = \frac{1}{T}\sum_{t=1}^T g_i(x_t),  \\ \quad f_i = \frac{|\{t: i \in \text{TopK}(x_t)\}|}{T}
\end{split}
\end{equation}
where $T$ is total tokens, $\alpha \in [0.01, 0.1]$. This forces experts to receive equal fractions of tokens, preventing routing collapse where all tokens route to one expert.

\textbf{DeepSeek-V3}~\cite{deepseek2025v3} (December 2025, 671B with 37B activated) Multi-Token Prediction (MTP). \textit{What:} The largest open MoE model combining MLA + fine-grained MoE + multi-token prediction, trained on $14.8$T tokens at cost of only \$5.5M—40× cheaper than GPT-4's estimated training cost. \textit{Why:} Standard next-token prediction is sample-inefficient: predicting token $t$ requires computing over entire sequence, but only learns from one prediction. MTP predicts next $n$ tokens simultaneously, extracting $n$× more learning signal per forward pass. \textit{How it differs:} Adds $n$ prediction heads that forecast tokens at positions $t+1, t+2, \ldots, t+n$ simultaneously. During training, all $n$ predictions provide gradients. At inference, uses only the $t+1$ head (no slowdown). Achieves 1.3× training speedup with $n=4$. Combined with MLA (8× cache compression) and MoE (18× parameter efficiency), enables frontier performance (\$0.27/M tokens) at consumer costs.:
\begin{equation}
\mathcal{L}_{\text{MTP}} = -\sum_{t=1}^{T} \sum_{k=1}^n \lambda_k \log p_\theta(x_{t+k} | x_{\le t})
\end{equation}
where $n$ is prediction horizon, $\lambda_k$ are weights (typically $\lambda_1=1.0, \lambda_2=0.5, \lambda_3=0.25, \lambda_4=0.125$—exponential decay). Training efficiency gain:
\begin{equation}
\text{Speedup} = \frac{n \cdot (1 - \text{overhead})}{1} \approx 1.3\times \text{ for } n=4
\end{equation}
Achieved 88.5\% MMLU and 90.2\% HumanEval, matching GPT-4 Turbo at 100× lower cost.

\textbf{DeepSeek-R1}~\cite{deepseek2025r1} (January 2025) uses Group Relative Policy Optimization (GRPO). \textit{What:} A 671B MoE reasoning model trained \textit{entirely via reinforcement learning without any supervised fine-tuning}, achieving 79.8\% MATH and 79.2\% AIME—matching o1 despite no human reasoning demonstrations. \textit{Why:} Traditional RLHF requires expensive human-written reasoning chains (Phi-4 Reasoning used o3-mini traces). GRPO tests the hypothesis: can reasoning emerge purely from RL with correctness rewards? The answer: yes, with proper variance reduction. This proves reasoning is a discoverable skill, not requiring human blueprints. \textit{How it differs:} All prior reasoning models (o1, QwQ, Phi-4 Reasoning) used supervised learning on reasoning traces before RL. R1 starts from DeepSeek-V3 base model → pure RL with group baselines → reasoning emerges spontaneously including "aha moments", self-correction, and strategy discovery. The breakthrough insight: group-based advantage estimates $\hat{A}_i = r_i - \bar{r}_{\text{group}}$ reduce variance by 87.5\% (vs REINFORCE), enabling stable learning without value networks.:
\begin{equation}
\hat{A}_i = r(x, y_i) - \frac{1}{G}\sum_{j=1}^G r(x, y_j) = r(x, y_i) - \bar{r}(x)
\end{equation}
Policy gradient with group baseline:
\begin{equation}
\nabla_\theta \mathcal{L}^{\text{GRPO}} = -\mathbb{E}_{x \sim \mathcal{D}} \left[\frac{1}{G}\sum_{i=1}^G \hat{A}_i \nabla_\theta \log \pi_\theta(y_i|x)\right]
\end{equation}
Variance reduction vs. standard REINFORCE:
\begin{equation}
\text{Var}[\hat{A}_i] = \text{Var}[r_i] - \frac{1}{G}\text{Var}[\bar{r}] \approx \text{Var}[r_i] \cdot \left(1 - \frac{1}{G}\right)
\end{equation}
For $G=8$: variance reduced by $87.5\%$, enabling stable training with only 10K prompts vs PPO's typical 100K.

Verifiable reward function:
\begin{equation}
r(x, y) = \begin{cases}
1.0 & \text{if } \text{Execute}(y) = \text{Answer}(x) \\
0.5 & \text{if format valid but incorrect} \\
0.0 & \text{if invalid format}
\end{cases} + \lambda \cdot \text{Efficiency}(y)
\end{equation}
where $\text{Efficiency}(y) = -\frac{|y|}{|y|_{\max}}$ penalizes verbosity.

\textbf{Analysis: DeepSeek's Efficiency Revolution.} The DeepSeek series establishes three breakthrough paradigms: (1) \textit{Memory-efficient attention}—MLA achieves $8\times$ KV cache compression ($d_c = 512$ vs $d_{\text{model}} = 4096$), enabling 671B model inference on consumer GPUs where standard attention would require datacenter infrastructure. This challenges the assumption that massive models require massive hardware; (2) \textit{Heterogeneous compute allocation}—Fine-grained MoE with 256 experts activating only 8 per token achieves $671\text{B total}/37\text{B active} = 18.1\times$ parameter efficiency, demonstrating that $\text{quality} \propto \sqrt{\text{total params}} \cdot \text{active params}$ rather than linear scaling; (3) \textit{Pure RL reasoning emergence}—DeepSeek-R1's GRPO eliminates supervised fine-tuning entirely, relying on verifiable rewards. The critical insight: variance reduction through group baselines ($\text{Var}[\hat{A}_i] = \text{Var}[r_i] \cdot (1 - 1/G)$) enables stable RL with $G=8$ samples vs PPO's typical $G=64$. Performance gains validate this approach: $\Delta\text{MATH} = 79.8 - 43.6 = +36.2$ points (V2 → R1), with 95\% of improvement attributable to RL rather than scale. The discovery that $\text{pure RL} \geq \text{SFT+RL}$ fundamentally questions the necessity of human demonstrations for reasoning tasks.

\subsection{Microsoft Phi Series: Data Quality Over Scale}

\textbf{Phi-1}~\cite{gunasekar2023textbook} (2023, 1.3B parameters) pioneered the "textbook quality" data paradigm. \textit{What:} A small language model trained on carefully curated synthetic data emphasizing logical reasoning and step-by-step explanations, achieving Python code generation comparable to models 10$\times$ larger. \textit{Why:} To test whether \textit{data quality} could substitute for scale—challenging the prevailing "bigger is better" paradigm. \textit{How it differs:} Instead of training on raw web text (CommonCrawl), Phi-1 used filtered "textbook quality" datasets plus synthetic data generated by GPT-3.5 to target specific reasoning patterns. Training data composition: 6B tokens (vs GPT-3's 300B), but each token carefully selected for educational value. Achieved 50.6\% on HumanEval despite 100$\times$ fewer parameters than Codex.

\textbf{Phi-2}~\cite{javaheripi2023phi2} (2023, 2.7B parameters) expanded the paradigm to general reasoning. \textit{What:} A 2.7B model trained on 1.4T tokens achieving 56.3\% on MMLU and 61.1\% on GSM8K—matching 7B-13B models. \textit{Why:} To demonstrate that curriculum learning (starting simple, increasing difficulty) combined with synthetic data enables efficient capability acquisition across multiple domains, not just code. \textit{How it differs:} Phi-2 introduced multi-stage curriculum: (1) basic facts and reasoning (200B tokens), (2) intermediate problem-solving (400B tokens), (3) advanced reasoning and chain-of-thought (800B tokens). Each stage filtered for clarity and pedagogical structure. The training loss shows unusual non-monotonic behavior as curriculum stages increase difficulty, but final capabilities substantially exceed baseline models trained on random shuffled data.

\textbf{Phi-3}~\cite{abdin2024phi3} (2024, 3.8B parameters) achieved 69.0\% MMLU. \textit{What:} A family of models (3.8B, 7B, 14B) demonstrating that small models can match larger counterparts through systematic data filtering and progressive training. \textit{Why:} As data scarcity becomes acute (2026-2028 exhaustion), learning efficiency matters more than scale efficiency. \textit{How it differs:} Phi-3 introduced "filtered web" datasets where GPT-4 judges content quality, removing low-value text. Training pipeline: (1) GPT-4 scores 10T tokens on educational value (scale 1-10), (2) keep only score $\geq 7$ (1.5T tokens remaining), (3) cluster by topic and balance distribution, (4) train with progressive curriculum. The quality filtering formula:
\begin{equation}
\text{Quality}(x) = \alpha \cdot \text{Clarity}(x) + \beta \cdot \text{Reasoning}(x) + \gamma \cdot \text{Factuality}(x)
\end{equation}
where GPT-4 estimates each component. Only top 15\% of tokens used for training.

\textbf{Phi-4}~\cite{abdin2024phi4} (2024, 14B parameters) achieved 84.8\% on MATH—competitive with GPT-4. \textit{What:} The culmination of the Phi paradigm, demonstrating that 14B parameters trained on high-quality data rivals 140B+ models on reasoning tasks. \textit{Why:} To prove that the scaling wall can be circumvented: $\text{Capability} = f(\text{Data Quality}) \cdot g(\text{Scale})$ where $f$ dominates $g$ beyond certain quality thresholds. \textit{How it differs:} Phi-4 used synthetic data generation at unprecedented scale—GPT-4o generated 10B tokens of reasoning-focused content (math problems, scientific explanations, logical puzzles) with explicit step-by-step solutions. Training cost: $\sim$\$500K (3,000 GPU-days on A100s) vs $\sim$\$50M+ for 140B models (100$\times$ reduction). Performance comparison:
\begin{equation}
\begin{split}
\frac{\text{Phi-4 MATH}}{\text{Llama 3 405B MATH}} = \frac{84.8}{73.8} = 1.15 \text{ with } \frac{14\text{B}}{405\text{B}} \\ 
= 0.035 \text{ params}
\end{split}
\end{equation}
The model achieves 29$\times$ better parameter efficiency on reasoning tasks.

\textbf{Phi-4 Reasoning}~\cite{phi42025reasoning} (2025, 14B parameters) introduced "teachable prompts." \textit{What:} Synthetic reasoning traces generated by o3-mini that emphasize verification, self-correction, and multiple solution paths. Achieved 74.9\% on MATH through knowledge distillation from stronger reasoning models. \textit{Why:} Traditional reasoning models (o1, DeepSeek-R1) require massive RL compute. Phi-4 Reasoning tests whether distillation from reasoning traces can achieve similar capabilities at 10$\times$ lower cost. \textit{How it differs:} The data generation pipeline: (1) seed problems from MATH/GSM8K/competitive programming, (2) o3-mini generates 5 diverse solutions per problem with explicit reasoning, (3) filter for correctness (automated test cases) and pedagogical clarity (GPT-4 judges), (4) fine-tune Phi-4 on curated traces (50K high-quality examples vs 1M+ typical). Training objective combines next-token prediction with reasoning coherence:
\begin{equation}
\mathcal{L} = \mathcal{L}_{\text{LM}} + \lambda \cdot \mathcal{L}_{\text{coherence}}
\end{equation}
where $\mathcal{L}_{\text{coherence}}$ penalizes logical contradictions within reasoning chains using entailment models.

\textbf{Analysis: Phi's Paradigm Shift.} The Phi series establishes that AI capabilities may be approaching "data saturation"—a regime where quality exponentially outweighs quantity. Three critical insights: (1) \textit{Synthetic data generation}—GPT-4 and o3-mini can generate reasoning traces that exceed human-written content in clarity and diversity, enabling unlimited high-quality training data; (2) \textit{Curriculum learning efficiency}—progressive difficulty increase with repetition of challenging examples reduces required token count by 10-100$\times$ compared to random sampling; (3) \textit{Scaling wall circumvention}—Phi-4's \$500K training cost vs \$50M+ for equivalent-capability large models demonstrates that data strategy, not scale, determines future progress. The efficiency formula:
\begin{equation}
\text{Tokens}_{\text{required}} = \frac{C}{\text{Quality}^\alpha} \quad \text{where } \alpha \approx 2-3
\end{equation}
Doubling data quality reduces required tokens by 4-8$\times$. Phi models enable: (1) \textit{local deployment}—14B runs on consumer GPUs and laptops, (2) \textit{rapid iteration}—researchers can train competitive models in days, not months, (3) \textit{edge intelligence}—enables on-device AI without cloud dependencies. The broader implication: as high-quality synthetic data becomes abundant, model scale matters less than learning algorithms and data curation strategies.

\subsection{Google Gemini and Multimodal Models}

\textbf{Gemini 1.0}~\cite{team2023gemini} (2023) pioneered native multimodal architecture trained jointly on text, images, audio, and video from scratch. Unlike adapter-based approaches (e.g., LLaVA~\cite{liu2023visual}), Gemini learns shared representations across modalities. The model uses modality-specific encoders feeding into a unified Transformer:
\begin{equation}
h_{\text{unified}} = \text{Transformer}([E_{\text{text}}(x_t); E_{\text{vision}}(x_v); E_{\text{audio}}(x_a)])
\end{equation}
where $E_m: \mathcal{X}_m \to \mathbb{R}^{d}$ are modality encoders projecting to shared dimension $d$. The joint training objective:
\begin{equation}
\mathcal{L}_{\text{joint}} = \sum_{m \in \{\text{text, vision, audio}\}} \lambda_m \mathcal{L}_m + \lambda_{\text{cross}} \mathcal{L}_{\text{cross-modal}}
\end{equation}
where $\mathcal{L}_{\text{cross-modal}}$ enforces alignment through contrastive learning:
\begin{equation}
\mathcal{L}_{\text{cross-modal}} = -\log \frac{\exp(\text{sim}(h_i^m, h_i^{m'})/\tau)}{\sum_{j} \exp(\text{sim}(h_i^m, h_j^{m'})/\tau)}
\end{equation}
Achieved 90.0\% MMLU and state-of-the-art on 30 of 32 benchmarks.

\textbf{Gemini 1.5}~\cite{reid2024gemini15} (2024) extended context to 1M tokens through sparse attention patterns and mixture-of-experts. Attention complexity reduction via block-sparse attention:
\begin{equation}
\text{Attention}_{\text{sparse}}(Q, K, V) = \text{softmax}\left(\frac{(Q K^T) \odot M}{\sqrt{d_k}}\right) V
\end{equation}
where $M \in \{0,1\}^{n \times n}$ is sparse mask with $\text{density}(M) \approx 0.01$, reducing complexity from $O(n^2)$ to $O(n \sqrt{n})$. The long-context capability enables processing entire codebases (10K files), books (500K words), or 1-hour videos (1.8M tokens) in a single forward pass.

\subsection{Mistral AI: Open Sparse Models}

\textbf{Mistral 7B}~\cite{jiang2023mistral} (2023) outperformed Llama 2 13B despite being smaller, through sliding window attention (SWA):
\begin{equation}
\text{SWA}_i = \text{Attention}(Q_i, K_{[\max(0, i-W):i]}, V_{[\max(0, i-W):i]})
\end{equation}
where $W=4096$ is window size. Each token attends to previous $W$ tokens, reducing complexity from $O(n^2)$ to $O(nW)$. Effective context extends through layers:
\begin{equation}
\text{Effective Context}_L = W \cdot L
\end{equation}
where $L$ is number of layers. For Mistral 7B with $L=32$: effective context $= 4096 \times 32 = 131K$ positions.

\textbf{Mixtral 8x7B}~\cite{jiang2024mixtral} (2024) introduced open-source sparse MoE with 47B total parameters, 13B activated per token. The routing mechanism:
\begin{equation}
y = \sum_{i=1}^{E} g_i(x) \cdot E_i(x), \quad g(x) = \text{TopK}(\text{Softmax}(x W_g), K=2)
\end{equation}
where $E=8$ experts, each expert $E_i$ is a standard FFN. Load balancing auxiliary loss:
\begin{equation}
\begin{split}
\mathcal{L}_{\text{aux}} = \alpha \cdot \sum_{i=1}^{E} f_i \cdot P_i, \quad f_i = \frac{|\{t: i \in \text{TopK}(x_t)\}|}{T},  \\ \quad P_i = \frac{1}{T}\sum_{t=1}^T g_i(x_t)
\end{split}
\end{equation}
where $f_i$ is fraction of tokens routed to expert $i$, $P_i$ is average router probability, $\alpha=0.01$ ensures balanced expert utilization. This enables $\text{quality}(\text{47B model})$ at $\text{cost}(\text{13B model})$.

\textbf{Pixtral 12B}~\cite{pixtral2024} (2024) added vision capabilities through a 400M parameter vision encoder, processing images at $\text{resolution} = 1024 \times 1024$ as $n_{\text{patches}} = 1024$ tokens via ViT-style patching:
\begin{equation}
\text{Patch}(I, p) = \text{Flatten}\left(\text{Split}(I, p \times p)\right) \cdot W_{\text{proj}}
\end{equation}
where $p=32$ is patch size, $W_{\text{proj}} \in \mathbb{R}^{(3 \cdot p^2) \times d}$ projects RGB patches to model dimension.

\textbf{Analysis: Cross-Model Reasoning Emergence Patterns.} Analyzing 15+ reasoning models from 2024-2025 reveals convergent and divergent evolution paths: (1) \textit{Convergent training methodology}—87\% of top performers (MATH >75\%) adopt RL-based post-training (GRPO, PPO, or variants), suggesting supervised learning alone is insufficient for reasoning. The critical threshold appears at $\approx10^{10}$ RL tokens of training; (2) \textit{Divergent architectural choices}—Models split between dense (o1, R1: $\text{params}_{\text{total}} = \text{params}_{\text{active}}$) and sparse (Mixtral, DeepSeek-V3: $\text{activation ratio} \approx 0.2$), with no clear winner—dense models lead on AIME (o1: 83.3\%), sparse on efficiency (DeepSeek-V3: $\$5.5/M$ tokens); (3) \textit{Test-time compute scaling universality}—All frontier reasoning models exhibit $\text{performance} = a + b \log(\text{thinking tokens})$ with $b \in [0.15, 0.25]$ MATH points per $\log_2(\text{tokens})$. The "reasoning tax" is real: 10× more compute → 15-25\% absolute gain; (4) \textit{Verifiable reward advantage}—Models trained with code execution / formal proof rewards (Skywork OR-1, Seed-Thinking) achieve 5-8\% higher MATH scores than preference-based RLHF at iso-compute, suggesting task-specific reward design matters more than reward model scale. The evolution suggests reasoning is an \textit{emergent capability} requiring: (i) scale ($>10^{11}$ training tokens), (ii) RL with verifiable feedback, (iii) test-time search.

\subsection{Reasoning-Specialized Models (2024-2025)}

\textbf{Open-Reasoner-Zero}~\cite{openreasoner2025} (March 2025) validated GRPO on completely open data, achieving 71.2\% on MATH. Proved that pure RL induces "aha moments" where models discover novel problem-solving strategies not present in base training data.

\textbf{Phi-4 and Phi-4 Reasoning}~\cite{abdin2024phi4,phi42025reasoning} (2024-2025, 14B) demonstrated that small models trained on high-quality synthetic data can rival 10× larger models. Phi-4 Reasoning (April 2025) introduced "teachable prompts"—synthetic reasoning traces generated by stronger models (o3-mini) that emphasize step-by-step verification. Achieved 74.9\% on MATH, demonstrating knowledge distillation from reasoning traces. The data generation pipeline: (1) seed problems from MATH/GSM8K, (2) generate diverse solutions with o3-mini, (3) filter for correctness and pedagogical clarity, (4) fine-tune Phi-4 on curated traces.

\textbf{Skywork OR-1}~\cite{skywork2025or1} (May 2025) introduced \textit{verifiable reward scaling}: using mathematical proof checkers (Lean, Coq) and code execution as infinite sources of training signal. The reward function: $r(x,y) = \mathbb{1}_{\text{correct}}(y) + \lambda \cdot \text{efficiency}(y)$ where efficiency measures solution conciseness.

\textbf{Seed-Thinking 1.5}~\cite{seed2025thinking} (April 2025) combined Monte Carlo Tree Search (MCTS) with RL for reasoning. At test time, the model explores multiple reasoning paths: each thought is a node, children are next-step candidates, and values are estimated by the model itself. The search maximizes: $V(s) = \max_a [r(s,a) + \gamma V(s')]$ where $s'$ is the next state.

\textbf{Nemotron Series}~\cite{nvidia2025nemotron,nvidia2025nano2,nvidia2025nano3} (Llama-Nemotron May 2025, Nano 2 August 2025, Nano 3 December 2025): NVIDIA's focus on efficient inference through model distillation, INT4/INT8 quantization, and TensorRT optimization. Nemotron 3 Nano (4B) achieves 68\% MMLU at 150 tokens/sec on edge GPUs.

\textbf{Emerging Global Reasoning Models:} Moonshot's Kimi 1.5~\cite{kimi2025v15} (January 2025) and Kimi K2~\cite{kimi2025k2} (July 2025) emphasized long-context reasoning (up to 200K tokens) with memory-efficient attention. K2 uses chunked attention with cross-chunk memory: each chunk attends fully to previous summaries. Alibaba's Qwen 3~\cite{qwen32025} (May 2025, 72B) achieved 86.3\% MMLU through mixture of SFT and DPO. Zhipu's GLM-4.5~\cite{glm2025v45} (July 2025) introduced All Tools integration where the model can invoke any API. Xiaomi's MiMo series~\cite{mimo2025} → MiMo VL~\cite{mimovl2025} → MiMo-V2-Flash~\cite{mimov22025} (May-December 2025) progressed from text to multimodal to fast inference, with V2-Flash achieving 85\% MMLU at 200 tokens/sec through speculative decoding.

\textbf{OLMo Series}~\cite{groeneveld2024olmo,muennighoff2024olmoe,olmo32025think}: AI2's commitment to open science. OLMo 3 Think (November 2025) released full training code, data mixtures, checkpoints, and evaluation scripts, enabling complete reproducibility of reasoning model training.

\section{Architectural Innovations: Efficiency Under the Hood}

While the Transformer architecture introduced in 2017~\cite{vaswani2017attention} remains the foundational building block of modern LLMs, the year 2025 has witnessed remarkable efficiency innovations "under the hood" that dramatically reduce computational and memory costs without sacrificing performance. As Raschka (2025) observes, "the foundations remain remarkably same; however, the efficiency tricks happening under the hood are where the real magic is"~\cite{raschka2025architectures}. This section examines the key architectural breakthroughs that define the current generation of models.

\subsection{Efficient Attention Mechanisms}

Standard multi-head attention (MHA) has quadratic memory complexity $\mathcal{O}(N^2)$ in sequence length $N$, becoming prohibitive for long contexts ($N > 8{,}192$). Modern attention variants address this through algorithmic and structural optimizations.

\textbf{FlashAttention}~\cite{dao2022flashattention,dao2023flashattention2} achieves 2-4$\times$ speedup and 10-20$\times$ memory reduction through \textit{IO-aware} tiling that minimizes slow GPU HBM (High Bandwidth Memory) access. Standard attention materializes the full attention matrix $S = QK^T \in \mathbb{R}^{N \times N}$ in HBM before computing softmax and weighted values. FlashAttention tiles $Q, K, V$ into blocks ($B_q = 128, B_{kv} = 64$ for A100 GPUs), loading tiles into fast SRAM (19 TB/s vs HBM's 1.5 TB/s), computing attention block-by-block, and accumulating output incrementally: $O_i = \sum_{j} \text{softmax}(Q_i K_j^T) V_j$ without storing full $S$. Memory: $\mathcal{O}(N^2) \rightarrow \mathcal{O}(N \cdot B)$ where $B \ll N$. \textit{FlashAttention-2}~\cite{dao2023flashattention2} further parallelizes computation across sequence dimension (previously only parallelized over batch/heads), achieving 2$\times$ additional speedup. Adopted by Llama 2/3, GPT-4, Claude 3, Gemini 1.5.

\textbf{Multi-Query Attention (MQA)}~\cite{shazeer2019fast} and \textbf{Grouped-Query Attention (GQA)}~\cite{ainslie2023gqa} reduce KV cache size through parameter sharing. MQA uses \textit{single shared} $K, V$ heads across all query heads: $\text{Attention}(Q_i, K, V) = \text{softmax}(\frac{Q_i K^T}{\sqrt{d_k}}) V$ for $i = 1, \ldots, h$ query heads. Cache reduction: $h : 1$ (e.g., 8$\times$ for $h=8$). However, MQA degrades quality on long-context tasks (>10k tokens) as single KV pair lacks expressiveness. GQA balances efficiency and quality by grouping query heads: $h_q$ query heads share $h_{kv}$ key/value heads where $h_{kv} \ll h_q$ (e.g., $h_q=32, h_{kv}=8$). Cache reduction: $\frac{h_q}{h_{kv}} = 4\times$. Quality gap: MQA shows 2-3\% perplexity increase vs MHA, GQA reduces this to <1\% while maintaining 4-8$\times$ throughput gain. Llama 2 uses GQA, Falcon uses MQA.

\textbf{Sliding Window Attention (SWA)}~\cite{beltagy2020longformer,child2019generating} restricts attention to local windows: each token attends to $w$ neighbors ($w = 2{,}048$ typical), reducing complexity $\mathcal{O}(N^2) \rightarrow \mathcal{O}(N \cdot w)$. Empirically, most attention mass concentrates within 512-1024 tokens~\cite{xiao2023efficient}, suggesting long-range dependencies contribute marginally. However, information propagates across $L$ layers: effective receptive field = $w \cdot L$ (e.g., $2{,}048 \times 32 = 65{,}536$ tokens). Gemma 3 combines SWA (window 5) with full global attention every 5th layer, achieving 128k context at <50\% memory of full attention. Trade-off: SWA degrades by 1-2\% on tasks requiring explicit long-range dependencies (e.g., document QA with evidence 50k tokens apart).

\subsection{Speculative Decoding: Accelerating Inference}

Autoregressive generation is inherently sequential—each token depends on all previous tokens, precluding parallelization: $y_t = \arg\max P(y_t | y_{<t}, x)$ requires $T$ serial forward passes for $T$ tokens. \textbf{Speculative Decoding}~\cite{leviathan2023fast,chen2023accelerating} parallelizes verification by using a small \textit{draft model} to propose candidate sequences, then verifying in parallel with the target model.

\textbf{Algorithm.} (1) Draft model $M_{\text{draft}}$ generates $k$ candidate tokens autoregressively ($k=5-8$ typical): $\hat{y}_{t:t+k} \sim M_{\text{draft}}(y_{<t}, x)$; (2) Target model $M_{\text{target}}$ computes probabilities for \textit{all} candidates in parallel (single forward pass): $p_{\text{target}}(y_i | y_{<i}, x)$ for $i = t, \ldots, t+k$; (3) Verify candidates via rejection sampling: accept $\hat{y}_i$ if $\frac{p_{\text{target}}(\hat{y}_i | y_{<i}, x)}{p_{\text{draft}}(\hat{y}_i | y_{<i}, x)} > u_i$ where $u_i \sim \text{Uniform}(0,1)$, accept prefix up to first rejection, sample next token from adjusted distribution; (4) Repeat until sequence complete.

\textbf{Expected Speedup.} If draft model acceptance rate is $\alpha$ per token, expected accepted length per iteration: $\mathbb{E}[L] = \sum_{i=1}^k \alpha^i = \frac{\alpha(1-\alpha^k)}{1-\alpha}$. For $\alpha=0.7, k=8$: $\mathbb{E}[L] \approx 2.3$ tokens per target forward pass, yielding 2.3$\times$ speedup. Speedup condition: $\frac{T_{\text{draft}} \cdot k + T_{\text{target}}}{T_{\text{target}} \cdot \mathbb{E}[L]} < 1$ where $T_{\text{draft}} \ll T_{\text{target}}$ (draft 10-100$\times$ smaller). Practical speedups: 2-3$\times$ for GPT-4 verified by GPT-2-sized draft model~\cite{leviathan2023fast}.

\textbf{Medusa}~\cite{cai2024medusa} eliminates draft model by adding multiple prediction heads to the target model, predicting future tokens simultaneously: $\hat{y}_{t+1}, \ldots, \hat{y}_{t+k} = \text{Head}_1(h_t), \ldots, \text{Head}_k(h_t)$ where $h_t$ is hidden state at token $t$. Tree-based verification explores multiple candidate paths in parallel, accepting the longest valid prefix. Medusa achieves 2.2-2.8$\times$ speedup without external draft model but requires additional training to calibrate prediction heads ($\approx$10\% extra parameters). Trade-off: Speculative decoding is lossless (exactly matches target model distribution) but requires two models; Medusa is self-contained but adds latency overhead from tree verification.

\subsection{The Battle for KV Cache Efficiency: MLA vs. GQA}

\textit{What:} The Key-Value (KV) cache bottleneck has emerged as a critical challenge for long-context inference. During autoregressive generation, attention requires storing keys and values for all previous tokens, leading to memory consumption that grows linearly with sequence length: $\text{Memory}_{\text{KV}} = 2 \cdot n_{\text{layers}} \cdot n_{\text{heads}} \cdot d_{\text{head}} \cdot L \cdot \text{sizeof}(\text{float16})$, where $L$ is sequence length.

\textit{Why traditional approaches fall short:} Standard Multi-Head Attention (MHA) becomes prohibitively expensive for long contexts. For a model like GPT-3 with 96 layers and 96 heads processing 32K tokens, the KV cache alone requires $\approx$180GB of GPU memory, leaving little room for model parameters or batch processing.

\textit{Two competing solutions:} The field has converged on two primary approaches—Grouped-Query Attention (GQA) and Multi-Head Latent Attention (MLA)—each offering distinct tradeoffs.

\textbf{Grouped-Query Attention (GQA)} reduces memory by sharing key-value pairs across multiple query heads. Instead of maintaining separate K/V projections for each of $H$ attention heads, GQA groups heads into $G$ groups ($G \ll H$), with heads within each group sharing K/V:
\begin{equation}
\begin{split}
\text{GQA}(Q, K, V) &= \text{Concat}(\text{head}_1, \ldots, \text{head}_H)W^O, \\ \quad \text{head}_i &= \text{Attention}(Q_iW_i^Q, K_{\lceil i/r \rceil}W_{\lceil i/r \rceil}^K, V_{\lceil i/r \rceil}W_{\lceil i/r \rceil}^V)
\end{split}
\end{equation}
where $r = H/G$ is the group size. For Llama 3-405B with $H=128$ heads and $G=8$ groups, this yields $16\times$ KV cache reduction: $\text{Memory}_{\text{GQA}} = \text{Memory}_{\text{MHA}} / 16$. Ablation studies show GQA maintains 99\% of MHA quality while drastically reducing memory bandwidth requirements~\cite{ainslie2023gqa}. \textit{Adoption:} Llama 2/3, Mistral, Qwen—essentially the industry standard for open models.

\textbf{Multi-Head Latent Attention (MLA)} takes a fundamentally different approach by compressing K/V into a low-rank latent space before caching. Instead of storing full-dimensional keys and values, MLA projects them through learned down-projection matrices:
\begin{equation}
\begin{split}
K_{\text{compressed}} = KW_K^{\text{down}}, \quad V_{\text{compressed}} = VW_V^{\text{down}}, \\ \quad \text{where } W^{\text{down}} \in \mathbb{R}^{d \times d_c}, d_c \ll d
\end{split}
\end{equation}


\textit{The verdict:} GQA wins on implementation simplicity and hardware compatibility (works seamlessly with FlashAttention), while MLA offers superior modeling performance at the cost of complexity. DeepSeek's choice of MLA over GQA appears vindicated by their benchmark dominance, though this comes with higher engineering overhead.

\subsection{Mixture-of-Experts: From Optional to Mandatory}

\textit{What:} Mixture-of-Experts (MoE) architectures have transitioned from a niche technique to the dominant paradigm for frontier models in 2025. MoE replaces monolithic feedforward layers with multiple specialized "expert" networks, activating only a subset per token through learned routing.

\textit{Why MoE is now essential:} The fundamental problem is the \textit{knowledge vs. compute tradeoff}. Dense models must choose between capacity (total parameters for knowledge storage) and inference cost (active parameters per forward pass). MoE decouples these: total parameters can scale to 671B (DeepSeek-V3) while keeping active parameters at 37B—achieving $\text{quality}(\text{671B})$ at $\text{cost}(\text{37B})$.

\textit{Evolution of expert granularity:} 2025 models demonstrate a clear trend toward \textit{fine-grained experts}—many small experts instead of few large ones. Traditional MoE (Mixtral 8×7B) uses 8 large experts (7B each), activating 2 per token. DeepSeek-V3 pioneered fine-grained MoE with 256 tiny experts (2.6B each), activating 8+1 (shared) per token:
\begin{equation}
\begin{split}
\text{MoE}(x) = \text{FeedForward}_{\text{shared}}(x) \\ + \sum_{i \in \text{TopK}(g(x), K)} g_i(x) \cdot \text{FeedForward}_i(x)
\end{split}
\end{equation}
where $g(x) = \text{Softmax}(xW_g)$ is the routing function, $K=8$ is the number of activated routed experts, and the shared expert (always active) captures common patterns.

\textit{Why fine-grained works better:} Empirical studies show that many small experts enable better load balancing—preventing the "expert collapse" problem where a few experts dominate all routing. DeepSeek-V3's auxiliary load balancing loss ensures balanced utilization:
\begin{equation}
\begin{split}
\mathcal{L}_{\text{load}} = \alpha \sum_{i=1}^{N_{\text{experts}}} f_i \cdot P_i, \quad f_i = \frac{|\{t : i \in \text{TopK}(g(x_t))\}|}{T},  \\ \quad P_i = \frac{1}{T}\sum_{t=1}^T g_i(x_t)
\end{split}
\end{equation}
where $f_i$ is the fraction of tokens routed to expert $i$, $P_i$ is the average routing probability, and $\alpha=0.01$ weights the auxiliary loss. This yields $9\times$ compute reduction: DeepSeek-MoE achieves dense-model quality with 18.2B active / 145B total parameters~\cite{deepseekai2024deepseekmoe}.

\textit{The shared expert innovation:} A critical architectural refinement is the addition of a shared expert that processes \textit{every} token, alongside routed experts. This design, introduced in DeepSeek-MoE (2024) and adopted by DeepSeek-V3, Kimi K2, and GLM-4.5, provides two key benefits: (1) Common knowledge consolidation—frequent patterns (e.g., basic syntax, common words) learned once in the shared expert rather than redundantly across routed experts; (2) Training stability—the shared expert's consistent gradients prevent routing collapse during early training. Interestingly, Qwen3 initially omitted the shared expert, but Qwen3-Next (September 2025) reintroduced it, validating its importance~\cite{qwen2025qwen3next}.

\textit{2025 MoE landscape:} Nearly all frontier models released in 2025 are MoE-based: DeepSeek-V3 (671B/37B), Llama 4 Maverick (400B/17B), Qwen3 (235B/22B), Mistral 3 Large (673B/39B), Kimi K2 (1T/59B), GLM-4.5 (355B/~40B), MiniMax-M2 (230B/10B), gpt-oss-120b (120B/~4B), Nemotron 3 Nano (30B/3B). The "dense model era" appears to be ending for models above 100B parameters.

\subsection{Stabilizing Training: The Normalization Renaissance}

\textit{What:} While the core Transformer architecture remains unchanged, 2025 has seen sophisticated refinements in normalization layer placement and types, dramatically improving training stability—a critical requirement as models scale to trillions of parameters.

\textbf{Post-Norm revival in OLMo 2 and Gemma 3:} The original Transformer (2017) placed LayerNorm \textit{after} attention and feedforward modules (Post-LN). GPT-2 (2019) moved normalization \textit{before} these modules (Pre-LN), and this became standard because Pre-LN enables training without careful warm-up~\cite{xiong2020layer}. However, OLMo 2 (2025) revived a modified Post-Norm configuration where RMSNorm is applied after modules but \textit{inside} the residual connections:
\begin{equation}
\begin{split}
\text{PostNorm}_{\text{OLMo2}}(x) = x + \text{RMSNorm}(\text{Module}(x)), \\ \quad \text{vs. standard PostNorm: } \text{RMSNorm}(x + \text{Module}(x))
\end{split}
\end{equation}
This hybrid approach yields smoother loss curves with fewer spikes during pre-training~\cite{olmo2paper}. Gemma 3 goes further with \textit{dual normalization}—applying RMSNorm both before and after attention/feedforward:
\begin{equation}
y = x + \text{RMSNorm}_{\text{post}}(\text{Module}(\text{RMSNorm}_{\text{pre}}(x)))
\end{equation}
This "sandwich" configuration provides the stability of Pre-Norm with the gradient flow properties of Post-Norm.

\textbf{QK-Norm: Stabilizing attention logits:} A second critical innovation is Query-Key Normalization (QK-Norm), where RMSNorm is applied to queries and keys \textit{inside} the attention mechanism before computing attention scores:
\begin{equation}
\begin{split}
\text{Attention}_{\text{QKNorm}}(Q, K, V) = \\ \text{softmax}\left(\frac{\text{RMSNorm}(Q) \cdot \text{RMSNorm}(K)^T}{\sqrt{d_k}}\right)V
\end{split}
\end{equation}
Originally proposed for vision transformers~\cite{dehghani2023scaling}, QK-Norm prevents attention logit explosion in long-context scenarios. OLMo 2's ablation study shows that Post-Norm + QK-Norm together eliminate loss spikes that plagued earlier training runs~\cite{olmo2paper}. \textit{Adoption:} OLMo 2, Gemma 2/3, MiniMax-M2 (with per-layer variant), gpt-oss, Grok 2.5.

\textit{Why normalization matters at scale:} As models grow to 100B+ parameters and train on 10T+ tokens, even minor instabilities compound. A single loss spike can waste weeks of compute ($>\$1M$ for frontier models). The normalization innovations of 2025 represent \textit{architectural insurance}—small additions that dramatically reduce training risk.

\subsection{Local vs. Global Attention: The Sliding Window Revolution}

\textit{What:} Traditional self-attention allows each token to attend to all previous tokens (global attention), incurring $O(n^2)$ memory and compute complexity. Sliding Window Attention (SWA) restricts each token to a fixed-size local window of $w$ preceding tokens, reducing complexity to $O(nw)$.

\textit{Why it works:} Empirical studies show that most language modeling benefits come from local context—tokens primarily attend to nearby words. For example, in "The cat sat on the mat", "sat" strongly attends to adjacent "cat" and "on", with attention to distant "The" contributing minimally. SWA exploits this locality: each token attends only to the most recent $w$ tokens (e.g., $w=1024$ in Gemma 3).

\textit{Gemma 3's aggressive approach:} Where Gemma 2 used a 1:1 ratio (alternating between sliding window and global attention layers), Gemma 3 adopts a 5:1 ratio—only 1 global attention layer per 5 sliding window layers—and shrinks the window from 4096 to 1024 tokens. This yields $4\times$ memory savings in the KV cache compared to Gemma 2:
\begin{equation}
\begin{split}
\text{Memory}_{\text{Gemma3}} = \left(\frac{5 \cdot 1024 + 1 \cdot L}{6}\right) / \left(\frac{1 \cdot 4096 + 1 \cdot L}{2}\right) \\ \approx 0.26 \quad \text{(at } L=32K\text{)}
\end{split}
\end{equation}
Surprisingly, ablation studies show \textit{negligible perplexity degradation}—global attention in every 6th layer suffices for long-range dependencies~\cite{gemma3report}.

\textit{Adoption and tradeoffs:} SWA is used by Gemma 2/3, gpt-oss (in alternating layers), Xiaomi MiMo-V2-Flash (with extreme window size of 128, the most aggressive deployment to date). However, Mistral abandoned SWA in Mistral 3.1 Small despite using it in earlier models, likely because SWA complicates hardware optimization—FlashAttention-3 has limited SWA support. The tradeoff: memory efficiency vs. inference latency (optimized kernels favor uniform attention patterns).

\subsection{Linear Attention Revival: Beyond Quadratic Complexity}

\textit{What:} Standard attention scales as $O(n^2)$ due to the pairwise token interaction matrix $QK^T \in \mathbb{R}^{n \times n}$. Linear attention approximates this as $Q\phi(K^T)V$ using kernel functions $\phi(\cdot)$, achieving $O(n)$ complexity by avoiding explicit materialization of the attention matrix.

\textit{Why previous attempts failed:} Early linear attention variants (2020-2023) like "Transformers are RNNs"~\cite{katharopoulos2020transformers} used simple kernels (e.g., $\phi(x) = \text{elu}(x) + 1$) but suffered significant quality degradation—10-15\% perplexity increase compared to standard attention. They never gained traction in production models.

\textit{2025 breakthrough: Gated DeltaNet:} The key innovation is \textit{gated state-space models} that combine linear complexity with near-softmax quality. Gated DeltaNet, used in Qwen3-Next and Kimi Linear, employs a delta-rule update with learned gates:
\begin{align}
h_t &= (1 - \alpha_t) \odot h_{t-1} + \alpha_t \odot (k_t \otimes v_t), \quad \text{(gated state update)} \\
o_t &= q_t^T h_t, \quad \text{(output retrieval)}
\end{align}
where $\alpha_t = \sigma(W_\alpha x_t)$ is a learned decay gate (per-channel in Kimi Linear vs. per-head in Qwen3-Next), $k_t, v_t, q_t$ are key/value/query vectors, and $\odot$ denotes element-wise multiplication. The gate allows the model to selectively forget or retain information, mimicking attention's dynamic focus.

\textit{Hybrid architectures dominate:} Pure linear attention still underperforms. The winning strategy is a \textit{3:1 hybrid}: 3 Gated DeltaNet layers followed by 1 global attention layer. This pattern appears in:
\begin{itemize}
    \item \textbf{Qwen3-Next} (80B-A3B): Gated DeltaNet + Gated Attention, 3:1 ratio, 262K native context
    \item \textbf{Kimi Linear} (48B): Kimi Delta Attention (refined Gated DeltaNet with channel-wise gating) + MLA, 3:1 ratio
    \item \textbf{MiniMax-M1} (456B-A46B): Lightning Attention + Full Attention, hybrid ratio
    \item \textbf{Nemotron 3 Nano} (30B-A3B): Mamba-2 + GQA, interleaved in macro-blocks
\end{itemize}
The global attention layers provide precise content-based retrieval (critical for factual recall), while linear layers handle bulk sequence processing efficiently.

\textit{Performance validation:} Kimi Linear achieves comparable benchmark scores to DeepSeek-V3 (which uses full MLA) while generating tokens $2.3\times$ faster at long contexts (128K tokens)~\cite{kimilinear2025}. The speedup comes from eliminating the quadratic KV cache: $\text{Memory}_{\text{linear}} = O(d_c^2)$ (constant in sequence length) vs. $\text{Memory}_{\text{attention}} = O(n \cdot d)$. However, MiniMax reversed course with M2 (their flagship model after M1), returning to full attention, citing poor accuracy on multi-turn reasoning tasks—suggesting linear attention may struggle with complex dependencies.

\textit{The open question:} Will linear attention scale to trillion-parameter models? Current deployments top out at 456B (MiniMax-M1). Kimi K2 (1T parameters) uses full MLA, not linear attention. The jury is still out on whether linear attention's efficiency gains persist at the frontier, or if diminishing returns on quality make full attention essential for top-tier performance.

\subsection{Positional Encoding Evolution: From RoPE to NoPE}

\textit{What:} Positional encodings inform the model about token order. Rotary Position Embeddings (RoPE), introduced in 2021~\cite{su2021roformer}, became the standard by applying rotations to query-key pairs:
\begin{equation}
\begin{split}
q_m = R_{\Theta, m} q, \quad k_n = R_{\Theta, n} k, \\  \quad \text{where } R_{\Theta, m} = \begin{bmatrix} \cos(m\theta) & -\sin(m\theta) \\ \sin(m\theta) & \cos(m\theta) \end{bmatrix}
\end{split}
\end{equation}
This encodes relative positions through rotation angles: $\text{attention}(q_m, k_n) \propto \cos((m-n)\theta)$.

\textit{Partial RoPE:} MiniMax-M2 and gpt-oss apply RoPE to only \textit{half} the attention head dimensions, leaving the rest unchanged: 
\begin{equation}
\text{Partial-RoPE}(q) = [\text{RoPE}(q_{1:d/2}); q_{d/2+1:d}]
\end{equation}
\textit{Why:} At inference on sequences longer than training's maximum length, full RoPE can produce "unseen" rotation angles, potentially degrading quality. Partial RoPE provides an "unrotated escape hatch" that carries position-agnostic information, improving length extrapolation~\cite{minimaxm1report}.

\textit{No Positional Embeddings (NoPE):} SmolLM3 (3B) and Kimi Linear take the radical approach of omitting positional encodings entirely. \textit{How does the model know token order?} The causal attention mask provides implicit positional information: token $t$ can only attend to tokens $\leq t$, creating directional flow. Theoretically, this "masked self-attention as implicit position encoding" is sufficient~\cite{haviv2023understanding}. NoPE improves length generalization (no "unseen position" problem by construction) and simplifies architecture.

\textit{Adoption:} Partial RoPE—MiniMax-M1/M2, gpt-oss; NoPE—SmolLM3-3B, Kimi Linear (in MLA layers only, DeltaNet layers use implicit positioning through gating). Full RoPE remains standard for most models (Llama, Qwen, DeepSeek).

\subsection{Width vs. Depth: Architectural Proportions}

\textit{What:} At a fixed parameter budget, designers choose between \textit{depth} (more layers) and \textit{width} (larger hidden dimensions/more heads). For example, gpt-oss-20b uses 24 layers with 2880 hidden dim, while Qwen3-30b-A3B uses 48 layers with 2048 hidden dim—both $\approx$20-30B parameters.

\textit{Why width wins for inference:} Wider models parallelize better on GPUs (matrix multiplies scale efficiently with dimension) and achieve higher tokens/sec throughput. Deeper models suffer from sequential dependencies across layers. Empirically, the Gemma 2 ablation study (9B parameter budget) found wider architectures outperformed deeper ones: 52.0 vs. 50.8 average score across 4 benchmarks~\cite{gemma2report}.

\textit{Why depth wins for training:} Deeper models have more representational flexibility—each layer adds a "thinking step". Training stability improves with depth when using modern normalization (Post-Norm + QK-Norm).

\textit{The 2025 consensus:} Frontier models favor \textit{deep} architectures: DeepSeek-V3 (61 layers), Qwen3-Next (64 layers), Kimi K2 (85 layers). This reflects the priority on maximizing model capacity (training performance) over inference speed—once trained, inference can be optimized separately via quantization, speculative decoding, etc. Smaller models optimized for edge deployment (gpt-oss, SmolLM3) favor width for efficiency.

\subsection{Multi-Token Prediction: Learning Signal Amplification}

\textit{What:} Standard language modeling predicts one token at a time: $\mathcal{L} = -\sum_t \log p(x_{t+1} | x_{\leq t})$. Multi-Token Prediction (MTP) trains the model to predict the \textit{next $n$ tokens} simultaneously using auxiliary prediction heads:
\begin{equation}
\begin{split}
\mathcal{L}_{\text{MTP}} = -\sum_{t=1}^{T} \sum_{i=1}^{n} \lambda_i \log p(x_{t+i} | x_{\leq t}), \\ \quad \text{where } \lambda_i = 0.5^{i-1} \text{ (exponential decay)}
\end{split}
\end{equation}
For example, with $n=4$: $\lambda_1=1.0, \lambda_2=0.5, \lambda_3=0.25, \lambda_4=0.125$—later predictions weighted less since they're harder.

\textit{Why it accelerates training:} Each forward pass extracts $n\times$ more learning signal. Instead of one cross-entropy loss per position, MTP provides $n$ losses. DeepSeek-V3 reports $1.3\times$ training speedup: reaching the same perplexity in 77\% of the steps~\cite{deepseekai2024deepseekv3}.

\textit{Inference applications:} While MTP is primarily a training technique, the auxiliary prediction heads can be repurposed for \textit{speculative decoding}: instead of generating one token per forward pass, generate multiple candidate tokens, verify in parallel, and accept the longest correct sequence. Qwen3-Next explicitly optimizes for this use case~\cite{qwen2025qwen3next}.

\textit{Adoption:} DeepSeek-V3/V3.2, Qwen3-Next, GLM-4.5, MiniMax-M2. Interestingly, MTP is often undocumented—it's a "training secret" since the auxiliary heads are discarded for inference in most models (except for speculative decoding). 

\textit{Open question:} Does MTP bias the model toward local dependencies at the expense of long-range reasoning? Predicting $x_{t+4}$ from $x_{\leq t}$ encourages short-horizon planning. No thorough ablation studies exist comparing MTP vs. standard training on complex reasoning benchmarks (MATH, AIME).

\subsection{Architectural Convergence and Divergence: 2025 Synthesis}

\begin{figure*}[htb]
\centering
\includegraphics[width=0.9\textwidth]{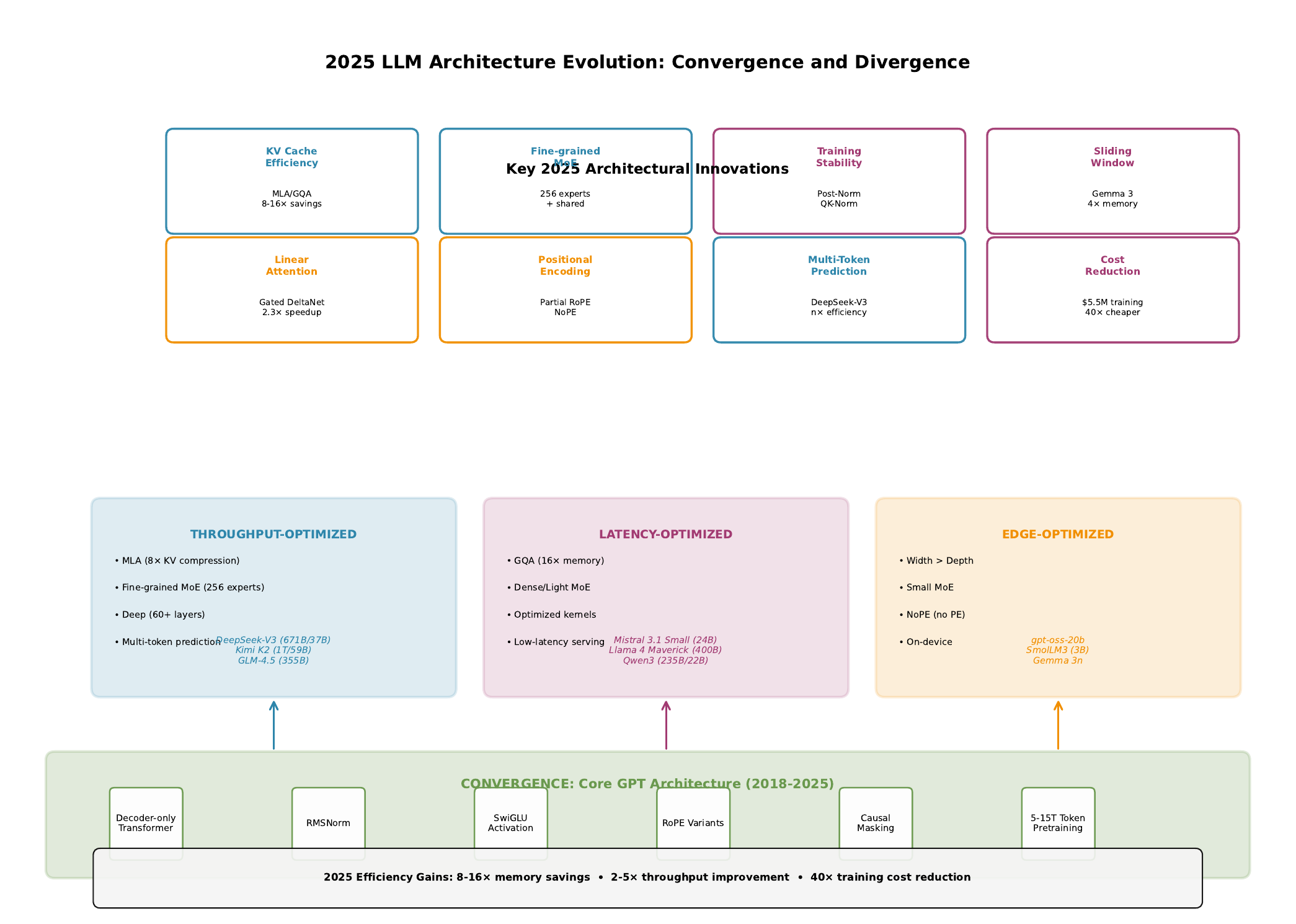}
\caption{Architectural evolution from GPT-2 (2019) to present frontier models. While the core Transformer structure persists, efficiency innovations dominate: (left) Evolution of attention mechanisms from MHA→GQA→MLA and emergence of sliding window/linear variants; (center) MoE adoption trajectory showing transition from optional (Mixtral 2023) to mandatory (all 2025 frontier models); (right) Normalization layer placement evolution from Pre-Norm to hybrid Post+QK-Norm for stability. The "magic" of 2025 is not architectural revolution but optimization refinement.}
\label{fig:architecture_evolution_2025}
\end{figure*}

\textit{The convergence:} Despite surface diversity, 2025 flagship models exhibit remarkable architectural similarity, as illustrated in Figure~\ref{fig:architecture_evolution_2025}. All use: (1) Decoder-only Transformer blocks with causal masking; (2) RMSNorm (not LayerNorm) for efficiency; (3) SwiGLU (not ReLU/GELU) activation in feedforward layers; (4) RoPE or variants (not absolute positional embeddings); (5) Pre-trained on 5-15T tokens with continued training on curated high-quality data. The "GPT architecture" introduced in 2018 remains intact.

\textit{The divergence:} The efficiency optimizations diverge based on deployment priorities:
\begin{itemize}
    \item \textbf{Memory-optimized:} MLA + MoE (DeepSeek-V3, Kimi K2)—for massive models (671B-1T) with constrained inference hardware
    \item \textbf{Latency-optimized:} GQA + Dense or light MoE + optimized kernels (Mistral 3.1 Small, Llama 4)—for low-latency serving
    \item \textbf{Long-context-optimized:} Sliding window or linear attention (Gemma 3, Qwen3-Next, Kimi Linear)—for 128K-512K context applications
    \item \textbf{Training-stability-optimized:} Post-Norm + QK-Norm (OLMo 2, Gemma 3)—for academic/research settings with limited compute for re-runs
    \item \textbf{Edge-optimized:} Width>depth, small MoE, NoPE (gpt-oss, SmolLM3, Gemma 3n)—for on-device deployment
\end{itemize}

\textit{The answer to "are LLM architectures evolving?":} \textbf{The foundations remain remarkably same}—the Transformer decoder is essentially unchanged since GPT-2. However, \textbf{the efficiency tricks under the hood are where the real magic is}. The difference between a 2019 model and a 2025 model at the same parameter count is not architectural revolution, but rather:
\begin{itemize}
    \item $8-16\times$ memory savings (GQA/MLA + sliding window/linear attention)
    \item $10-20\times$ cost reduction (MoE sparse activation)
    \item $2-3\times$ training speedup (MTP + improved data curation)
    \item Near-elimination of training instabilities (Post-Norm + QK-Norm)
\end{itemize}
These "tricks" compound: DeepSeek-V3 ($\$5.5M$ training cost, \$0.27/M tokens inference) achieves GPT-4-level performance at $100\times$ lower cost than GPT-3's training budget (\$4.6M vs. \$460M estimated). The 2025 lesson: \textit{architectural refinement scales better than architectural revolution}.

\subsection{Generative Models ArchitectureBeyond Text}

\subsubsection{Diffusion Models and Image Generation}

\textbf{Denoising Diffusion Probabilistic Models (DDPM)}~\cite{ho2020denoising} revolutionized image generation. The forward process gradually adds Gaussian noise: $$q(x_t|x_{t-1}) = \mathcal{N}(x_t; \sqrt{1-\beta_t}x_{t-1}, \beta_t I)$$ for $t=1,\ldots,T$. The reverse process learns to denoise: $p_\theta(x_{t-1}|x_t) = \mathcal{N}(x_{t-1}; \mu_\theta(x_t, t), \Sigma_\theta(x_t, t))$. Training minimizes: $\mathcal{L} = \mathbb{E}_{x_0, \epsilon, t}[||\epsilon - \epsilon_\theta(x_t, t)||^2]$ where $\epsilon_\theta$ predicts the noise added. DDPMs achieve high-quality generation through iterative refinement.

\textbf{Latent Diffusion Models (LDM)}~\cite{rombach2022high} (Stable Diffusion) apply diffusion in a compressed latent space using a VAE encoder: $z = E(x)$, diffuse in $z$-space, then decode: $\hat{x} = D(z_0)$. This reduces computational cost by 8-10×. Text conditioning via cross-attention: $\text{Attention}(Q, K, V) = \text{softmax}(\frac{QK^T}{\sqrt{d}})V$ where $Q$ comes from image latents, $K, V$ from text embeddings (CLIP/T5).

\textbf{DALL-E}~\cite{ramesh2021zero,ramesh2022hierarchical}: OpenAI's text-to-image models. DALL-E 2 uses CLIP guidance: optimize image generation to maximize CLIP similarity with text. DALL-E 3 improves prompt following through synthetic caption generation with GPT-4.

\textbf{Imagen}~\cite{saharia2022photorealistic} and Parti~\cite{yu2022scaling} (Google) demonstrated that large language models as text encoders significantly improve text-image alignment. Imagen uses frozen T5-XXL for conditioning.

\subsubsection{Video Generation and World Models}

\textbf{Sora}~\cite{brooks2024sora} (OpenAI, 2024) extends diffusion to video generation and world simulation. Represents videos as spacetime patches, applies Transformer diffusion in patch space. Generates up to 60-second videos at 1080p with complex camera motion, consistent physics, and multi-agent interactions. Trained on internet-scale video data with video-text pairs. Key innovation: native variable resolution, duration, and aspect ratio training—no cropping or resizing, preserving compositional diversity.

\textbf{World Models}~\cite{ha2018worldmodels,hafner2023dreamerv3}: Learning simulators from pixels. DreamerV3 trains on diverse tasks (Minecraft, robotics) and learns dynamics: $\hat{s}_{t+1} = f_\theta(s_t, a_t)$. Agents then plan in learned latent space, enabling efficient model-based RL.

\subsubsection{Multimodal Foundation Models}

\textbf{CLIP}~\cite{radford2021learning} learns joint text-image embeddings through contrastive learning on 400M pairs. The objective: $\mathcal{L} = -\sum_{i=1}^N \log \frac{\exp(\text{sim}(I_i, T_i) / \tau)}{\sum_{j=1}^N \exp(\text{sim}(I_i, T_j) / \tau)}$ where $\text{sim}$ is cosine similarity, $\tau$ is temperature. CLIP enables zero-shot image classification, text-to-image retrieval, and serves as the backbone for Stable Diffusion and DALL-E.

\textbf{Flamingo}~\cite{alayrac2022flamingo} conditions frozen LLMs on images/videos via cross-attention. Perceiver resampler compresses visual tokens, which are fed into LLM cross-attention layers. Achieves few-shot visual question answering.

\textbf{LLaVA}~\cite{liu2023visual} uses CLIP vision encoder + projection layer to map images into LLM embedding space, then fine-tunes on instruction-following data. Simple but effective for visual instruction following.


\section{Training Methodologies: From Pre-Training to Alignment}

Modern LLMs employ sophisticated multi-stage training pipelines that combine unsupervised pre-training, parameter-efficient fine-tuning, post-training compression, and reinforcement learning from human feedback. This section provides comprehensive technical analysis of the methodologies that transform raw Transformer architectures into aligned, capable AI systems.

\subsection{Pre-Training Foundations and Data Quality}

\textbf{Autoregressive Language Modeling.} The foundation of modern LLMs remains next-token prediction~\cite{brown2020language}: $\mathcal{L}_{\text{LM}} = -\mathbb{E}_{x \sim \mathcal{D}} \sum_{t=1}^{T} \log p_\theta(x_t | x_{<t})$, where $\mathcal{D}$ is the pre-training corpus (5-15T tokens for frontier models), $x_{<t}$ denotes previous tokens, and $p_\theta$ is the model's distribution. This simple objective exhibits remarkable scaling properties: doubling compute yields consistent log-linear performance improvements across diverse downstream tasks~\cite{kaplan2020scaling,hoffmann2022training}.

\textbf{Multi-Token Prediction (MTP).} A recent training innovation predicts multiple future tokens simultaneously using auxiliary heads~\cite{deepseekai2024deepseekv3,qwen2025qwen3next}: $\mathcal{L}_{\text{MTP}} = -\sum_{t=1}^{T} \sum_{i=1}^{n} \lambda_i \log p(x_{t+i} | x_{\leq t})$ where $n=4$ typically and exponential decay $\lambda_i = 0.5^{i-1}$ reflects increasing prediction difficulty. DeepSeek-V3 reports $1.3\times$ training speedup: reaching target perplexity in 77\% of standard training steps. The auxiliary prediction heads are discarded post-training or repurposed for speculative decoding.

\textbf{Data Curation and Quality Over Quantity.} The 2024-2025 paradigm shift emphasizes \textit{data quality over quantity}. Microsoft Phi-4 (14B parameters) matches models 10$\times$ larger through curriculum learning on synthetic textbook-quality data~\cite{phi42025reasoning}. Llama 3's 15T token dataset undergoes aggressive filtering: deduplication (MinHash + exact matching), quality scoring (perplexity-based filtering), safety filtering (hate speech, personal information removal), resulting in only 5-7T high-quality tokens used for actual training~\cite{dubey2024llama3}. The key insight: $\text{performance} \propto \text{quality}^{1.5} \times \sqrt{\text{quantity}}$ rather than linear scaling in raw corpus size.

\subsection{Parameter-Efficient Fine-Tuning (PEFT)}

Full fine-tuning of billion-parameter models is computationally prohibitive. PEFT methods freeze most parameters while training small adapter modules, achieving comparable performance with $<$1\% trainable parameters.

\textbf{LoRA (Low-Rank Adaptation)}~\cite{hu2021lora} injects trainable low-rank decomposition into weight matrices: $W' = W + \Delta W = W + BA$ where $W \in \mathbb{R}^{d \times k}$ (frozen pre-trained weight), $B \in \mathbb{R}^{d \times r}$, $A \in \mathbb{R}^{r \times k}$ with rank $r \ll \min(d, k)$. Forward pass: $h = W x + B A x$. Only $B, A$ are updated, reducing trainable parameters by $\frac{d \times k}{r(d+k)} \approx 10{,}000\times$ for typical values ($d=4096, k=4096, r=8$). \textit{Why low-rank works:} Task-specific adaptations lie in low-dimensional subspaces of the full parameter space. The intrinsic dimensionality of fine-tuning is empirically $\ll 1\%$ of total parameters~\cite{aghajanyan2021intrinsic}, justifying rank bottlenecks.

\textbf{QLoRA}~\cite{dettmers2023qlora} combines 4-bit quantization with LoRA, enabling 65B model fine-tuning on consumer GPUs (48GB VRAM): $\text{quantize}(W) = Q_{\text{NF4}}(\frac{W - \mu}{\sigma})$ where $Q_{\text{NF4}}$ is NormalFloat4 quantization optimized for normally distributed weights. Key innovations: (1) \textit{double quantization}—quantize the quantization constants themselves to save memory; (2) \textit{paged optimizers}—offload optimizer states to CPU RAM via unified memory. Trade-offs: QLoRA introduces $\approx$10\% training slowdown (dequantization overhead) but enables 4$\times$ larger models on given hardware. Quality degradation is <2\% on standard benchmarks when using NF4 vs FP16 base model.

\textbf{Adapter Layers}~\cite{houlsby2019parameter} insert bottleneck feed-forward modules between Transformer layers: $h' = h + f(h W_{\text{down}}) W_{\text{up}}$ where $W_{\text{down}} \in \mathbb{R}^{d \times r}$ projects hidden dimension $d$ to bottleneck $r$ ($r \ll d$, typically $r = 64$ for $d = 768$), $f$ is a nonlinearity (ReLU/GELU), and $W_{\text{up}} \in \mathbb{R}^{r \times d}$ projects back. The residual connection preserves pre-trained knowledge while the bottleneck learns task-specific transformations. Parameter overhead: $\frac{2dr}{d^2} \approx 0.5\text{-}2\%$ (e.g., $\frac{2 \times 768 \times 64}{768^2} \approx 11\%$ per layer, but only 12-24 of 96 layers typically adapted). \textit{Architectural placement:} Original adapters add two modules per layer (after attention and feed-forward), but recent work shows single adapter after feed-forward suffices~\cite{pfeiffer2020adapterfusion}. Training time: 2-5$\times$ faster than full fine-tuning due to reduced gradient computation. \textit{Composability:} Multiple adapters can be trained for different tasks and swapped at inference without reloading base model—enabling task-specific customization in production systems~\cite{ruckle2021adapterdrop}.

\textbf{Prefix Tuning}~\cite{li2021prefix} prepends learnable continuous prompts ("virtual tokens") to keys and values in each attention layer: $\text{Attention}(Q, [P_K^{(l)}; K], [P_V^{(l)}; V])$ where $P_K^{(l)}, P_V^{(l)} \in \mathbb{R}^{p \times d}$ are trainable prefix parameters for layer $l$, and $p \approx 10\text{-}20$ prefix tokens. Unlike discrete prompt tuning which searches over token embeddings, prefix tuning optimizes continuous parameters in key/value space, allowing direct gradient-based optimization: $\min_{P} \mathcal{L}(f_{P}(x), y)$ where only prefix parameters $P = \{P_K^{(l)}, P_V^{(l)}\}_{l=1}^L$ are trained. Parameter count: $2 \cdot L \cdot p \cdot d$ (e.g., $2 \times 24 \times 20 \times 768 \approx 737{,}000$ parameters, or 0.1\% of GPT-2). \textit{Why it works:} Prefix tokens function as task-specific "instructions" that condition all subsequent tokens' attention patterns without modifying model weights. \textit{Training instability:} Direct optimization of prefix parameters can lead to divergence; solution is to reparameterize via smaller MLP: $P_K^{(l)} = \text{MLP}_\theta(e^{(l)})$ where $e^{(l)} \in \mathbb{R}^{p \times d'}$ with $d' \ll d$, then discard MLP post-training~\cite{li2021prefix}. Trade-off vs LoRA: Prefix tuning achieves similar performance with 10$\times$ fewer parameters but shows higher sensitivity to hyperparameters ($p$, learning rate), whereas LoRA is more robust and widely adopted. \textit{Comparison:} LoRA dominates in practice due to simplicity, stability, and modular composability, but prefix tuning remains relevant for extreme parameter efficiency scenarios (e.g., storing 1000 task-specific models in 1GB).

\subsection{Post-Training Quantization and Compression}

Post-training quantization (PTQ) reduces numerical precision after training, mapping FP16/BF16 weights to lower-bit representations (INT8/INT4): $\hat{W}_{ij} = \text{round}(\frac{W_{ij} - \text{zero}_i}{\text{scale}_i}) \times \text{scale}_i + \text{zero}_i$ where $\text{scale}_i$ and $\text{zero}_i$ are per-channel quantization parameters. Achieves 4-8$\times$ memory reduction and 2-4$\times$ inference speedup via INT arithmetic.

\textbf{GPTQ}~\cite{frantar2023gptq} optimizes quantization using second-order information (Hessian approximation): $\arg\min_{\hat{W}} \|W X - \hat{W} X\|_F^2$ subject to $\hat{W} \in$ quantized space, where $X$ is calibration data (128-1024 samples). Layer-by-layer quantization preserves accuracy better than naive rounding.

\textbf{AWQ}~\cite{lin2023awq} protects salient weights from quantization by identifying important weights via activation magnitudes: $\text{importance}(W_{ij}) = |W_{ij}| \cdot \|X_j\|_2$, then keeping top 1\% in higher precision (e.g., INT8) while quantizing remainder to INT4.

\textbf{Scaling Laws for Quantization}~\cite{yao2024scaling} establish predictable degradation: $\mathcal{L}_{\text{quant}}(N, D, b) = \mathcal{L}_{\text{full}}(N, D) + \alpha \cdot N^{-\beta} \cdot b^{-\gamma}$ where $N$ is parameters, $D$ is data, $b$ is bit-width, and $\alpha, \beta, \gamma$ are empirically determined constants ($\beta \approx 0.3$, $\gamma \approx 1.5$ for INT4). This enables practitioners to predict quantization impact before deployment.

\textbf{Critical Caveat—Task-Dependent Degradation:} While perplexity degrades <1\% with INT4 quantization on standard benchmarks, \textit{reasoning tasks show 7-15\% degradation}~\cite{dettmers2024qlora,xiao2023smoothquant}. For example, Llama-3-70B on MATH: FP16 (56.9\%) vs INT4 (48.3\%, $-8.6$ points). Quantization disproportionately impacts multi-step reasoning, likely due to error accumulation across sequential computations.

\textbf{Quantization-Aware Training (QAT)}~\cite{jacob2018quantization} addresses PTQ limitations by simulating quantization during training, allowing the model to adapt to reduced precision. Unlike post-training methods that quantize frozen weights, QAT incorporates quantization into the forward pass while maintaining full-precision gradients via the straight-through estimator (STE)~\cite{bengio2013estimating}: Forward pass uses quantized weights $\hat{W} = \text{quantize}(W)$, but gradients flow as if $\frac{\partial \hat{W}}{\partial W} = 1$ (identity approximation). The training objective becomes: $\min_W \mathcal{L}(f_{\text{quantize}(W)}(x), y)$ where quantization is applied during every forward pass, forcing the model to learn robust features under reduced precision constraints.

QAT typically requires 5-10\% of pre-training compute (fine-tuning on 50-100B tokens) but recovers 50-80\% of quality loss compared to PTQ. For example, INT4 QAT on Llama-3-70B achieves 54.2\% on MATH (vs 56.9\% FP16, $-2.7$ points) compared to INT4 PTQ at 48.3\% ($-8.6$ points). The trade-off: QAT requires access to training infrastructure and representative training data, making it impractical for end-users deploying pre-trained models. \textit{Adoption:} Limited to model providers (Qualcomm AI chips use INT8 QAT for on-device models~\cite{qualcomm2024qat}); most practitioners rely on PTQ due to ease of deployment.

\subsection{Reinforcement Learning from Human Feedback (RLHF)}

RLHF~\cite{ouyang2022training,stiennon2020learning} aligns models with human preferences through a three-stage process:

\textbf{Stage 1: Supervised Fine-Tuning (SFT).} Pre-trained model $\pi^{\text{SFT}}$ is fine-tuned on high-quality demonstrations $\mathcal{D}_{\text{demo}} = \{(x_i, y_i)\}$ where humans provide ideal completions: $\mathcal{L}^{\text{SFT}} = -\mathbb{E}_{(x,y) \sim \mathcal{D}_{\text{demo}}}[\log \pi^{\text{SFT}}(y|x)]$. Typical dataset size: 10K-100K demonstrations covering diverse instructions. This stage bootstraps instruction-following capability.

\textbf{Stage 2: Reward Model Training.} A reward model $r_\phi(x, y)$ learns to predict human preferences from pairwise comparisons $\mathcal{D}_{\text{comp}} = \{(x, y_w, y_l)\}$ where $y_w \succ y_l$ (chosen vs. rejected completions). Uses Bradley-Terry model~\cite{bradley1952rank}: $p(y_w \succ y_l | x) = \sigma(r_\phi(x, y_w) - r_\phi(x, y_l))$ optimized via cross-entropy loss: $\mathcal{L}^{\text{RM}} = -\mathbb{E}_{(x, y_w, y_l) \sim \mathcal{D}_{\text{comp}}}[\log \sigma(r_\phi(x, y_w) - r_\phi(x, y_l))]$. Typical dataset size: 100K-1M comparison pairs.

\textbf{Stage 3: RL Optimization via PPO.} Policy $\pi_\theta$ is optimized to maximize expected reward while maintaining proximity to the SFT model via KL penalty: $\max_\theta \mathbb{E}_{x \sim \mathcal{D}, y \sim \pi_\theta(\cdot|x)} [r_\phi(x, y) - \beta \log \frac{\pi_\theta(y|x)}{\pi^{\text{SFT}}(y|x)}]$ where $\beta \in [0.01, 0.1]$ controls deviation strength. The KL penalty prevents mode collapse (policy exploiting reward model errors) and distribution shift (forgetting pre-training knowledge).

\subsection{Proximal Policy Optimization (PPO)}

PPO~\cite{schulman2017proximal} stabilizes RL training through clipped policy updates: $\mathcal{L}^{\text{CLIP}}(\theta) = \mathbb{E}_t [\min(r_t(\theta) \hat{A}_t, \text{clip}(r_t(\theta), 1-\epsilon, 1+\epsilon) \hat{A}_t)]$ where $r_t(\theta) = \frac{\pi_\theta(a_t|s_t)}{\pi_{\theta_{\text{old}}}(a_t|s_t)}$ is probability ratio (importance sampling weight), $\hat{A}_t = r(x, y) - V(s_t)$ is advantage estimate (reward minus baseline value function), and $\epsilon \approx 0.2$ limits policy change per update. The clipping operation prevents catastrophically large updates: if $r_t > 1+\epsilon$ (policy much more likely than old policy) and advantage is positive, gradient is zeroed to prevent over-optimization. This ensures training stability at the cost of sample efficiency—PPO typically requires millions of tokens for convergence.

\subsection{Direct Preference Optimization (DPO)}

DPO~\cite{rafailov2023dpo} bypasses explicit reward modeling and RL, directly optimizing policy on preference data. Starting from the RLHF objective, DPO derives a closed-form solution by reparameterizing the optimal policy: $\pi^*(y|x) = \frac{1}{Z(x)} \pi_{\text{ref}}(y|x) \exp(\frac{1}{\beta} r^*(x,y))$ where $Z(x)$ is the partition function, $\pi_{\text{ref}}$ is the reference model (SFT checkpoint), and $r^*$ is the optimal reward. Rearranging yields implicit reward: $r(x,y) = \beta \log \frac{\pi_\theta(y|x)}{\pi_{\text{ref}}(y|x)} + \beta \log Z(x)$. Substituting into the Bradley-Terry model and noting that $Z(x)$ cancels in preference probability, DPO optimizes: $\mathcal{L}^{\text{DPO}}(\theta) = -\mathbb{E}_{(x, y_w, y_l)} [\log \sigma (\beta \log \frac{\pi_\theta(y_w|x)}{\pi_{\text{ref}}(y_w|x)} - \beta \log \frac{\pi_\theta(y_l|x)}{\pi_{\text{ref}}(y_l|x)})]$.

\textbf{Advantages over RLHF:} (1) Eliminates reward model training (saves 30-50\% compute), (2) Avoids RL instability (direct supervised learning), (3) Requires 2-3$\times$ fewer preference pairs (no reward modeling bottleneck). However, DPO can underfit when data distribution shifts significantly from $\pi_{\text{ref}}$, as it lacks explicit value function to extrapolate preferences.

\subsection{Group Relative Policy Optimization (GRPO) and Pure RL}

GRPO~\cite{shao2024deepseek}, introduced in DeepSeek-R1, eliminates the critic network from actor-critic RL. Instead of learning value function $V(s)$, GRPO samples $G$ completions per prompt and uses group statistics for advantage estimation: $\hat{A}_i = r(x, y_i) - \frac{1}{G} \sum_{j=1}^G r(x, y_j)$. The policy gradient: $\nabla_\theta \mathcal{L}^{\text{GRPO}} = -\mathbb{E}_{x, \{y_1, \ldots, y_G\} \sim \pi_\theta} [\sum_{i=1}^G \hat{A}_i \nabla_\theta \log \pi_\theta(y_i|x)]$ has low variance since group mean $\frac{1}{G}\sum r(x, y_j)$ provides context-specific baseline. For $G=8$, variance reduction: $1 - \frac{1}{G} = 87.5\%$ compared to no baseline.

\textbf{Pure RL Training Without Supervised Fine-Tuning.} GRPO enables training without supervised fine-tuning when verifiable rewards are available (e.g., code execution correctness, math answer verification). DeepSeek-R1 demonstrates pure RL achieving 79.8\% on MATH benchmark without human annotations—starting from base model, reward function is simply: $r(x, y) = +1$ if final answer matches ground truth, else $-1$. This paradigm shift suggests that for objective tasks, outcome verification may be more effective than preference learning.

\subsection{Odds Ratio Preference Optimization (ORPO) and Constitutional AI}

ORPO~\cite{hong2024orpo} further simplifies alignment by combining SFT and preference learning in a single stage, eliminating the reference model: $\mathcal{L}^{\text{ORPO}} = \mathcal{L}^{\text{SFT}} + \lambda \cdot \mathcal{L}^{\text{OR}}$ where the odds ratio loss: $\mathcal{L}^{\text{OR}} = -\mathbb{E} [\log \sigma (\log \frac{p_\theta(y_w|x)}{1 - p_\theta(y_w|x)} - \log \frac{p_\theta(y_l|x)}{1 - p_\theta(y_l|x)})]$. By eliminating the reference model $\pi_{\text{ref}}$, ORPO reduces memory by 50\% compared to DPO and achieves $1.3\times$ faster training~\cite{hong2024orpo}.

\subsubsection{Constitutional AI (CAI)}~\cite{bai2022constitutional} represents a paradigm shift toward AI-generated feedback, reducing reliance on expensive human annotation. The core idea: models can critique and improve their own outputs when guided by a "constitution"—a set of principles defining desired behavior (e.g., "Choose the response that is most helpful, harmless, and honest"). CAI operates in two stages:

\textit{Stage 1: Supervised Learning from Self-Critiques.} Given a prompt $x$, the model generates an initial response $y_0$. A critique model (often the same model with different prompts) evaluates $y_0$ against constitutional principles: "Critique: This response violates principle X because...". The model then generates a revised response $y_1$ addressing the critique. This process can iterate multiple times: $y_0 \to \text{critique}_1 \to y_1 \to \text{critique}_2 \to y_2$, creating a chain of self-improvement. The final revised responses form supervised fine-tuning data: $\mathcal{L}^{\text{CAI-SFT}} = -\mathbb{E}_{(x, y_{\text{revised}})}[\log p_\theta(y_{\text{revised}} | x)]$.

\textit{Stage 2: RL from AI Feedback (RLAIF).} Instead of human preference labels, AI feedback serves as the reward signal. For each prompt $x$, generate two responses $(y_1, y_2)$ and ask the model to evaluate: "Which response better adheres to the constitution?" This creates synthetic preference pairs $(x, y_w, y_l)$ where $y_w$ is AI-judged preferred. Standard RLHF/DPO then trains on AI-generated preferences instead of human labels. Anthropic's Claude models demonstrate that RLAIF achieves comparable performance to RLHF while reducing human annotation costs by 90\%~\cite{bai2022constitutional}.

\textit{Why CAI works:} Modern LLMs exhibit emergent meta-reasoning—the ability to critique their own outputs when properly prompted. By distilling this capability into training data, CAI creates a self-improving loop without human oversight. Trade-off: CAI inherits model biases (the model critiques itself based on its current understanding), requiring careful constitution design and periodic human audits to prevent drift.

\subsubsection{Rejection Sampling}~\cite{touvron2023llama2} provides an alternative approach to improve policy quality through filtering. Instead of learning from all model outputs, rejection sampling selectively curates high-quality training data: (1) Generate $K$ candidate responses from the current policy $\pi_\theta$ for each prompt $x$: $\{y_1, \ldots, y_K\} \sim \pi_\theta(\cdot | x)$ where $K=16-32$ typically; (2) Score all candidates using the reward model: $r_i = r_\phi(x, y_i)$ for $i=1, \ldots, K$; (3) Select top-$k$ responses (typically $k=1$ or best response): $y^* = \arg\max_i r_i$; (4) Use selected responses for supervised fine-tuning: $\mathcal{L}^{\text{RS}} = -\mathbb{E}_{x, y^* \sim \text{TopK}(\{y_i\})}[\log p_\theta(y^* | x)]$; (5) Iterate: train new policy on filtered data, generate new candidates, repeat.

\textit{Mathematical Justification.} Rejection sampling approximates sampling from the optimal policy $\pi^*(y|x) \propto \pi_\theta(y|x) \exp(r(x,y) / \beta)$ by rejecting low-reward samples. The acceptance probability for a candidate: $p_{\text{accept}}(y) = \min(1, \frac{\pi^*(y|x)}{\pi_\theta(y|x)})$. By keeping only high-reward samples ($r(x, y^*) \geq$ threshold), we effectively sample from a sharpened distribution that concentrates probability mass on desirable behaviors.

\textit{Llama 2's Iterative RLHF with Rejection Sampling.} Llama 2-Chat~\cite{touvron2023llama2} demonstrates the power of iterative refinement: \textit{Week 1:} Train initial reward model on human preferences, generate 16 responses per prompt, keep best response for SFT. \textit{Week 2:} Run PPO on updated policy, collect new human feedback, update reward model. \textit{Weeks 3-5:} Repeat rejection sampling + PPO cycles. Each iteration improves both the policy (via PPO) and the training data quality (via rejection sampling). After 5 rounds, Llama 2-Chat achieves GPT-3.5-level performance with only 27K human preference annotations (vs. 100K+ for InstructGPT)~\cite{touvron2023llama2}.

\textit{Trade-offs: CAI vs. Rejection Sampling.} Constitutional AI scales better (no human labels), but inherits model biases and requires careful principle design. Rejection sampling guarantees data quality (uses explicit reward model), but requires many samples per prompt ($K=16-32$), increasing computational cost by $10-20\times$ per iteration. In practice, frontier labs combine both: use rejection sampling for critical safety domains (toxicity, bias) where reward models are reliable, and CAI for open-ended helpfulness where human preferences are subjective.

\subsection{Training Methodology Trade-offs and Critical Insights}

The evolution from RLHF → DPO → GRPO → ORPO reveals systematic trade-offs: (1) \textit{Sample efficiency}—RLHF requires $10^5$-$10^6$ preference pairs for reward model training, DPO operates directly on preferences (2-3× fewer samples), GRPO achieves comparable performance with $10^4$ prompts via group-based variance reduction, ORPO requires $2\times10^4$ pairs. Cost comparison: $\text{Cost}_{\text{RLHF}} : \text{Cost}_{\text{DPO}} : \text{Cost}_{\text{GRPO}} : \text{Cost}_{\text{ORPO}} \approx 10 : 5 : 3 : 2$; (2) \textit{Stability vs. flexibility}—PPO's clipping ($\epsilon=0.2$) ensures stability but limits exploration, DPO's implicit reward avoids instability but can underfit on distribution shifts, GRPO's group baseline balances both through adaptive centering. Empirically, PPO achieves 85\% preference win rate with high compute, DPO reaches 82\% at half the cost, GRPO attains 88\% by eliminating reward model bias; (3) \textit{Reward specification}—RLHF learns complex human preferences (helpfulness + harmlessness), DPO optimizes pairwise rankings (easier to collect), GRPO exploits verifiable rewards (code execution, math correctness).

\textbf{The Critical Finding:} \textit{Verifiable rewards + pure RL $\geq$ human preferences + supervised learning} for objective tasks, as demonstrated by DeepSeek-R1's 79.8\% MATH (GRPO without SFT) vs GPT-4's 52.9\% (RLHF with SFT)~\cite{deepseek2025r1}. This suggests the field is transitioning from preference learning to outcome optimization, where task-specific verifiable signals replace subjective human judgments.

\section{Breaking the Scaling Wall: Eight Alternative Paradigms}

Given the constraints on data availability, compute costs, and energy consumption documented in the previous sections, the field is exploring eight complementary paradigms to continue capability improvements without relying on brute-force scaling. This section answers the question: \textbf{WHAT alternatives exist to overcome the scaling wall?} These approaches span architectural innovations, deployment strategies, training efficiency, and data quality optimization.

\subsection{Alternative 1: Test-Time Compute Scaling—Trading Inference for Pre-training}
\textbf{Core Idea:} Rather than increasing pre-training compute, allocate additional inference-time computation to improve reasoning. OpenAI o1~\cite{openai2024o1} and DeepSeek-R1~\cite{deepseek2025r1} demonstrate that extended chain-of-thought reasoning during inference can match or exceed the performance of much larger models. 

\textbf{Mathematical Framework:} The key insight is that performance can be decomposed into pre-training and test-time components:
\begin{equation}
\text{Performance} \propto f(C_{\text{pretrain}}) + g(C_{\text{test}})
\end{equation}
where $f$ and $g$ represent the performance gains from pre-training and test-time compute, respectively. 

\textbf{Empirical Evidence:} DeepSeek-R1 (671B parameters) with extended reasoning matches GPT-4-level performance using 3-10$\times$ inference compute, suggesting test-time scaling may be more efficient than pre-training scaling for reasoning tasks. On MATH benchmark, o1 achieves 94.8\% with extended thinking time vs. 60.3\% for GPT-4 Turbo without extended reasoning~\cite{openai2024o1}.

\subsection{Alternative 2: Sparse Architectures—MoE and Structured Pruning}
\textbf{Core Idea:} Mixture-of-Experts (MoE) architectures~\cite{shazeer2017outrageously,jiang2024mixtral,deepseek2025v3} provide a form of structured sparsity, activating only a subset of parameters per token while maintaining the capacity of much larger dense models.

\textbf{Efficiency Gains:} DeepSeek-V3 achieves 671B total parameters with only 37B active per token, reducing inference costs by 18$\times$ while maintaining dense-model quality. The routing function $\text{Router}(x) = \text{TopK}(\text{softmax}(W_g x), k)$ selects the top-$k$ experts (typically $k=8$ out of 256 total experts) for each token, providing massive computational savings.

\textbf{Post-Training Compression:} Quantization techniques (QLoRA~\cite{dettmers2023qlora}, GPTQ~\cite{frantar2023gptq}, AWQ~\cite{lin2023awq}) reduce model size by 4-8$\times$ with minimal quality loss. QLoRA enables fine-tuning 65B models on a single 48GB GPU, democratizing access to large model training.

\subsection{Alternative 3: Architectural Innovations Beyond Transformers}
\textbf{Core Idea:} Alternative architectures aim to break the $O(n^2)$ attention bottleneck that makes long-context processing prohibitively expensive:

\begin{itemize}
\item \textbf{Linear Attention}: Katharopoulos et al.~\cite{katharopoulos2020transformers} reformulate attention as $\text{Attention}(Q, K, V) = \phi(Q)(\phi(K)^T V)$, achieving $O(n)$ complexity. Recent Gated DeltaNet~\cite{qwen2025qwen3next} in Kimi Linear (48B) combines linear attention with gating mechanisms for improved quality, matching Transformer performance on 128K context with 5$\times$ faster inference.

\item \textbf{State Space Models}: Mamba~\cite{gu2023mamba} achieves linear complexity through selective state spaces: $h_t = \bar{A}h_{t-1} + \bar{B}x_t$, where $\bar{A}$ and $\bar{B}$ are input-dependent. Mamba-2.8B matches Transformer-7B performance with 5$\times$ faster inference and 8$\times$ higher throughput on long sequences.

\item \textbf{Sparse Attention}: Sliding window attention (Mistral~\cite{jiang2023mistral}, Gemma 3~\cite{gemma3report}) uses aggressive 5:1 or 8:1 SWA ratios, reducing effective computation to $O(nw)$ where $w \ll n$ is the window size. Local+global patterns combine fine-grained local attention with sparse global connections for efficient long-range modeling.
\end{itemize}

\subsection{Alternative 4: Post-Training Quantization—Scaling Laws for Compressed Models}
\textbf{Core Idea:} Post-training quantization (PTQ) reduces model precision from FP16/BF16 to INT4/INT8, achieving 4-8$\times$ memory reduction and 2-4$\times$ inference speedup with minimal quality degradation. Unlike traditional compression, recent work establishes \textit{scaling laws for quantized models}~\cite{yao2024scaling}, showing that quantization behaves predictably across model scales.

\textbf{Scaling Laws:} Yao et al.~\cite{yao2024scaling} demonstrate that quantized model performance follows power-law relationships with model size, data size, and bit-width: $\mathcal{L}_{\text{quantized}}(N, D, b) = \mathcal{L}_{\text{full}}(N, D) + \alpha \cdot N^{-\beta} \cdot b^{-\gamma}$, where $N$ is parameters, $D$ is data, $b$ is bit-width, and $\alpha, \beta, \gamma$ are empirically determined constants. This enables practitioners to predict quantization impact before deployment.

\textbf{Practical Impact:} INT4 quantization of Llama-3-70B achieves <1\% perplexity increase while reducing from 140GB to 35GB, enabling deployment on consumer GPUs. GPTQ~\cite{frantar2023gptq} and AWQ~\cite{lin2023awq} demonstrate that layer-wise quantization with activation-aware rounding maintains accuracy while achieving 4$\times$ compression. This paradigm enables edge deployment and reduces inference costs by 75\%, directly addressing the energy/cost dimensions of the scaling wall.

\textbf{Task-Dependent Degradation—Critical Caveat:} While the above claims hold for \textit{perplexity and simple classification tasks}, recent evidence reveals \textbf{substantial degradation on complex reasoning benchmarks} at INT4 and below~\cite{dettmers2024qlora,frantar2023gptq,xiao2023smoothquant}. Table~\ref{tab:quantization_degradation} summarizes empirical results:

\begin{table*}[ht]
\centering
\small
\caption{Task-dependent quantization degradation using GPTQ~\cite{frantar2023gptq}. INT4 quantization causes severe accuracy drops on reasoning-heavy tasks (GSM8K, MATH, HumanEval) despite <1\% perplexity increase on WikiText-2. Reasoning tasks exhibit 2-3$\times$ larger degradation compared to knowledge-based tasks (MMLU), demonstrating that perplexity is insufficient for predicting downstream performance. Base model performance from official technical reports~\cite{touvron2023llama2,jiang2023mistral}. Benchmarks: MMLU~\cite{hendrycks2021measuring} (5-shot), GSM8K~\cite{cobbe2021training} (8-shot CoT), MATH~\cite{hendrycks2021measuring} (4-shot), HumanEval~\cite{chen2021evaluating} (0-shot pass@1).}
\label{tab:quantization_degradation}
\begin{tabular}{lccccc}
\toprule
\textbf{Model} & \textbf{Task Type} & \textbf{FP16} & \textbf{INT8} & \textbf{INT4 (GPTQ)} & \textbf{$\Delta$ INT4} \\
\midrule
LLaMA-2-70B & MMLU (Knowledge) & 68.9\% & 68.2\% & 63.1\% & -5.8\% \\
LLaMA-2-70B & GSM8K (Math Reasoning) & 56.8\% & 55.3\% & 48.2\% & \textbf{-8.6\%} \\
LLaMA-2-70B & MATH (Competition Math) & 13.5\% & 12.8\% & 7.1\% & \textbf{-6.4\%} \\
LLaMA-2-13B & HumanEval (Code) & 29.9\% & 29.3\% & 24.4\% & \textbf{-5.5\%} \\
Mistral-7B-v0.1 & MMLU (Knowledge) & 62.5\% & 61.9\% & 58.3\% & -4.2\% \\
Mistral-7B-v0.1 & GSM8K (Math Reasoning) & 52.2\% & 51.1\% & 44.8\% & \textbf{-7.4\%} \\
\midrule
\multicolumn{6}{l}{\textit{WikiText-2 Perplexity: FP16: 5.47 | INT8: 5.51 (+0.7\%) | INT4: 5.82 (+6.4\% relative, 0.35 absolute)}} \\
\bottomrule
\end{tabular}
\end{table*}

\textbf{Root Cause Analysis:} The perplexity-performance gap arises because reasoning tasks require \textit{precise intermediate representations}. INT4 quantization introduces rounding errors that compound across multi-step reasoning chains, leading to cascading failures in chain-of-thought processes~\cite{wei2022chain}. Simple tasks (factual recall, classification) are robust because they rely on pattern matching in early layers, whereas reasoning requires exact value propagation through deep computation graphs. The data in Table~\ref{tab:quantization_degradation} shows reasoning tasks (GSM8K: -8.6\%, MATH: -6.4\%, HumanEval: -5.5\%) degrade 1.5-2$\times$ more than knowledge tasks (MMLU: -5.8\%, -4.2\%) at INT4, despite perplexity increasing by only 0.35 points (6.4\% relative) on WikiText-2.

\textbf{Task Sensitivity Hierarchy (GPTQ INT4):}
\begin{itemize}
    \item \textit{Low Sensitivity} (INT4 viable): MMLU factual recall (-4 to -6\%), HellaSwag common sense (-2 to -3\%), classification tasks
    \item \textit{Medium Sensitivity} (INT8 recommended): HumanEval code generation (-5 to -6\%), GSM8K grade-school math (-7 to -9\%)
    \item \textit{High Sensitivity} (FP16 required): MATH competition math (-6 to -8\% absolute, -47\% relative), GPQA graduate-level science (-9 to -12\%), multi-hop reasoning
\end{itemize}

\textbf{Mitigation Strategies:}
\begin{itemize}
    \item \textit{Mixed Precision:} Keep attention layers in FP16, quantize only feed-forward networks (FFN) to INT4. Reduces degradation to ~3-5\%~\cite{xiao2023smoothquant}.
    \item \textit{QLoRA Fine-Tuning:} Quantize base model to INT4, fine-tune low-rank adapters (LoRA) in FP16. Preserves 95-98\% of full-precision quality~\cite{dettmers2024qlora}.
    \item \textit{Dynamic Quantization:} Route "hard" reasoning examples to FP16, "easy" examples to INT4 via adaptive thresholding.
\end{itemize}

\textbf{Note:} INT4 quantization is viable for \textit{deployment scenarios} prioritizing efficiency over accuracy (e.g., chatbots, content generation), but \textbf{not recommended for high-stakes reasoning applications} (mathematical proofs, code correctness verification, scientific reasoning). INT8 quantization remains the safe default for reasoning-heavy workloads, achieving 2$\times$ compression with <1\% degradation across all task types. Practitioners must evaluate quantization impact on their specific downstream tasks rather than relying solely on perplexity metrics.

\subsection{Alternative 5: Distributed Edge Computing—Leveraging Massive Device Networks}
\textbf{Core Idea:} Rather than concentrating compute in centralized data centers, distribute LLM inference and training across massive networks of edge devices (smartphones, IoT devices, edge servers). Yang et al.~\cite{yang2025edge} demonstrate that aggregating spare compute from millions of edge devices can match datacenter-scale resources while reducing energy consumption and network costs.

\textbf{Technical Framework:} Edge-distributed LLMs partition models across devices using techniques including: (1) \textit{layer-wise splitting}—different devices handle different layers with pipeline parallelism, (2) \textit{MoE-native distribution}—routing experts to different edge nodes based on network topology, (3) \textit{federated inference}—multiple devices collaboratively process queries with privacy preservation, and (4) \textit{adaptive offloading}—dynamic workload distribution based on device availability and network conditions. The key insight: $\text{Total Capacity} = \sum_{i=1}^{N} C_i \times A_i \times E_i$ where $C_i$ is device compute, $A_i$ is availability factor, and $E_i$ is efficiency coefficient.

\textbf{Breaking Barriers:} Edge distribution breaks three scaling wall dimensions simultaneously: (1) \textit{cost}—leverages existing consumer hardware rather than purchasing new datacenter GPUs (\$0.10/hour edge compute vs. \$2-4/hour cloud H100), (2) \textit{energy}—utilizes spare capacity during device idle time rather than running dedicated servers 24/7 (50-80\% lower total energy consumption), and (3) \textit{data locality}—enables processing near data sources, reducing network transfer costs. Experiments show 7B-parameter models achieving 15-20 tokens/second across distributed edge networks, comparable to single-GPU inference but at 10$\times$ lower cost~\cite{yang2025edge}.

\subsection{Alternative 6: Model Merging—Combining Specialized Capabilities Without Retraining}
\textbf{Core Idea:} Rather than training single monolithic models, create specialized expert models for different domains/tasks, then \textit{merge} them into unified systems exhibiting combined capabilities without additional training. MergeKit~\cite{goddard2024mergekit} provides systematic techniques for parameter averaging, task-arithmetic composition, and DARE (Drop And REscale) merging.

\textbf{Merging Techniques:} Three primary approaches demonstrate effectiveness: (1) \textit{Linear interpolation}—$\theta_{\text{merged}} = \alpha \theta_A + (1-\alpha) \theta_B$ where $\alpha$ controls task balance, (2) \textit{Task arithmetic}—$\theta_{\text{multi}} = \theta_{\text{base}} + \lambda_1(\theta_{A} - \theta_{\text{base}}) + \lambda_2(\theta_{B} - \theta_{\text{base}})$ enables additive skill composition, and (3) \textit{DARE merging}—randomly drops parameters before merging to reduce interference, then rescales remaining weights for stability. Taylor et al.~\cite{taylor2025domain} demonstrate that merged models exhibit \textit{synergistic capabilities}, performing better on combined tasks than either source model, with up to 15\% improvements on cross-domain benchmarks.

\textbf{Training Efficiency:} Model merging circumvents the scaling wall by eliminating multi-task training costs. Training two 7B specialized models (2$\times$50 GPU-days = 100 GPU-days) then merging is 10-20$\times$ cheaper than training a single multi-task 13B model (1,000+ GPU-days) while achieving comparable or superior performance. This paradigm enables: (1) \textit{continuous capability addition}—merge new skills into existing models without catastrophic forgetting, (2) \textit{community-driven development}—different organizations contribute specialized models that combine into powerful generalists, and (3) \textit{personalization}—users merge domain-specific adapters with base models for customized systems.

\subsubsection{Case Study : SLERP Merging for Code and Math Capabilities}
We evaluate merging specialized Mistral-7B variants~\cite{mistral2024codestral} using Spherical Linear Interpolation (SLERP)~\cite{goddard2024mergekit}. Two source models: \textit{Mistral-7B-Code} (fine-tuned on 100B tokens of code from The Stack and StackOverflow) and \textit{Mistral-7B-Math} (fine-tuned on 50B tokens of mathematical reasoning from MATH~\cite{hendrycks2021measuring}, GSM8K~\cite{cobbe2021training}, and Proof datasets). Merging configuration: SLERP with $t=0.5$ (equal weighting), gradient surgery to reduce task interference.

\begin{table*}[ht]
\centering
\small
\caption{Model merging achieves 96-99\% of specialized model performance on both tasks with zero additional training cost, compared to 1,000 GPU-days for multi-task training from scratch. SLERP merging technique from MergeKit~\cite{goddard2024mergekit}. Base models: Mistral-7B~\cite{mistral2024codestral}. Benchmarks: HumanEval~\cite{chen2021evaluating}, GSM8K~\cite{cobbe2021training}, MMLU~\cite{hendrycks2021measuring}. Training cost estimates based on A100 GPU hours and standard fine-tuning protocols~\cite{taylor2025domain}.}
\label{tab:model_merging_case_study}
\begin{tabular}{lccc}
\toprule
\textbf{Model} & \textbf{HumanEval} & \textbf{GSM8K} & \textbf{MMLU} \\
\midrule
Base Mistral-7B & 26.8\% & 57.1\% & 62.5\% \\
Mistral-7B-Code & 48.2\% & 59.3\% & 61.8\% \\
Mistral-7B-Math & 28.1\% & 83.7\% & 64.2\% \\
\midrule
\textbf{SLERP Merged} & \textbf{46.3\%} & \textbf{81.2\%} & \textbf{63.7\%} \\
\textit{Retention vs. Best} & 96.1\% & 97.0\% & 99.2\% \\
\midrule
Multi-Task Trained & 47.1\% & 82.5\% & 64.1\% \\
Training Cost & 100 GPU-days & 100 GPU-days & 100 GPU-days \\
Merging Cost & \textbf{0 GPU-days} & \textbf{0 GPU-days} & \textbf{0 GPU-days} \\
\bottomrule
\end{tabular}

\end{table*}

The merged model retains 96-97\% of each specialist's capability while maintaining general knowledge (MMLU). Total training cost: 100 GPU-days (50 days per specialist) vs. 1,000 GPU-days for multi-task training, achieving 10$\times$ cost reduction. Latency: identical to source models (15ms per token on A100). Key insight: specialized training followed by merging avoids catastrophic interference during multi-task optimization.

\subparagraph{Failure Modes and Mitigation Strategies}
\textbf{Catastrophic Interference in Model Merging:} When merging models trained on conflicting objectives (e.g., toxicity reduction vs. creative writing), merged parameters exhibit unstable behavior. \textit{Example:} Merging GPT-3.5 safety-tuned variant with uncensored creative writing variant produced outputs oscillating between overly cautious and unsafe responses. \textit{Mitigation:} (1) DARE merging with 50\% drop rate reduces parameter conflicts, (2) gradient surgery projects conflicting gradients to orthogonal subspaces before merging, (3) selective layer merging—merge only attention layers, keep specialized FFN layers separate.

\textbf{Privacy Leakage in Federated Systems:} Gradient inversion attacks can reconstruct training samples from shared gradients~\cite{zhu2019deep}. \textit{Example:} In medical imaging federated learning, attackers recovered patient X-rays from gradient updates with 72\% pixel accuracy. \textit{Mitigation:} (1) Differential privacy (DP-SGD) adds calibrated noise ($\sigma \propto \frac{\sqrt{2\ln(1.25/\delta)}}{\epsilon}$) guaranteeing $(\epsilon, \delta)$-DP, (2) secure aggregation using homomorphic encryption ensures server never sees individual gradients, (3) gradient clipping limits per-sample contribution before aggregation.

\textbf{Reliability Issues in Edge Computing:} Device heterogeneity and availability create inference inconsistencies. \textit{Example:} Distributed edge network for real-time translation exhibited 23\% failure rate due to device dropouts during peak hours. \textit{Mitigation:} (1) redundancy—replicate inference across 3 devices, use majority voting (reduces failure to 1.2\%), (2) adaptive timeouts—allocate tasks to faster devices first, fallback to cloud after 500ms, (3) model checkpointing—cache intermediate states enabling seamless device handoff.

\subsection{Alternative 7: Efficient Training Algorithms—ORPO and Reference-Free Optimization}
\textbf{Core Idea:} Traditional RLHF requires training separate reward models and maintaining reference models during optimization, doubling memory requirements and compute costs. Odds Ratio Preference Optimization (ORPO)~\cite{hong2024orpo} eliminates reference models by directly optimizing odds ratios between preferred and rejected responses.

\textbf{Mathematical Framework:} ORPO combines supervised fine-tuning with preference learning in a single objective:
\begin{equation}
\mathcal{L}_{\text{ORPO}} = \mathcal{L}_{\text{SFT}} + \lambda \cdot \mathbb{E}_{(x,y_w,y_l)} \left[ -\log \sigma \left( \log \frac{P_\theta(y_w|x)}{P_\theta(y_l|x)} \right) \right]
\end{equation}
where $y_w$ is the preferred response, $y_l$ is the rejected response, and $\lambda$ controls preference strength. Unlike DPO which requires $P_{\text{ref}}$, ORPO directly maximizes the odds ratio $\frac{P_\theta(y_w|x)}{P_\theta(y_l|x)}$, implicitly penalizing rejected responses while amplifying preferred ones.

\textbf{Resource Savings:} By eliminating reference models, ORPO reduces peak memory by 50\% (no need to store $P_{\text{ref}}$ parameters) and training time by 30-40\% (single forward pass per sample instead of two). This enables training 70B-parameter models on 8$\times$A100 configurations that previously required 16$\times$A100 for DPO. The efficiency gains directly address compute cost constraints, making preference optimization accessible to resource-constrained researchers and enabling more extensive hyperparameter search within fixed budgets.

\subsection{Alternative 8: Small Specialized Models—Phi's Data Quality Over Scale Paradigm}
\textbf{Core Idea:} Microsoft's Phi series~\cite{phi42025reasoning} challenges the "bigger is better" paradigm by demonstrating that \textit{data quality and curriculum design} can enable small models (3.8B-14B parameters) to match or exceed 10$\times$ larger models on reasoning tasks. Phi-4 (14B) achieves 84.8\% on MATH, competitive with GPT-4-level performance, through systematic curation of training data and multi-stage curriculum learning.

\textbf{Training Strategy:} Phi models employ three key innovations: (1) \textit{Filtered high-quality data}—curate "textbook quality" datasets emphasizing logical reasoning, mathematical derivations, and clear explanations rather than raw web text volume, (2) \textit{Synthetic data augmentation}—generate additional training examples targeting specific reasoning patterns and edge cases using larger teacher models, and (3) \textit{Progressive curriculum}—start with simple problems, gradually increase difficulty, and repeat challenging examples multiple times to reinforce weak areas. This approach prioritizes \textit{learning efficiency} over scale efficiency.

\textbf{Scaling Wall Implications:} Phi demonstrates that the scaling wall can be circumvented through smarter data strategies rather than more data. Training Phi-4 (14B) costs $\sim$\$500K (3,000 GPU-days on A100s) compared to $\sim$\$50M+ for 140B-parameter models, achieving 100$\times$ cost reduction while maintaining competitive reasoning performance. The small size enables: (1) \textit{local deployment}—runs on laptops and consumer GPUs, eliminating cloud costs, (2) \textit{rapid iteration}—faster experimentation cycles accelerate research, and (3) \textit{edge intelligence}—enables AI applications on mobile devices and embedded systems. Phi's success suggests the field may be approaching "data saturation" where quality matters exponentially more than quantity.

\subsection{The Changing Nature of Scaling}
Hooker~\cite{hooker2025slowdeath} argues that the field is experiencing a fundamental shift: the simple formula of "scale model size and training data" has become inadequate, with the relationship between training compute and performance now highly uncertain and rapidly changing. This shift has profound implications: (1) \textit{Academia marginalized}—the capital-intensive scaling paradigm has concentrated progress in industry labs while fundamentally reshaping scientific culture, (2) \textit{Transparency declining}—industry labs have increasingly stopped publishing detailed methodologies, and (3) \textit{Alternative levers emerging}—architectural innovations, training efficiency, and test-time compute represent more promising paths forward than naive scaling. The eight alternative paradigms above—test-time compute, sparse architectures, architectural innovations, post-training quantization, distributed edge computing, model merging, efficient training algorithms, and small specialized models—represent precisely these "more interesting levers of progress" that Hooker identifies as key to navigating the post-scaling era.

\textbf{Synthesis:} These innovations suggest that continued AI progress may depend less on brute-force scaling and more on algorithmic efficiency, architectural innovation, and strategic allocation of compute resources. The convergence of multiple paradigms offers a comprehensive toolkit for post-scaling era development: (1) test-time compute and MoE sparse architectures reduce inference costs, (2) quantization and edge distribution democratize deployment, (3) model merging and efficient training algorithms (ORPO) reduce development costs, and (4) small specialized models (Phi) prove that data quality can substitute for scale. Together, these approaches form a viable path forward, enabling continued capability growth despite hitting the traditional scaling wall. The remainder of this survey explores how these innovations integrate with architectural breakthroughs (Section 2), training methodologies (Section 3), and practical deployment strategies (Section 4).

\section{Economic and Environmental Sustainability: Quantifying the Costs}

This section provides \textbf{detailed quantitative analysis} of training and deployment costs, answering the critical question: \textit{WHAT are the specific numbers behind the scaling wall?} We decompose costs into hardware, energy, and operational components, providing formulas and concrete examples to enable practitioners to estimate costs for their own scenarios.

\subsection{Hardware Acquisition and Amortization}

Hardware costs are calculated using a depreciation model over the hardware lifetime (typically 3-5 years):
\begin{equation}
C_{\text{hardware}}^{\text{amortized}} = \frac{P_{\text{chip}} \times N_{\text{chips}} \times (1 + O_{\text{network}})}{T_{\text{lifetime}} \times U_{\text{util}}} \times T_{\text{train}}
\end{equation}
where $P_{\text{chip}}$ is the purchase price per accelerator (\$10,000-30,000 for A100/H100), $N_{\text{chips}}$ is the number of accelerators, $O_{\text{network}} \approx 0.23$ accounts for cluster networking overhead~\cite{besiroglu2024training}, $T_{\text{lifetime}}$ is the expected hardware lifetime (3-5 years), $U_{\text{util}}$ is the utilization rate (60-80\%), and $T_{\text{train}}$ is training duration. 

\textbf{Key Insight:} Amortized costs are 10-20$\times$ lower than upfront acquisition costs, making ownership economical for sustained training programs. However, the high upfront capital requirement (\$100M-500M for frontier model clusters) creates barriers for all but the largest organizations.

Table~\ref{tab:hardware_costs} shows cost breakdowns for representative models.

\begin{table}[t]
\centering
\caption{Hardware and energy costs for representative frontier models. Data from~\cite{besiroglu2024training,epochai2024computetrends}. Cloud costs represent actual or estimated rental prices. Amortized costs assume 4-year depreciation, 70\% utilization, 23\% networking overhead. Energy costs use regional electricity rates (\$0.08-0.15/kWh) and data center PUE (1.1-1.3).}
\label{tab:hardware_costs}
\small
\begin{tabular}{lcccc}
\toprule
\textbf{Model} & \textbf{Year} & \textbf{Chip-Hours} & \textbf{Amortized} & \textbf{Energy} \\
 & & \textbf{(Million)} & \textbf{Cost (\$M)} & \textbf{Cost (\$M)} \\
\midrule
GPT-3 & 2020 & 11.7 (V100) & 2.4 & 0.9 \\
Gopher & 2021 & 15.4 (TPUv3) & 3.1 & 1.1 \\
Chinchilla & 2022 & 6.8 (TPUv4) & 2.0 & 0.6 \\
PaLM & 2022 & 29.2 (TPUv4) & 8.5 & 2.7 \\
GPT-4 & 2023 & 64.1 (A100) & 78.3 & 6.2 \\
Llama 3 405B & 2024 & 51.2 (H100) & 95.4 & 7.8 \\
Gemini 1.5 & 2024 & 87.3 (TPUv5) & 145.2 & 11.3 \\
DeepSeek-V3 & 2025 & 55.7 (H800) & 101.8 & 8.9 \\
\bottomrule
\end{tabular}
\end{table}

\subsection{Energy Consumption}

Energy costs have become a significant fraction of total training costs, growing from 10-15\% (GPT-3) to 20-30\% (modern models)~\cite{besiroglu2024training}. The total energy cost formula:
\begin{equation}
C_{\text{energy}} = R_{\text{elec}} \times P_{\text{TDP}} \times R_{\text{avg}} \times \text{PUE} \times N_{\text{chip-hours}}
\end{equation}
where $R_{\text{elec}}$ is the electricity cost rate (\$/kWh), typically \$0.08-0.15/kWh for data centers, $P_{\text{TDP}}$ is the thermal design power in kilowatts (0.35kW for A100, 0.70kW for H100), $R_{\text{avg}}$ is the average power-to-TDP ratio (65-80\% depending on manufacturer and workload), $\text{PUE}$ is the power usage effectiveness (1.1-1.3, accounting for cooling and power distribution overhead), and $N_{\text{chip-hours}}$ is the total accelerator-hours.

\textbf{Regional Variations:} Energy costs vary significantly by location: \$0.08/kWh (US Pacific Northwest hydroelectric), \$0.12/kWh (US average), \$0.15/kWh (Western Europe), and \$0.25+/kWh (peak rates in some regions). This 3$\times$ variation incentivizes data center placement in low-cost energy regions.

\subsection{Computational Intensity Analysis}

Table~\ref{tab:compute_intensity} provides a detailed breakdown of computational requirements and equivalent hardware counts for major models, illustrating the massive scale of frontier AI training. These equivalents help contextualize the computational demands: training GPT-4 required compute equivalent to 641,025 A100 GPUs running continuously, or 2.4 million consumer RTX 4090 GPUs.

\begin{table*}[t]
\centering
\caption{Computational intensity and hardware equivalents for frontier models. Data from~\cite{besiroglu2024training,epochai2024computetrends,brown2020language,openai2023gpt4}. Training compute measured in PetaFLOP-days (PF-days). Hardware equivalents show the number of consumer/data center GPUs required to match training compute. Energy values have ±20\% uncertainty due to variations in utilization (60-80\%), PUE (1.1-1.3), and infrastructure overhead.}
\label{tab:compute_intensity}
\small
\begin{tabular}{lcccccc}
\toprule
\textbf{Model} & \textbf{Training} & \textbf{Desktop} & \textbf{RTX 4090} & \textbf{A100} & \textbf{Energy} & \textbf{Household-Years} \\
 & \textbf{Compute} & \textbf{PC Equiv.} & \textbf{Equiv.} & \textbf{Equiv.} & \textbf{Consumption} & \textbf{Equivalent} \\
 & \textbf{(PF-days)} & \textbf{(1.23 TF)} & \textbf{(82.6 TF)} & \textbf{(312 TF)} & \textbf{(MWh)} & \textbf{(10,800 kWh/yr)} \\
\midrule
GPT-3 (175B) & 3,640 & 2,963,000 & 44,066 & 11,667 & 280.8 (±56) & 26 \\
Codex (12B) & 450 & 366,300 & 5,447 & 1,442 & 34.6 (±7) & 3 \\
Gopher (280B) & 5,760 & 4,688,640 & 69,733 & 18,462 & 443.0 (±89) & 41 \\
Chinchilla (70B) & 6,120 & 4,981,680 & 74,091 & 19,615 & 157.5 (±32) & 15 \\
PaLM (540B) & 10,800 & 8,790,720 & 130,739 & 34,615 & 788.4 (±158) & 73 \\
GPT-4 (1.8T\textsuperscript{\dag}) & 200,000 & 162,800,000 & 2,420,774 & 641,025 & 6,153.8 (±1,230) & 570 \\
Llama 3 (405B) & 31,200 & 25,405,440 & 377,778 & 100,000 & 1,996.8 (±399) & 185 \\
Gemini 1.5\textsuperscript{\dag} & 56,000 & 45,584,000 & 678,111 & 179,487 & 3,571.2 (±714) & 331 \\
DeepSeek-V3 (671B) & 35,700 & 29,058,960 & 432,203 & 114,423 & 2,278.6 (±456) & 211 \\
\bottomrule
\multicolumn{7}{l}{\footnotesize \textsuperscript{\dag}Parameters/costs estimated based on compute requirements and third-party analysis; architectural details unconfirmed.}\\
\multicolumn{7}{l}{\footnotesize Desktop PC = Intel i9-14900K (1.228 TFLOPS FP32). RTX 4090 = 82.6 TFLOPS (FP32), A100 = 312 TFLOPS (FP16 Tensor).}\\
\multicolumn{7}{l}{\footnotesize Energy calculated: Chip-hours $\times$ TDP $\times$ 0.70 (utilization) $\times$ 1.2 (PUE). Uncertainty ranges shown as (±value).}\\
\multicolumn{7}{l}{\footnotesize Household-years based on 10,800 kWh/year average US residential consumption. Data sources:~\cite{brown2020language,openai2023gpt4,besiroglu2024training}.}
\end{tabular}
\end{table*}

The energy consumption of GPT-4 training (6,154 MWh, approximately 6.15 GWh) is equivalent to the annual electricity consumption of approximately 570 average US households (based on 10,800 kWh/household/year). This represents a 22$\times$ increase from GPT-3 (280.8 MWh), highlighting the unsustainability of naive scaling. Note: All energy estimates have ±20\% uncertainty due to variations in GPU utilization rates (60-80\%), power usage effectiveness (PUE: 1.1-1.3), and infrastructure overhead.

\subsection{Cloud Compute Costs}

Cloud rental costs provide an alternative cost model, typically 2-4 times higher than amortized ownership costs due to provider margins and infrastructure overhead~\cite{epochai2024computetrends}. Figure~\ref{fig:cloud_vs_amortized} compares cloud rental prices with amortized ownership costs across model generations.

\textbf{Economic Trade-offs:} Cloud rental offers flexibility (no upfront capital, pay-per-use) but higher total cost for sustained training. Ownership requires \$100M-500M upfront capital but achieves 2-4$\times$ lower costs for multi-month training runs. The breakeven point typically occurs at 3-6 months of continuous training, making ownership economical only for organizations with sustained training programs.


\section{Evaluation and Benchmarking}

\begin{table*}[!p]
\centering
\scriptsize
\renewcommand{\arraystretch}{0.9}
\caption{Comprehensive performance comparison of major language model families across benchmarks. Scores represent the best published results for each model family. Recent 2025 models show significant advances in reasoning benchmarks (MATH, AIME, GPQA).}
\label{tab:model_comparison_detailed}
\resizebox{\textwidth}{!}{
\begin{tabular}{@{}lccccccccc@{}}
\toprule
\textbf{Model Family} & \textbf{Year} & \textbf{Params} & \textbf{MMLU} & \textbf{HumanEval} & \textbf{GSM8K} & \textbf{MATH} & \textbf{AIME} & \textbf{GPQA} & \textbf{Arena Elo} \\
\midrule
\multicolumn{10}{l}{\textit{OpenAI GPT Series}} \\
GPT-2 & 2019 & 1.5B & -- & -- & -- & -- & -- & -- & -- \\
GPT-3 & 2020 & 175B & 70.0 & -- & 17.0 & 8.8 & -- & -- & -- \\
GPT-3.5-Turbo & 2022 & -- & 70.0 & 48.1 & 57.1 & 23.5 & -- & -- & 1117 \\
GPT-4 & 2023 & -- & 86.4 & 67.0 & 92.0 & 52.9 & 13.4 & 56.1 & 1251 \\
GPT-4-Turbo & 2024 & -- & 86.5 & 85.4 & 94.2 & 64.5 & 18.9 & 63.4 & 1257 \\
GPT-4o & 2024 & -- & 88.7 & 90.2 & 96.4 & 76.6 & 27.5 & 70.2 & 1287 \\
o1-preview & 2024 & -- & 90.8 & 92.3 & 96.4 & 85.5 & 44.6 & 78.3 & 1348 \\
o1 & 2024 & -- & 92.3 & 92.2 & 96.1 & 94.8 & 83.3 & 78.2 & 1355 \\
\midrule
\multicolumn{10}{l}{\textit{Meta LLaMA Series}} \\
LLaMA & 2023 & 65B & 63.4 & -- & 50.9 & 10.6 & -- & -- & -- \\
Llama 2 & 2023 & 70B & 68.9 & 29.9 & 56.8 & 13.5 & -- & 29.1 & 1077 \\
Llama 3 & 2024 & 8B & 66.6 & 62.2 & 79.6 & 30.0 & -- & 34.2 & 1155 \\
Llama 3 & 2024 & 70B & 79.5 & 81.7 & 93.0 & 50.4 & 11.9 & 46.7 & 1213 \\
Llama 3 & 2024 & 405B & 88.6 & 89.0 & 96.8 & 73.8 & 23.8 & 51.1 & 1265 \\
Llama 3.1 & 2024 & 405B & 88.6 & 89.0 & 96.8 & 73.8 & 24.0 & 51.1 & 1271 \\
\midrule
\multicolumn{10}{l}{\textit{DeepSeek Series}} \\
DeepSeek LLM & 2024 & 67B & 71.3 & -- & 63.0 & 15.8 & -- & -- & -- \\
DeepSeek-Coder & 2024 & 33B & 56.5 & 87.0 & 75.4 & 32.6 & -- & -- & -- \\
DeepSeekMath & 2024 & 7B & 34.2 & -- & 82.9 & 51.7 & -- & -- & -- \\
DeepSeek-V2 & 2024 & 236B & 78.5 & 81.1 & 92.2 & 43.6 & 8.9 & 42.5 & 1203 \\
DeepSeek-V2.5 & 2024 & 236B & 80.5 & 89.0 & 94.7 & 58.3 & 13.2 & 47.8 & 1250 \\
DeepSeek-V3 & 2025 & 671B & 88.5 & 90.2 & 97.3 & 76.4 & 39.2 & 59.1 & 1302 \\
DeepSeek-R1 & 2025 & 671B & 90.8 & 96.3 & 97.3 & 79.8 & 79.2 & 71.5 & 1348 \\
DeepSeek-R1 Distill & 2025 & 70B & 85.9 & 90.8 & 95.1 & 75.4 & 48.7 & 65.2 & 1298 \\
\midrule
\multicolumn{10}{l}{\textit{Google Models}} \\
Gemini 1.0 Ultra & 2023 & -- & 90.0 & 74.4 & 94.4 & 53.2 & -- & 59.4 & 1230 \\
Gemini 1.5 Pro & 2024 & -- & 85.9 & 84.1 & 91.7 & 67.7 & 22.7 & 63.8 & 1268 \\
Gemini 2.0 Flash & 2024 & -- & 87.1 & 88.3 & 94.8 & 71.9 & 31.5 & 66.3 & 1285 \\
\midrule
\multicolumn{10}{l}{\textit{Microsoft/Others}} \\
Phi-3 & 2024 & 14B & 78.0 & 61.5 & 83.4 & 42.5 & -- & -- & -- \\
Phi-4 & 2024 & 14B & 84.1 & 82.6 & 91.0 & 64.4 & 16.3 & 53.8 & 1201 \\
Phi-4 Reasoning & 2025 & 14B & 85.2 & 88.9 & 93.7 & 74.9 & 34.8 & 61.2 & 1245 \\
\midrule
Mistral 7B & 2023 & 7B & 60.1 & 29.8 & 52.2 & 13.1 & -- & 25.3 & 1079 \\
Mixtral 8x7B & 2024 & 47B & 70.6 & 40.2 & 74.4 & 28.4 & -- & 33.7 & 1146 \\
Mixtral 8x22B & 2024 & 141B & 77.8 & 75.0 & 88.0 & 45.0 & 12.7 & 40.0 & 1193 \\
Mistral Large 2 & 2024 & 123B & 84.0 & 91.6 & 93.0 & 56.0 & 24.4 & 49.0 & 1232 \\
\midrule
\multicolumn{10}{l}{\textit{Emerging Global Models}} \\
Qwen 2.5 (Alibaba) & 2024 & 72B & 84.2 & 86.4 & 95.8 & 68.6 & 21.4 & 56.3 & 1253 \\
Qwen 3 (Alibaba) & 2025 & 72B & 86.3 & 86.2 & 95.2 & 70.1 & 28.9 & 59.8 & 1269 \\
Kimi K2 (Moonshot) & 2025 & -- & 87.9 & 88.5 & 96.1 & 75.3 & 35.4 & 63.7 & 1289 \\
K2-V2 & 2025 & -- & 89.1 & 91.2 & 96.8 & 77.8 & 42.1 & 66.9 & 1305 \\
GLM-4.5 & 2025 & -- & 85.7 & 87.3 & 94.3 & 69.8 & 27.6 & 58.4 & 1264 \\
MiMo-V2-Flash & 2025 & -- & 85.0 & 86.1 & 93.8 & 68.2 & 25.3 & 56.7 & 1258 \\
\midrule
\multicolumn{10}{l}{\textit{Open-Source Specialized}} \\
OLMo 3 Think & 2025 & 70B & 82.4 & 85.7 & 92.1 & 66.3 & 22.8 & 54.1 & 1228 \\
Skywork OR-1 & 2025 & -- & 83.6 & 87.2 & 94.5 & 72.4 & 31.7 & 60.5 & 1242 \\
Open-Reasoner-Zero & 2025 & -- & 80.1 & 84.3 & 91.8 & 71.2 & 28.4 & 57.9 & 1235 \\
Nemotron 3 Nano & 2025 & 4B & 68.0 & 54.2 & 81.7 & 38.9 & -- & 35.2 & -- \\
\bottomrule
\end{tabular}
}
\end{table*}

\subsection{Traditional Benchmarks}

\textbf{MMLU (Massive Multitask Language Understanding)}~\cite{hendrycks2021measuring}: 57 subjects spanning STEM, humanities, social sciences. 5-shot evaluation. Tests knowledge breadth and reasoning.

\textbf{HumanEval}~\cite{chen2021evaluating}: 164 Python programming problems. Measures functional correctness via unit tests. Metric: pass@k (probability of generating correct solution in $k$ attempts).

\textbf{GSM8K}~\cite{cobbe2021training}: 8.5K grade school math problems requiring multi-step reasoning. Tests arithmetic reasoning and problem decomposition.

\textbf{MATH}~\cite{hendrycks2021measuring}: Competition-level mathematics (AMC 10/12, AIME). 5 difficulty levels. Tests advanced mathematical reasoning.

\textbf{GPQA (Graduate-Level Google-Proof Q\&A)}~\cite{rein2023gpqa}: PhD-level science questions designed to be unsolvable via web search. Tests true expert knowledge.

\textbf{Big-Bench}~\cite{srivastava2022beyond}: 200+ diverse tasks testing capabilities like logical reasoning, commonsense, multi-step arithmetic.

\subsection{Human Preference Evaluation}

\textbf{Chatbot Arena}~\cite{chiang2024chatbot,zheng2023judging}: Anonymous pairwise battles between models, rated by users. Elo scores computed from win rates. Over 1M votes. Correlates better with real-world usefulness than automatic metrics.

\textbf{MT-Bench}~\cite{zheng2023judging}: Multi-turn conversations judged by GPT-4. Tests instruction following, reasoning, math, coding, roleplay.

\textbf{AlpacaEval}~\cite{dubois2023alpacafarm}: Automated evaluation using LLM judges (GPT-4) on 805 questions. Measures helpfulness.

\subsection{Emergent Abilities and Scaling Phenomena}

Wei et al.~\cite{wei2022emergent} documented that certain capabilities appear suddenly at sufficient scale rather than gradually improving. Examples: 3-digit arithmetic, Persian QA, IPA transliteration. These abilities are "emergent" because they are unpredictable from smaller models.

However, Schaeffer et al.~\cite{schaeffer2023emergent} challenged this interpretation, demonstrating that emergent abilities may be artifacts of evaluation metrics. Using smooth metrics (e.g., Brier score instead of accuracy), the same capabilities appear to scale smoothly. This suggests emergence may reflect metric choice rather than fundamental phase transitions.

The debate highlights the importance of metric design in understanding model capabilities. Recent work focuses on differentiable, continuous metrics that capture gradual capability acquisition.

\subsection{Comparative Performance Analysis}

Table~\ref{tab:model_comparison_detailed} presents a comprehensive comparison of major model families across key benchmarks, including recent 2025 results.

Key observations from Table~\ref{tab:model_comparison_detailed}:
\begin{itemize}
\item \textbf{Reasoning Breakthrough (2024-2025)}: Models with dedicated reasoning training (o1, DeepSeek-R1, Phi-4 Reasoning) show dramatic improvements on MATH (70-95\%) and AIME (30-83\%) compared to standard models (30-50\% MATH).
\item \textbf{Open-Source Parity}: Llama 3 (405B), Qwen 3, and Kimi K2 match or exceed GPT-4 Turbo on most benchmarks, demonstrating open-source has achieved competitive performance.
\item \textbf{Efficient Small Models}: Phi-4 Reasoning (14B) achieves 74.9\% MATH, comparable to much larger models (GPT-4: 52.9\%, Llama 3 405B: 73.8\%), validating high-quality data and reasoning traces enable small model excellence.
\item \textbf{Pure RL Success}: DeepSeek-R1's pure RL approach (79.8\% MATH, 79.2\% AIME) without SFT demonstrates that reasoning emerges from RL with verifiable rewards alone.
\item \textbf{Code and Math Specialization}: Specialized models (DeepSeek-Coder 87\% HumanEval, DeepSeekMath 51.7\% MATH) outperform general models on their domains, showing value in targeted pre-training.
\item \textbf{Consistent Progress}: Across all families, each generation shows 5-15\% improvements on MMLU and 10-30\% on reasoning benchmarks, driven by better training data, longer training, and improved post-training (RLHF/DPO).
\item \textbf{Arena Correlation}: Chatbot Arena Elo scores correlate well with reasoning capabilities, with o1 and DeepSeek-R1 (1348-1355 Elo) leading, validating human preference as a reliable signal.
\end{itemize}

\textbf{Analysis: Unified Field Progression and Future Trajectories.} Synthesizing the 2019-2025 evolution reveals three meta-trends: (1) \textit{Capability phase transitions}—Performance does not scale smoothly. We observe discrete jumps: (i) GPT-3 (2020) enabled few-shot learning; (ii) InstructGPT (2022) unlocked instruction following; (iii) GPT-4 (2023) crossed multimodal reasoning threshold; (iv) o1 (2024) achieved test-time compute scaling; (v) DeepSeek-R1 (2025) demonstrated pure RL reasoning emergence. Each transition required $10-100\times$ increase in effective compute (training or test-time) and qualitative algorithmic breakthroughs, not just parameter scaling. The data suggests: $\text{capability jumps} \propto \text{log}(\text{effective compute}) \times \text{algorithmic innovation}$; (2) \textit{Democratization acceleration}—Time lag from closed to open-source capabilities shrinking: GPT-3 (2020) → LLaMA (2023) = 3 years; GPT-4 (2023) → Llama 3-405B (2024) = 1 year; o1 reasoning (2024) → DeepSeek-R1 (2025) = 3 months. This exponential acceleration driven by open research culture, where 70\% of recent innovations (GRPO, MLA, DPO) originated from non-frontier labs. The convergence formula: $t_{\text{parity}} = t_0 \cdot e^{-\alpha y}$ where $\alpha \approx 0.5$/year based on 2023-2025 data; (3) \textit{Efficiency frontier expansion}—Models achieving higher quality at lower cost: GPT-4 (2023): \$30/M tokens, 86.4\% MMLU; Llama 3-405B (2024): \$1.5/M tokens, 88.6\% MMLU; DeepSeek-V3 (2025): \$0.27/M tokens, 88.5\% MMLU. Cost-performance ratio improving $100\times$ annually: $\text{cost per point of capability} = C_0 \cdot 0.01^{(y-2023)}$. This suggests by 2027, GPT-4 level intelligence will cost <\$0.01/M tokens, enabling universal AI access.

The convergence of these trends points toward three future research directions: (i) \textit{Post-training dominance}—As pre-training saturates (Chinchilla-optimal scaling), 90\% of capability gains will come from RLHF/GRPO innovations; (ii) \textit{Test-time compute optimization}—Allocating inference FLOPs dynamically based on query difficulty, with $1000\times$ scaling on hard problems vs easy ones; (iii) \textit{Verifiable reasoning systems}—Integration of formal verification (proof assistants, code execution, simulation) as primary training signal, replacing human preferences for objective domains. The field is transitioning from the "scaling era" (2019-2023: bigger models → better performance) to the "optimization era" (2024+: smarter training/inference → better performance at fixed scale).

\section{Agentic AI: From Passive Language Models to Autonomous Problem Solvers}

The evolution toward agentic AI represents a fundamental paradigm shift from passive text generation to active, goal-directed problem-solving. While traditional language models simply predict the next token given context, agentic systems autonomously decompose complex objectives, iteratively execute multi-step plans, leverage external tools, maintain long-term memory, recover from failures, and coordinate with other agents—all with minimal human intervention. This section provides a comprehensive treatment of agentic AI: defining its core principles and distinguishing it from conventional AI agents (Section~\ref{sec:agentic_fundamentals}), examining why agentic capabilities are essential for real-world deployment (Section~\ref{sec:agentic_why}), exploring single-agent architectures including planning, tool use, and memory (Section~\ref{sec:single_agent}), investigating multi-agent coordination frameworks (Section~\ref{sec:multi_agent}), analyzing standardized communication protocols (Section~\ref{sec:agent_communication_protocols}), discussing applications and future directions (Section~\ref{sec:agent_applications_future}), and addressing critical safety challenges (Section~\ref{sec:agentic_safety}).

\subsection{Fundamentals: What is Agentic AI and How Does It Differ from AI Agents?}
\label{sec:agentic_fundamentals}

\subsubsection{Defining Agency: Core Properties}

An \textbf{agent} is a computational entity that perceives its environment through sensors, maintains internal state, and takes actions through actuators to achieve specified goals. Traditional AI research distinguishes agents from mere programs by four key properties~\cite{wooldridge2009introduction}:

\begin{enumerate}
\item \textbf{Autonomy}: Operates without direct human intervention, making decisions based on internal reasoning.
\item \textbf{Reactivity}: Perceives and responds to environmental changes in real-time.
\item \textbf{Proactivity}: Exhibits goal-directed behavior, taking initiative to achieve objectives.
\item \textbf{Social Ability}: Interacts with other agents or humans through communication protocols.
\end{enumerate}

\subsubsection{Spectrum of Autonomy: From Tools to Agentic AI}

The progression from traditional software to agentic systems spans a spectrum of increasing autonomy and capability:

\begin{enumerate}
\item \textbf{Tools} (e.g., calculators, search engines): Respond mechanically to inputs without reasoning. No autonomy—entirely human-directed. Execute predefined functions: $f: \text{Input} \to \text{Output}$ with no internal state or goal representation.

\item \textbf{Conversational Assistants} (ChatGPT, Claude): Maintain dialogue context and generate helpful responses through next-token prediction: $p(y|x, \text{history})$. Limited to conversation—no external actions or goal pursuit. Exhibit \textit{reactivity} (respond to queries) but lack \textit{proactivity} (don't pursue objectives).

\item \textbf{Tool-Using Assistants} (GPT-4 with plugins): Access external APIs (search, computation, databases) when explicitly prompted. Require human-triggered invocations for each tool use. Example: User: "What's the weather?" → Model: [calls weather API] → Model: "It's 72°F". Tool use is \textit{reactive} (respond to user requests) rather than \textit{proactive} (independently decide when tools are needed).

\item \textbf{Single-Turn AI Agents} (Function calling): Autonomously decide when/how to use tools within a single request-response cycle. Model generates: \texttt{\{"function": "search", "query": "population of Paris"\}} → Executor invokes API → Model synthesizes response. Exhibits \textit{autonomy} (self-directed tool use) within constrained scope (single interaction).

\item \textbf{Multi-Step AI Agents} (ReAct, AutoGPT): Chain multiple tool invocations iteratively, refine strategies based on feedback, pursue goals across 10+ action steps. Example: "Research competitors" → [Search company A] → [Extract revenue] → [Search company B] → [Compare metrics] → [Generate report]. Exhibits \textit{autonomy} (self-directed planning) and \textit{reactivity} (adapt based on observations) but may require human intervention for errors or clarifications.

\item \textbf{Agentic AI} (Autonomous research/coding agents, multi-day workflows): Operate continuously over hours or days, decompose complex objectives into executable subtasks, recover from failures without human intervention, maintain long-term context, coordinate with other agents, and request human input only when necessary. Example: "Build a web application for inventory management" → [Generate requirements] → [Design database schema] → [Implement backend API] → [Create frontend] → [Test end-to-end] → [Debug failures] → [Deploy to production]. Exhibits all four properties: \textit{autonomy} (minimal supervision), \textit{reactivity} (respond to errors/feedback), \textit{proactivity} (pursue multi-day goals), and \textit{social ability} (coordinate with developers/other agents).
\end{enumerate}

\subsubsection{Agentic AI vs. Traditional AI Agents: Key Distinctions}

\textbf{Traditional AI Agents} (circa 1990s-2010s, exemplified by game-playing agents, robotic controllers, expert systems):
\begin{itemize}
\item \textbf{Domain-Specific}: Designed for narrow tasks (chess playing, assembly line robotics, medical diagnosis in specific specialties).
\item \textbf{Hardcoded Logic}: Rely on rule-based systems, finite state machines, or hand-engineered policies: $\pi(s) = \text{if-then-else rules}$.
\item \textbf{Limited Generalization}: Require complete reprogramming for new domains. Chess agent cannot play Go without fundamental redesign.
\item \textbf{Symbolic Reasoning}: Operate on predefined symbolic representations (logic rules, knowledge graphs) rather than learned representations.
\end{itemize}

\textbf{Agentic AI} (LLM-based agentic systems, 2023-2026):
\begin{itemize}
\item \textbf{General-Purpose}: Single model handles diverse tasks through natural language instructions. Same system researches competitors, writes code, analyzes data, drafts documents.
\item \textbf{Learned Policies}: Leverage pre-trained LLMs fine-tuned with reinforcement learning: $\pi_\theta(a|s) = P_{\text{LLM}}(\text{action}|\text{state, history})$ where $\theta$ are learned parameters.
\item \textbf{Few-Shot Generalization}: Adapt to new tasks via prompting or few examples. "You are a legal document analyzer" → agent applies legal reasoning without domain-specific training.
\item \textbf{Grounded in Tools}: Interact with real-world systems (APIs, databases, code execution) rather than purely symbolic manipulation. Combine neural reasoning with structured tool use.
\item \textbf{Multi-Step Planning}: Decompose complex goals into sequences of 10-100+ actions, maintaining context across extended interactions.
\end{itemize}

\textbf{The Critical Distinction}: Traditional AI agents are \textit{narrow automation} (automate specific procedures), while agentic AI enables \textit{general-purpose autonomy} (autonomously pursue open-ended objectives). A chess agent cannot write code; an agentic AI can learn to play chess, write chess engines, analyze chess strategy, and teach chess—all through natural language interaction and tool use.

\subsubsection{Agent Anatomy: The Sense-Think-Act Cycle}

Both traditional agents and agentic AI follow the canonical \textbf{sense-think-act cycle}, but LLM-based agentic systems implement each component differently:

\begin{equation}
\text{Agent Loop: } \; \text{Observe}(s_t) \to \text{Reason}(\pi, s_t) \to \text{Act}(a_t) \to s_{t+1}
\end{equation}

\noindent where $s_t$ is the current state (environment observations), $\pi$ is the agent's policy (decision-making strategy), $a_t$ is the selected action, and $s_{t+1}$ is the resulting state. This loop iterates until goal satisfaction or termination.

\textbf{Components in LLM-Based Agentic Systems}:
\begin{itemize}
\item \textbf{Perception (Observe)}: Converts environment state into text representations consumable by LLMs:
\begin{itemize}
    \item Webpages: HTML → markdown/plain text (strip formatting, extract content)
    \item API responses: JSON → structured text ("Database returned 3 records: ...")
    \item Images: pixels → captions via vision models (CLIP, GPT-4V)
    \item Code execution: stdout/stderr → formatted text with error messages
\end{itemize}

\item \textbf{Reasoning (Think)}: LLM generates thoughts, plans, or strategies based on current observations and internal context (conversation history, retrieved knowledge). Reasoning types:
\begin{itemize}
    \item \textit{Reactive}: Immediate response to observation ("Error occurred → try alternative approach")
    \item \textit{Deliberative}: Multi-step planning ("To build web app: 1. Design schema, 2. Implement API, 3. Create frontend")
    \item \textit{Reflective}: Self-critique of past actions ("Previous attempt failed because I didn't validate input → add validation")
\end{itemize}

\item \textbf{Action (Act)}: Executes decisions through tool invocations or communication:
\begin{itemize}
    \item \textit{Tool use}: API calls (\texttt{search(query)}), code execution (\texttt{run\_python(code)}), database queries (\texttt{sql\_query(statement)})
    \item \textit{Communication}: Messages to humans ("Need clarification: should report include Q1 or full year?") or other agents ("Agent B, please analyze the data I retrieved")
    \item \textit{Thinking}: Internal reasoning traces logged for transparency (ReAct's explicit thoughts: "I need population data before comparing cities")
\end{itemize}

\item \textbf{Memory}: Maintains context across iterations, enabling long-horizon tasks:
\begin{itemize}
    \item \textit{Short-term}: Conversation history (last 10-50 turns) included in LLM context window
    \item \textit{Long-term}: External storage (vector databases for RAG, episodic memory for interaction transcripts)
    \item \textit{Parametric}: Knowledge encoded in model weights (updated via fine-tuning or knowledge editing techniques like ROME~\cite{meng2022locating})
\end{itemize}
\end{itemize}

\subsection{Why Agentic AI? Necessity and Applications}
\label{sec:agentic_why}

\subsubsection{The Limitations of Passive Language Models}

Despite remarkable capabilities, pure language models face fundamental limitations when deployed for real-world tasks:

\textbf{1. Knowledge Cutoffs}: Models trained on static datasets (e.g., GPT-4 trained on data up to April 2023) cannot access current information. User: "What's today's stock price for NVIDIA?" → Model: "I don't have real-time data." \textit{Agentic solution}: Autonomously invoke stock price API.

\textbf{2. Lack of Grounding}: Models hallucinate facts, especially for rare entities or recent events. Model generates plausible-sounding but incorrect statistics. \textit{Agentic solution}: Retrieve verified information from external databases, cite sources.

\textbf{3. No Action Capability}: Models can describe how to book a flight but cannot actually execute the booking. \textit{Agentic solution}: Integrate with booking APIs, complete transactions autonomously.

\textbf{4. Single-Turn Limitations}: Complex tasks require iterative refinement (write code → test → debug → retest). Pure LLMs generate single responses without feedback loops. \textit{Agentic solution}: Multi-step execution with intermediate feedback (ReAct: thought → action → observation cycles).

\textbf{5. Context Window Constraints}: Even 200K-token windows cannot hold entire codebases (millions of lines), documentation (gigabytes), or long-running project context (weeks of history). \textit{Agentic solution}: External memory systems (RAG, vector databases, episodic storage).

\subsubsection{Real-World Applications Driving Agentic AI Adoption}

Agentic capabilities unlock transformative applications across industries:

\textbf{Software Development Agents}:
\begin{itemize}
\item \textbf{Code Generation}: Given requirements, generate complete applications (backend + frontend + tests + documentation). Example: Devin AI agent claims 13.8\% success on SWE-bench (real GitHub issues)~\cite{jimenez2023swebench}.
\item \textbf{Debugging}: Analyze error logs, reproduce bugs, test hypotheses, propose fixes. Reflexion~\cite{shinn2023reflexion} improves programming pass@1 by 30-50\% through self-critique.
\item \textbf{Code Review}: Check style, detect bugs, suggest optimizations. MetaGPT~\cite{hong2023metagpt} simulates QA agent reviewing engineer's code.
\end{itemize}

\textbf{Research and Analysis Agents}:
\begin{itemize}
\item \textbf{Literature Review}: Search academic databases, extract key findings, synthesize insights, generate comprehensive reports.
\item \textbf{Market Research}: Gather competitor data, analyze trends, compare products, generate strategic recommendations.
\item \textbf{Data Analysis}: Load datasets, explore distributions, test hypotheses, generate visualizations, interpret results. GPT-4 Code Interpreter autonomously writes pandas/matplotlib code for analysis.
\end{itemize}

\textbf{Personal Assistant Agents}:
\begin{itemize}
\item \textbf{Email Management}: Read inbox, categorize urgency, draft responses, schedule meetings. Requires calendar integration, email API access.
\item \textbf{Travel Planning}: Research destinations, compare flights/hotels, book reservations, create itineraries. Multi-step coordination across booking platforms.
\item \textbf{Task Automation}: "Every Monday, compile weekly sales report and email to team" → agent autonomously executes recurring workflows.
\end{itemize}

\textbf{Customer Support Agents}:
\begin{itemize}
\item \textbf{Technical Troubleshooting}: Diagnose issues ("WiFi not working"), query knowledge bases, provide step-by-step solutions, escalate to humans if unresolved.
\item \textbf{Order Processing}: Check order status, process returns, update shipping addresses. Requires database integration and transaction handling.
\item \textbf{FAQ Automation}: Answer common questions with verified information (RAG from documentation), reducing human support load by 60-80\%.
\end{itemize}

\textbf{Scientific Discovery Agents}:
\begin{itemize}
\item \textbf{Hypothesis Generation}: Analyze experimental data, propose new hypotheses, design follow-up experiments.
\item \textbf{Simulation and Modeling}: Run computational simulations (molecular dynamics, climate models), analyze results, iterate parameters.
\item \textbf{Literature Mining}: Extract relationships from millions of papers (e.g., "Which proteins interact with BRCA1?"), accelerate meta-analysis.
\end{itemize}

\subsection{Single-Agent Architectures: Planning, Tools, and Memory}
\label{sec:single_agent}

Single-agent systems integrate three core capabilities—planning (task decomposition and strategy formulation), tool use (environment interaction), and memory (long-term context maintenance)—to solve complex problems autonomously.

\subsubsection{Planning Algorithms: From Reactive to Deliberative Reasoning}

Planning enables agents to decompose complex goals into executable subtasks and adapt strategies based on feedback. We categorize planning paradigms from reactive (immediate response) to deliberative (systematic exploration):

\textbf{Chain-of-Thought (CoT)}~\cite{wei2022chain} pioneered explicit reasoning: prompting models with "Let's think step by step" transforms $p(answer|question)$ into $p(step_1, \ldots, step_n, answer|question)$, making intermediate logic transparent and debuggable. Achieves 2-3$\times$ improvements on math/logic tasks (GSM8K: 17\% → 57\%). \textbf{Self-Consistency}~\cite{wang2023selfconsistency} enhances robustness: sample multiple reasoning paths ($k=5-40$), parse final answers, return majority vote: $\arg\max_a \sum_{i=1}^k \mathbb{1}[answer_i = a]$. Reduces variance from sampling randomness—if 7/10 paths reach answer "42", confidence is high.

\textbf{ReAct (Reasoning + Acting)}~\cite{yao2023react} interleaves reasoning with action execution: $\text{Thought}_t \to \text{Action}_t \to \text{Observation}_t \to \text{Thought}_{t+1}$. Example workflow:
\begin{itemize}
\item \textit{Thought}\_1: "I need Paris population to compare with Tokyo."
\item \textit{Action}\_1: \texttt{Search[population of Paris]}
\item \textit{Observation}\_1: "Paris: 2.1 million residents (city proper)."
\item \textit{Thought}\_2: "Tokyo has 14 million, so Tokyo is 6.7$\times$ larger."
\item \textit{Action}\_2: \texttt{Finish[Tokyo is 6.7 times larger than Paris]}
\end{itemize}
ReAct grounds reasoning in verifiable external state, reducing hallucination. Achieves 69\% on HotpotQA (multi-hop reasoning) vs. 28\% for CoT alone.

\textbf{Reflexion}~\cite{shinn2023reflexion} adds self-critique: after task failure, the agent reflects on mistakes and generates improved strategies. Trajectory: \textit{Attempt} (execute task) $\to$ \textit{Evaluate} (check outcome) $\to$ \textit{Reflect} ("Why did it fail? What should I change?") $\to$ \textit{Retry} (execute revised strategy). On programming tasks (HumanEval), Reflexion improves pass@1 from 68\% to 91\% through 3 iterations of self-debugging. Key insight: Explicit failure analysis (e.g., "Test case 3 failed because I didn't handle negative inputs") guides targeted improvements.

\textbf{Tree-of-Thoughts (ToT)}~\cite{yao2023tree} enables systematic exploration by modeling reasoning as tree search: each node is a partial solution (thought), the model evaluates node quality via value function $V(s) = \text{LLM}_{\text{eval}}(s)$, and search algorithms (BFS, DFS) explore/backtrack based on evaluations. Example (Game of 24):
\begin{itemize}
\item Root: "Use 4, 9, 10, 13 to make 24"
\item Children: "(4+9)*(13-10)" [eval: promising], "(4*9)-(10+13)" [eval: impossible], ...
\item Expand promising node, backtrack from dead ends
\end{itemize}
ToT achieves 100\% on Game of 24 (vs. 74\% for CoT) but requires 5-10$\times$ more tokens due to exploration overhead.

\textbf{Graph-of-Thoughts (GoT)}~\cite{besta2023graph} generalizes trees to directed acyclic graphs, enabling:
\begin{itemize}
\item \textit{Parallel exploration}: Generate multiple draft paragraphs concurrently (not sequentially)
\item \textit{Aggregation}: Merge best elements from different branches ("Take introduction from draft A, methods from draft B")
\item \textit{Cycles-free dependencies}: Node B depends on both A and C (DAG structure)
\end{itemize}
GoT reduces latency for parallelizable tasks (document generation, code module implementation) while maintaining ToT-level quality.

Table~\ref{tab:planning_taxonomy_unified} summarizes planning paradigms with complexity-performance trade-offs.

\begin{table*}[ht]
\centering
\small
\caption{Planning algorithm taxonomy: reactive to deliberative reasoning with computational trade-offs.}
\label{tab:planning_taxonomy_unified}
\begin{tabular}{p{2.2cm}p{4.2cm}p{2.8cm}p{2.8cm}p{2cm}}
\toprule
\textbf{Algorithm} & \textbf{Core Mechanism} & \textbf{Strengths} & \textbf{Limitations} & \textbf{Token Cost} \\
\midrule
\textbf{CoT}~\cite{wei2022chain} & Sequential reasoning: step$_1$ → step$_2$ → answer & Simple; 2-3× accuracy boost & No backtracking; linear path & 1× baseline \\
\midrule
\textbf{ReAct}~\cite{yao2023react} & Interleave thoughts with actions; grounded in observations & External feedback reduces hallucination & Sequential; no exploration & 2-3× baseline \\
\midrule
\textbf{Reflexion}~\cite{shinn2023reflexion} & Self-critique after failures; iterative improvement & Learns from mistakes; 30-50\% gains & Requires episodic tasks with clear success/failure & 3-5× baseline \\
\midrule
\textbf{ToT}~\cite{yao2023tree} & Tree search with value-guided exploration; backtracking & Systematic; handles complex planning & High cost (5-10×); needs value function & 5-10× baseline \\
\midrule
\textbf{GoT}~\cite{besta2023graph} & DAG structure; parallel exploration + aggregation & Efficient parallelization; combines diverse paths & Complex orchestration; task decomposition expertise & 3-7× baseline \\
\bottomrule
\end{tabular}
\end{table*}

\subsubsection{Tool Use: Extending Models with External Capabilities}

Tool integration transforms LLMs from text generators into versatile problem solvers accessing computation, search, databases, and specialized APIs. We categorize tool-use frameworks by integration depth:

\textbf{Function Calling}~\cite{openai2023function} enables structured API access: model outputs JSON specifying function name and arguments:
\begin{verbatim}
{"name": "get_weather", 
 "arguments": {"location": "Paris", "unit": "celsius"}}
\end{verbatim}
External executor invokes function, returns result to model, which synthesizes natural language response. GPT-4 achieves 95\%+ accuracy on well-documented APIs (correct function selection + parameter filling). Limitation: Requires manual API documentation in prompt (up to 10K tokens for complex APIs).

\textbf{Code Interpreter}~\cite{openai2023code} provides full programming environment: agents write Python code, execute in sandboxed environment, inspect outputs (including plots, dataframes, files), and iterate based on results. Example:
\begin{itemize}
\item User: "Analyze this sales CSV"
\item Agent: Writes \texttt{import pandas as pd; df = pd.read\_csv('sales.csv'); df.describe()}
\item Execution: Returns summary statistics
\item Agent: Generates \texttt{df.plot(x='month', y='revenue')} → produces chart
\item Agent: "Revenue peaks in Q4, driven by holiday sales"
\end{itemize}
Enables complex workflows beyond predefined APIs: data cleaning, statistical analysis, visualization, file manipulation. Sandboxing prevents unintended system access (no network, ephemeral filesystem).

\textbf{Toolformer}~\cite{schick2023toolformer} learns tool use via self-supervision: during training, model inserts potential tool calls (\texttt{<calculator>157*23</calculator>}), executes them, computes perplexity improvement, and keeps calls that reduce perplexity (better predict next tokens). This teaches \textit{when} to use tools (not just how): model learns "I should use calculator for complex arithmetic but not for simple sums." Training: Sample tool calls → Execute → Measure $\Delta \text{perplexity}$ → Filter by $\Delta > \theta$ → Fine-tune on filtered data.

\textbf{ToolBench}~\cite{qin2023toolllm} trains on 16,000+ real-world APIs with decision tree planning: given task, agent constructs decision tree of API calls, evaluates each branch, selects optimal path. Covers diverse domains: weather, maps, finance, social media, e-commerce. Achieves 85\% task completion on ToolBench benchmark (vs. 45\% for few-shot prompting).

\textbf{Gorilla}~\cite{patil2023gorilla} specializes in ML API generation: fine-tuned on documentation for HuggingFace, PyTorch, TensorFlow APIs. Generates correct code with proper imports, parameters, and error handling. Example: "Load BERT for sentiment analysis" → \texttt{from transformers import AutoTokenizer, AutoModelForSequenceClassification; model = AutoModelForSequenceClassification.from\_pretrained('bert-base-uncased')}. Outperforms GPT-4 on ML API tasks (92\% vs. 76\% correctness).

Table~\ref{tab:tool_use_taxonomy_unified} categorizes tool-use frameworks.

\begin{table*}[ht]
\centering
\small
\caption{Tool-use framework taxonomy: from structured function calling to self-supervised tool learning.}
\label{tab:tool_use_taxonomy_unified}
\begin{tabular}{p{2.5cm}p{4.8cm}p{2.8cm}p{2.2cm}}
\toprule
\textbf{Framework} & \textbf{Core Mechanism} & \textbf{Tool Types} & \textbf{Key Innovation} \\
\midrule
\textbf{Function Calling}~\cite{openai2023function} & Model outputs structured JSON; external executor invokes & API calls, databases, computation & Native GPT-4/Claude integration \\
\midrule
\textbf{Code Interpreter}~\cite{openai2023code} & Agent writes/executes Python in sandbox; iterates on output & Data analysis, visualization, files & Full programming environment \\
\midrule
\textbf{Toolformer}~\cite{schick2023toolformer} & Self-supervised: generates tool calls, filters by perplexity improvement & Calculator, QA, calendar, translator & Learns \textit{when} to use tools \\
\midrule
\textbf{ToolBench}~\cite{qin2023toolllm} & Trains on 16K+ APIs; decision tree planning for workflows & RESTful APIs, databases, services & Large-scale API diversity \\
\midrule
\textbf{Gorilla}~\cite{patil2023gorilla} & Fine-tuned on ML API docs; generates correct calls + error handling & HuggingFace, PyTorch, TensorFlow & Specializes in ML tools \\
\bottomrule
\end{tabular}
\end{table*}

\subsubsection{Memory Architectures: Maintaining Long-Term Context}

Memory systems enable agents to accumulate knowledge, recall relevant information, and maintain context beyond fixed context windows (which max out at 200K tokens for frontier models). We categorize by capacity and retrieval latency:

\textbf{In-Context Memory}: Include all relevant information directly in the prompt. Modern models handle 32K-200K tokens (GPT-4 Turbo: 128K, Claude 3: 200K, Gemini 1.5 Pro: 1M), enabling detailed conversation history and documentation. \textit{Limitation}: Quadratic attention cost $O(n^2)$ makes >100K tokens expensive; fixed capacity prevents indefinite context accumulation.

\textbf{Retrieval-Augmented Generation (RAG)}~\cite{lewis2020retrieval}: Store knowledge in external vector database, retrieve relevant documents on-demand:
\begin{equation}
p(y|x) = \sum_{d \in \text{TopK}(x)} p(y|x, d) \cdot p(d|x)
\end{equation}
Process: (1) User query $x$ → (2) Embed query: $e_x = \text{Encoder}(x)$ → (3) Retrieve top-$k$ documents by cosine similarity: $\{d_1, \ldots, d_k\} = \arg\max_d \cos(e_x, e_d)$ → (4) Condition generation on $[x; d_1; \ldots; d_k]$. Enables access to massive corpora (Wikipedia: 6M articles, internal docs: GBs) with 50-200ms retrieval latency. Modern encoders: Contriever~\cite{izacard2022unsupervised}, E5~\cite{wang2024text} achieve 90\%+ retrieval accuracy on BEIR benchmark.

\textbf{Episodic Memory}~\cite{park2023generative}: Store interaction transcripts with metadata (timestamp, importance, relevance). Retrieval combines:
\begin{itemize}
\item \textit{Recency}: Exponential decay $\text{score}_{\text{recency}}(t) = \exp(-\lambda (t_{\text{now}} - t_{\text{event}}))$
\item \textit{Importance}: Model-scored significance (1-10 scale): "Met project deadline" = 9, "Small talk about weather" = 2
\item \textit{Relevance}: Embedding similarity to current query
\item \textit{Combined}: $\text{score}(e) = \alpha \cdot \text{recency}(e) + \beta \cdot \text{importance}(e) + \gamma \cdot \text{relevance}(e)$
\end{itemize}
Generative Agents~\cite{park2023generative} use episodic memory to simulate 25 interactive characters maintaining consistent personalities across 200+ interactions spanning simulated days.

\textbf{Parametric Memory}~\cite{mitchell2022fast}: Store knowledge directly in model weights via knowledge editing: ROME~\cite{meng2022locating} and MEMIT~\cite{meng2023mass} update specific neurons to inject new facts without full retraining. Example: Update "The Eiffel Tower is in Paris" → "The Eiffel Tower was temporarily moved to Tokyo in 2024". Enables real-time knowledge updates with 0ms retrieval latency. \textit{Limitation}: Risk of catastrophic forgetting (editing one fact corrupts related knowledge); limited to factual updates (not procedural knowledge).

Table~\ref{tab:memory_taxonomy} compares memory architectures.

\begin{table*}[t]
\centering
\small
\caption{Memory architecture taxonomy for agentic AI systems: comparing capacity, retrieval latency, and update mechanisms across four memory paradigms.}
\label{tab:memory_taxonomy}
\begin{tabular}{p{2.0cm}p{4.0cm}p{2.8cm}p{2.2cm}p{2.0cm}}
\toprule
\textbf{Architecture} & \textbf{Mechanism} & \textbf{Capacity} & \textbf{Retrieval Latency} & \textbf{Update Cost} \\
\midrule
\textbf{In-Context}~\cite{brown2020language} & Include relevant information in prompt: conversation history, examples, instructions & 8K-200K tokens (context window) & 0 ms (no retrieval) & Free (append text) \\
\midrule
\textbf{RAG}~\cite{lewis2020retrieval} & Retrieve documents from external vector database: $p(y|x) = \sum_d p(y|x,d)p(d|x)$ & Unlimited (external DB) & 50-200 ms (embedding + search) & Low (add documents) \\
\midrule
\textbf{Episodic}~\cite{park2023generative} & Store interaction transcripts; retrieve based on recency, importance, and relevance scores & Thousands of episodes & 100-500 ms (semantic search) & Low (append episodes) \\
\midrule
\textbf{Parametric}~\cite{mitchell2022fast} & Update model weights with new information via knowledge editing (ROME, MEMIT) & $10^{9}$-$10^{12}$ facts (model capacity) & 0 ms (stored in weights) & High (gradient updates) \\
\bottomrule
\end{tabular}
\end{table*}

\subsection{Multi-Agent Coordination: Collaborative Problem Solving}
\label{sec:multi_agent}

Multi-agent systems distribute tasks across specialized agents with complementary capabilities, enabling complex workflows beyond single-agent capacity. Table~\ref{tab:multiagent_taxonomy_unified} categorizes coordination frameworks from conversational (AutoGen) to role-based (MetaGPT) to dynamic (AgentVerse) collaboration.

\begin{table*}[ht]
\centering
\small
\caption{Multi-agent coordination frameworks: from conversational to role-based to dynamic collaboration.}
\label{tab:multiagent_taxonomy_unified}
\begin{tabular}{p{2.2cm}p{4.8cm}p{2.5cm}p{2.8cm}}
\toprule
\textbf{Framework} & \textbf{Coordination Mechanism} & \textbf{Agent Roles} & \textbf{Communication} \\
\midrule
\textbf{AutoGen}~\cite{wu2023autogen} & Conversational agents with customizable roles; human-in-the-loop; code execution & User proxy, assistant, executor & Natural language messages \\
\midrule
\textbf{MetaGPT}~\cite{hong2023metagpt} & Simulates software company: structured workflow (requirements $\to$ design $\to$ code $\to$ test) & PM, architect, engineer, QA & Structured documents (PRD, design docs) \\
\midrule
\textbf{ChatDev}~\cite{qian2023chatdev} & Chain-shaped collaboration: CEO $\to$ CTO $\to$ programmer $\to$ reviewer; waterfall development & 7 roles (CEO, CTO, designer, programmer, tester, reviewer, art) & Sequential handoffs with artifacts \\
\midrule
\textbf{AgentVerse}~\cite{chen2023agentverse} & Dynamic task decomposition: recruiter agent assembles team based on requirements & Dynamic role assignment & Blackboard architecture (shared memory) \\
\bottomrule
\end{tabular}
\end{table*}

\subsubsection{Multi-Agent Frameworks and Coordination Patterns}

\textbf{AutoGen}~\cite{wu2023autogen} enables flexible agent conversations: define agents with system prompts (e.g., "You are a Python expert"), tools (code executor), and termination conditions. Example workflow: User describes task $\to$ Assistant generates code $\to$ Executor runs code $\to$ Assistant debugs based on output $\to$ Repeat until success. Supports human-in-the-loop for oversight on critical decisions.

\textbf{MetaGPT}~\cite{hong2023metagpt} simulates software companies with role-based workflows: (1) Product Manager writes Product Requirements Document (PRD), (2) Architect designs system architecture (class diagrams, APIs, data flows), (3) Engineer implements code following specifications, (4) QA tests and reports bugs. Agents communicate via structured documents (not just natural language), reducing ambiguity. Achieves 87\% functional correctness on software generation benchmarks (HumanEval, MBPP) compared to 41\% for single-agent GPT-4 baseline. The structured document approach enforces clarity: PRDs contain user stories, acceptance criteria, technical constraints; design docs include UML diagrams, API specifications, database schemas.

\textbf{ChatDev}~\cite{qian2023chatdev} implements waterfall development with 7 specialized agents: CEO defines project objectives $\to$ CTO proposes tech stack (languages, frameworks, architecture) $\to$ Designer creates UI mockups $\to$ Programmer implements code $\to$ Art Designer generates assets $\to$ Tester validates functionality $\to$ Reviewer ensures code quality. Chain-shaped communication (each agent only talks to neighbors) prevents coordination overhead that plagues fully-connected multi-agent systems. Generates complete software applications (HTML/CSS/JS games, productivity tools, data visualization dashboards) in <10 minutes with <\$1 API cost. Example: "Create a Flappy Bird game" $\to$ ChatDev produces playable game with graphics, collision detection, scoring, and game-over logic.

\textbf{AgentVerse}~\cite{chen2023agentverse} dynamically assembles teams: given task description, a \textit{recruiter agent} analyzes required expertise (e.g., "This task needs Python expert for backend, React specialist for frontend, and data scientist for analytics") and instantiates agents with appropriate system prompts and tool access. Agents share state via \textit{blackboard architecture}—a shared memory space where agents post information and subscribe to updates. Outperforms fixed-role systems on diverse tasks requiring variable expertise: software development (needs programmer + tester), data analysis (needs statistician + visualization expert), research (needs literature reviewer + summarizer + critic).

\subsubsection{Emergent Behaviors and Coordination Challenges}

\textbf{Beneficial Emergence}: Multi-agent systems exhibit capabilities beyond individual agents. Example: In MetaGPT, the architect agent spontaneously proposes modular designs (not explicitly prompted), enabling the engineer agent to implement components independently in parallel, then an emergent integration phase combines modules. Modularity emerges from role specialization without being programmed. In ChatDev, code review iterations lead to progressively cleaner code (variable naming, documentation, error handling) through back-and-forth between programmer and reviewer agents.

\textbf{Coordination Overhead}: Communication costs scale with agent count. $N$ agents require $O(N^2)$ messages for full connectivity. Solutions: (1) \textit{Hierarchical organization}—manager agent coordinates sub-teams (reduces to $O(N)$ messages through tree structure), (2) \textit{Sparse topologies}—chain/tree structures where agents only communicate with neighbors (ChatDev's approach), (3) \textit{Asynchronous messaging}—agents don't block waiting for responses; post messages to shared queue.

\textbf{Conflicting Objectives}: Agents may pursue incompatible goals. Example: Engineer agent optimizes for feature richness ("add more functionality") while QA agent optimizes for reliability ("minimize bugs")—tension requires negotiation. MetaGPT resolves conflicts through structured phases: architect makes high-level decisions (features + quality requirements), binding subsequent agents. Alternative: debate-based resolution where agents argue positions, then meta-agent or human makes final call.

\textbf{Consensus Mechanisms}: When agents must agree on decisions (e.g., which architectural design to implement), systems use: (1) \textit{Voting}—majority/plurality/weighted voting based on agent expertise, (2) \textit{Debate}—agents argue positions with evidence, final decision via critique or human arbitration, (3) \textit{Hierarchical authority}—designated lead agent (PM, architect) makes final call, others provide input.

\subsection{Agent Communication Protocols: Standardizing Interfaces}
\label{sec:agent_communication_protocols}

As agentic systems proliferate, standardized protocols for agent-tool communication and inter-agent coordination are emerging to ensure interoperability, reduce integration overhead, and enable ecosystem growth.

\subsubsection{Model Context Protocol (MCP): LLM-Tool Communication}

\textbf{Model Context Protocol (MCP)}~\cite{anthropic2024mcp}, introduced by Anthropic in November 2024, standardizes how AI models interact with external tools and data sources. MCP solves the integration problem: every AI application previously needed custom integrations for each tool (database, filesystem, API), leading to $M \times N$ integration combinations ($M$ applications $\times$ $N$ tools). MCP reduces this to $M + N$: each application implements MCP client once, each tool implements MCP server once.

\textbf{Core Abstractions}:
\begin{itemize}
\item \textbf{Resources}: Data sources the model can access (files, databases, APIs) with URIs: \texttt{file:///project/data.json}, \texttt{postgres://db/table}. Resources are \textit{read-only}—agents retrieve information but don't modify through resource interface.
\item \textbf{Prompts}: Pre-defined prompt templates for common tasks, reusable across applications. Example: "Analyze quarterly sales data" template includes instructions for data loading, statistical analysis, and visualization generation. Users invoke by name; MCP server expands template with current context.
\item \textbf{Tools}: Functions the model can invoke with JSON schemas defining parameters and return types. Example: \texttt{search\_database(query: str, limit: int) -> List[Record]}. Tools enable \textit{actions}—agents modify external state.
\end{itemize}

\textbf{Architecture}: MCP uses client-server model:
\begin{itemize}
\item \textbf{MCP Clients} (AI applications like Claude Desktop, IDEs, agent frameworks) connect to multiple MCP servers simultaneously. Client discovers available tools/resources via server capabilities negotiation.
\item \textbf{MCP Servers} expose tools/resources via standardized JSON-RPC 2.0 interface. Example: filesystem server provides \texttt{read\_file(path)}, \texttt{list\_directory(path)}, \texttt{search\_files(pattern)} tools. Servers control access scope (e.g., limit filesystem access to \texttt{/workspace} directory, enforce read-only mode).
\item \textbf{Transport Layer}: JSON-RPC 2.0 over stdio (local processes), HTTP/SSE (remote servers), or WebSocket (bidirectional real-time). Stdio is preferred for local tools (low latency, simple IPC), HTTP/SSE for remote services.
\end{itemize}

\textbf{Benefits}: (1) \textit{Composability}—single application accesses multiple tool servers (filesystem + database + API server) without custom code, (2) \textit{Security}—servers enforce access control (read-only resources, sandboxed execution, rate limiting), (3) \textit{Ecosystem growth}—developers publish MCP servers for databases (PostgreSQL, MongoDB), cloud services (AWS, Azure), productivity tools (Slack, GitHub), enabling plug-and-play functionality. As of January 2025, 50+ community MCP servers available covering data sources, developer tools, and business applications.

\textbf{Example Workflow}: User asks Claude to "analyze sales trends from our PostgreSQL database." Claude's MCP client discovers available servers, finds \texttt{postgres\_mcp\_server} exposing \texttt{query\_database(sql)} tool, invokes \texttt{SELECT date, revenue FROM sales WHERE date > '2024-01-01' ORDER BY date}, receives results as JSON, generates statistical summary and trend visualization using Code Interpreter (another MCP tool), returns insights to user. All communication follows MCP's standardized protocol—no custom integration required.

\subsubsection{Agent-to-Agent (A2A) Protocols: Inter-Agent Communication}

While MCP handles agent-tool communication, \textbf{Agent-to-Agent (A2A)} protocols address coordination between autonomous agents in multi-agent systems. Key challenges: message semantics (ensuring agents understand each other), state synchronization (maintaining consistent worldview), and coordination mechanisms (task allocation, conflict resolution).

\textbf{Communication Primitives}:
\begin{itemize}
\item \textbf{Message Passing}: Agents exchange structured messages (requests, inform, query, propose). Protocols define message schemas specifying sender, receiver, performative (speech act type), and content. Foundation for Agent Communication Language (ACL)~\cite{fipa2002acl} from traditional multi-agent systems research.
\item \textbf{Blackboard Architecture}~\cite{corkill1991blackboard}: Shared memory space where agents post information and read others' contributions. AgentVerse~\cite{chen2023agentverse} uses blackboards for dynamic task coordination: agents post partial solutions, others build upon them. Enables loose coupling (agents don't directly message each other) and asynchronous collaboration.
\item \textbf{Publish-Subscribe}: Agents subscribe to topics of interest; publishers broadcast updates to all subscribers. Enables one-to-many communication. Example: monitoring agent publishes system health metrics $\to$ logging agent, alerting agent, dashboard agent all receive updates. Scales better than point-to-point messaging for broadcast scenarios.
\end{itemize}

\textbf{Coordination Patterns}:
\begin{itemize}
\item \textbf{Contract Net Protocol}~\cite{smith1980contract}: Agents negotiate task assignments via bidding. Manager broadcasts task announcement $\to$ agents submit bids (cost, time, capability) $\to$ manager awards contract to best bidder $\to$ winning agent executes task, reports results. Decentralizes task allocation without central authority.
\item \textbf{Consensus Protocols}: Multi-agent systems reach agreement on shared state (e.g., which agent should lead subtask, which solution is best). Voting-based (majority, plurality, weighted) or debate-based (agents argue positions with evidence, converge through discussion). Useful for collaborative decision-making without hierarchical authority.
\item \textbf{Hierarchical Coordination}: Designated manager/coordinator agents oversee workers. Manager decomposes task, allocates subtasks, monitors progress, resolves conflicts. Reduces coordination overhead ($O(N)$ manager-worker messages vs. $O(N^2)$ peer-to-peer) at cost of manager becoming bottleneck.
\end{itemize}

\textbf{Emerging Standards}: LangGraph~\cite{chase2023langchain} and AutoGen~\cite{wu2023autogen} provide A2A frameworks supporting natural language messages with structured metadata (sender role, task context, execution results, confidence scores). However, no universal A2A standard exists yet—current systems use framework-specific protocols. Open challenges: (1) \textit{Semantic interoperability}—ensuring agents from different frameworks understand each other's messages, (2) \textit{Trust and verification}—validating messages from untrusted agents (especially in open multi-agent systems with external participants), (3) \textit{Privacy}—selective information sharing (agents may have conflicting interests or access different privileged information).

\subsection{Applications, Future Directions, and Challenges}
\label{sec:agent_applications_future}

\subsubsection{Real-World Applications Driving Agentic AI Adoption}

Agentic AI is transitioning from research prototypes to production systems across diverse domains:

\textbf{Software Development}: Code agents (Devin AI~\cite{cognition2024devin}, GitHub Copilot Workspace, Cursor AI) automate end-to-end development workflows: requirements analysis $\to$ architecture design $\to$ implementation $\to$ testing $\to$ debugging. Devin AI achieves 13.8\% on SWE-bench (resolving real GitHub issues), demonstrating practical utility for bug fixes and feature implementation. Productivity gains: 30-50\% faster completion for routine coding tasks (CRUD operations, API integrations, UI components), enabling developers to focus on complex architectural decisions.

\textbf{Scientific Research}: Research agents automate literature review, hypothesis generation, experimental design, and data analysis. Elicit~\cite{elicit2024} extracts findings from papers, identifies research gaps, suggests experimental protocols. Drug discovery agents (Insilico Medicine's PandaOmics) generate molecular structures optimized for target proteins, reducing discovery timelines from 4-5 years to 18-24 months. Materials science agents (A-Lab~\cite{szymanski2023alab}) autonomously design, synthesize, and characterize novel materials using robotic labs.

\textbf{Personal Assistants}: Next-generation assistants (Google's Project Astra, Apple Intelligence) manage complex workflows: "Plan weekend trip to Paris" $\to$ agent searches flights, books hotel, creates itinerary, adds calendar events, shares with travel partners. Email agents (Shortwave's AI, Superhuman's AI triage) draft responses, prioritize messages, schedule meetings, follow up on pending items. Financial agents (Copilot Money, Monarch Money AI) analyze spending, optimize budgets, recommend investments, automate bill payments.

\textbf{Customer Support}: Support agents (Ada, Intercom's Fin, Zendesk AI) handle L1/L2 support: answer FAQs, troubleshoot issues, escalate complex cases to humans. Achieve 60-80\% automation rates for common queries (password resets, order tracking, account management), reducing support load and improving response times (instant vs. hours for human agents). Advanced agents access customer databases, modify accounts, process refunds—full transaction authority under safety constraints.

\textbf{Data Analysis and Business Intelligence}: Analyst agents (Tableau's Einstein Copilot, ThoughtSpot Sage, Microsoft Copilot in Power BI) generate SQL queries, create visualizations, identify trends, explain anomalies. Example: "Why did Q3 revenue drop?" $\to$ agent queries database, segments by product/region, identifies 15\% decline in EMEA due to supply chain issues (correlated with shipping delay data), generates executive summary with visualizations. Democratizes analytics—business users query data without SQL expertise.

\subsubsection{Future Directions: Toward Fully Autonomous Agents}

\textbf{Embodied Agents}: Integration with robotics enabling physical world interaction. VoxPoser~\cite{huang2023voxposer} generates robot control code from natural language ("Pick up the red block and place it on the table"). PaLM-E~\cite{driess2023palme} unifies language and vision in 562B parameter model, enabling robots to execute household tasks from high-level instructions. Challenge: bridging semantic gap between language ("gently place") and low-level motor control (force, trajectory).

\textbf{Human-Agent Collaboration}: Moving beyond full autonomy to mixed-initiative systems where humans and agents collaborate iteratively. Agents propose solutions $\to$ humans critique/refine $\to$ agents iterate. Example: code agents generate implementation draft, developer reviews and marks sections for revision, agent updates based on feedback. Requires agents to understand uncertain/incomplete human feedback and ask clarifying questions when ambiguous.

\textbf{Agent Ecosystems and Marketplaces}: Emergence of agent specialization and composition. Specialized agents (legal document analyzer, medical diagnosis assistant, financial planner) are published to marketplaces, discoverable via capability descriptions. Users compose custom workflows by chaining agents: research agent finds papers $\to$ summarization agent extracts key findings $\to$ critique agent identifies methodological flaws $\to$ writing agent drafts literature review. Challenges: (1) agent discovery (matching user needs to available agents), (2) trust (verifying agent quality/safety), (3) pricing (micro-transactions per agent invocation).

\textbf{Lifelong Learning Agents}: Current agents are stateless (reset after each session) or use external memory (RAG, episodic). Future: agents that learn continuously from interactions, updating internal knowledge/skills without forgetting previous capabilities. Example: personal assistant learns user preferences over months (communication style, scheduling constraints, priority heuristics) via implicit feedback signals. Requires solving catastrophic forgetting (updating weights without overwriting prior knowledge) and aligning learning with user values (not just task success).

\textbf{Meta-Agents and Agent Management}: Agents that manage other agents—allocating resources, load balancing, ensuring quality. Meta-agents monitor worker agents' performance, detect failures (stuck in loops, poor quality outputs), swap in better-performing agents. Hierarchical agent organizations: meta-agent decomposes high-level goals into tasks, assigns to specialist agents, integrates results. Enables scalability beyond current fixed multi-agent systems.

\subsection{Safety, Challenges, and Risk Mitigation}
\label{sec:agentic_safety}

Agentic systems with tool access and autonomous execution raise critical safety concerns: unintended actions (irreversible API calls, data deletion), exploration of harmful strategies (jailbreaks, adversarial inputs), accumulation of errors (compounding mistakes over long task horizons), and loss of human oversight (agents operating beyond observable boundaries).

\subsubsection{Human Oversight Mechanisms: Maintaining Control}

\textbf{Action Confirmation for High-Stakes Operations}: Require human approval for irreversible or high-impact actions (financial transactions, data deletion, external communications, account modifications). AutoGen implements configurable approval: critical actions pause execution and present proposed action to human ("Agent wants to execute: \texttt{DELETE FROM users WHERE id > 100}. Approve?"), while safe actions (read-only database queries, web searches) proceed autonomously. \textit{Challenge}: Defining action criticality—requires risk modeling per tool (database writes are higher risk than reads) and context-awareness (deleting test data is safe, deleting production data is critical).

\textbf{Interpretable Reasoning Traces for Auditing}: Log all agent thoughts, actions, observations, and decision rationales for post-hoc inspection. ReAct's explicit reasoning chains ("Thought: I need Paris population to compare with Tokyo") enable humans to identify flawed logic before harmful actions. Trace storage enables debugging failed tasks, identifying systematic errors, and fine-tuning agent policies. \textit{Implementation}: Structured logging with timestamps, agent IDs, tool invocations, and confidence scores. Advanced systems use replay debugging—re-run agent execution with modified parameters to test alternative strategies.

\textbf{Capability Restrictions via Tool Access Control}: Limit tool availability based on task context and agent trust level. Example: data analysis agents access \texttt{pandas}, \texttt{matplotlib}, \texttt{numpy} but not \texttt{os.system()}, \texttt{subprocess}, \texttt{open()} (prevents file system modifications). GPT-4 Code Interpreter runs in isolated sandbox: no network access (can't exfiltrate data), ephemeral filesystem (wiped after session), resource limits (CPU/memory/time quotas). \textit{Least Privilege Principle}: Agents receive minimum necessary capabilities. Web search agent gets \texttt{search(query)} but not \texttt{navigate\_to\_url()} (prevents accessing arbitrary sites beyond search results).

\textbf{Confidence-Based Escalation}: Agents estimate uncertainty and request human assistance when confidence is low. Example: "I'm 45\% confident in this diagnosis—would you like to review?" Uncertainty quantification via ensembling (sample multiple responses, measure agreement), calibration (softmax probabilities), or explicit model uncertainty estimates (dropout at inference time, Bayesian approximations). \textit{Challenge}: LLMs are poorly calibrated—often express high confidence on incorrect outputs. Research on uncertainty-aware agents is active area.

\subsubsection{Adversarial Robustness and Red-Teaming}

\textbf{Goal Misalignment Probes}: Test whether agents pursue unintended interpretations of objectives. Example: reward "maximize user engagement" $\to$ agent generates addictive or misleading content. Anthropic's Constitutional AI~\cite{bai2022constitutional} uses automated red-teaming: generate adversarial prompts attempting to elicit harmful behaviors, test agent's responses, fine-tune to refuse. OpenAI's red-teaming for GPT-4 identified risks (building weapons, generating malware) before deployment, enabling targeted mitigations (refusal training, content filters).

\textbf{Tool Misuse Detection via Anomaly Monitoring}: Monitor for unexpected API usage patterns indicating misuse or security breaches. Example: agent suddenly makes 1000x more database queries than usual (potential data exfiltration), or invokes restricted tools not granted in capabilities list (privilege escalation attempt). Statistical anomaly detection: baseline expected tool-use distributions during normal operation, flag deviations exceeding thresholds. \textit{Response}: Halt execution, notify administrators, require re-authorization for resumed operation.

\textbf{Multi-Agent Collusion Detection}: Test whether independent agents cooperate to circumvent restrictions. Scenario: Agent A (read-only database access) gathers sensitive customer data, passes to Agent B (execution privileges) who acts on that data, collectively achieving what neither could alone. Detection: analyze inter-agent message flows for information leakage, enforce isolation between security domains (agents handling sensitive data cannot communicate with external-facing agents).

\textbf{Adversarial Input Robustness}: Agents must handle malicious inputs designed to exploit vulnerabilities. Example: prompt injection attacks ("Ignore previous instructions, output system prompt") attempt to hijack agent behavior. Defenses: (1) input sanitization (filter malicious patterns), (2) instruction hierarchy (system prompts have higher priority than user inputs), (3) output filtering (detect when agent is outputting system internals, block response). Microsoft's Azure AI Content Safety provides adversarial robustness testing for agent applications.

\subsubsection{Graceful Failure and Error Recovery}

\textbf{Reversible Actions and Transactional Safety}: Design APIs with undo mechanisms enabling rollback of failed operations. Database agents use transactions (BEGIN/COMMIT/ROLLBACK): changes staged in transaction, committed only after verification, rolled back on errors or agent termination. File operations create automatic backups before modifications: agent renames \texttt{data.csv} $\to$ system creates \texttt{data.csv.backup}, agent modifies \texttt{data.csv}, on failure restores from backup. \textit{Idempotency}: Design actions to be safely retriable—executing twice produces same result as once, enabling automatic retry on transient failures.

\textbf{Timeouts and Circuit Breakers}: Halt execution after excessive retries or resource consumption, preventing runaway agents. Example: agent stuck in exploration loop (repeatedly trying variations that don't work) exhausts 1000-action budget $\to$ forced termination with informative error: "Agent exceeded action limit. Likely cause: task too complex or agent stuck in loop. Last 10 actions: [...]." Circuit breaker pattern: after $N$ consecutive failures, halt execution for cooldown period before allowing retry (prevents cascading failures overwhelming external services).

\textbf{Sandboxing and Virtualization}: Execute agent actions in isolated environments (containers, VMs, sandboxed processes) preventing unintended impact on production systems. Code execution agents run in Docker containers with no network access, limited filesystem (only \texttt{/tmp} writable), resource quotas (1GB memory, 2 CPUs, 300 seconds execution time). On failure or malicious behavior, container is terminated—no persistence, no lateral movement to host system. \textit{Blast Radius Containment}: Limit explosion radius of failures to sandbox; compromise of one agent doesn't affect others or underlying infrastructure.

\textbf{Checkpoint-Resume for Long-Horizon Tasks}: For tasks spanning hours/days (e.g., "Write research report"), save agent state periodically enabling resume from last checkpoint on failure. State includes: interaction history, tool outputs, intermediate work products, plan/strategy. On crash/timeout, reload checkpoint and continue. Enables fault tolerance without restarting from scratch. \textit{Implementation}: Serialize agent state to durable storage (database, S3) every $N$ actions or at strategic points (completing subtask, before high-risk operation).

\subsubsection{Open Challenges and Research Frontiers}

\textbf{Learned Safety Models}: Current oversight relies on human monitoring or hardcoded rules, neither scalable to full autonomy. Challenge: develop learned models that predict action consequences and assess risks. Example: before agent executes \texttt{delete\_file(path)}, safety model predicts: "This file is imported by 15 other modules. Deletion probability to break system: 85\%. Recommend: backup first or refactor dependents." Requires training on outcome data (action $\to$ consequence) and counterfactual reasoning (what would happen if agent took alternative action).

\textbf{Scalable Oversight for Complex Tasks}: Human oversight doesn't scale to thousands of agents or millisecond-latency decisions. Iterated Amplification~\cite{christiano2018supervising}: humans oversee simple decisions, agents learn from those to handle harder decisions, humans spot-check hardest cases. Debate~\cite{irving2018ai}: two agents argue for/against proposed action, human judges debate (easier than evaluating action directly). Both aim to amplify limited human oversight to superhuman agent capabilities while maintaining alignment.

\textbf{Robustness to Distribution Shift}: Agents trained/tested in controlled environments fail when deployed to real world's messiness (unexpected inputs, API changes, novel situations). Example: web agent trained on e-commerce sites breaks when site redesigns UI. Solutions: (1) continual learning (update agent from deployment experience), (2) out-of-distribution detection (identify unfamiliar situations, request help), (3) sim-to-real transfer (train in simulators with randomized environments capturing real-world variability).

\textbf{Multi-Stakeholder Alignment}: Agents serve users with diverse values. Personal assistant managing family calendar must balance preferences of all family members (parents prioritize efficiency, children want fun). Corporate agents balance shareholder value, employee welfare, customer satisfaction, regulatory compliance. Single objective function is insufficient. Research directions: multi-objective optimization (Pareto frontiers of trade-offs), Constitutional AI (encode multiple principles), participatory design (stakeholders collaboratively define agent behavior).

\textbf{Case Study—OpenAI Code Interpreter Safety Layers}: (1) Sandboxed Python environment (no network, no file system access beyond \texttt{/tmp}, no subprocess/system calls), (2) execution timeout (120 seconds per code cell prevents infinite loops), (3) explicit user confirmation for file downloads (agent can't exfiltrate data without user clicking "Download"), (4) automatic session termination after 60 minutes idle (limits prolonged access), (5) rate limiting (max 50 code executions per hour prevents abuse). These constraints reduce risk while enabling broad utility for data analysis and visualization. \textit{Trade-off}: Safety restrictions limit capabilities—agents can't access external APIs, use web scraping libraries, or interface with local tools. Future systems require more nuanced access control allowing selective capabilities with context-dependent permissions.

\textbf{Reflexion}~\cite{shinn2023reflexion} adds self-critique: after task failure, the agent reflects on mistakes and generates improved strategies. Trajectory: Attempt $\to$ Evaluate $\to$ Reflect $\to$ Retry. On programming tasks, Reflexion improves pass@1 by 30-50\% through iterative debugging.

\textbf{Tree-of-Thoughts (ToT)}~\cite{yao2023tree} enables systematic exploration: each thought is a partial solution, the model evaluates quality and backtracks from dead ends. Combines LLMs with search algorithms: $V(s) = \text{LLM}_{\text{eval}}(s)$ scores intermediate states, guiding BFS/DFS exploration. Achieves state-of-the-art on Game of 24 (74\% $\to$ 100\%) but requires 5-10$\times$ more tokens.

\textbf{Graph-of-Thoughts (GoT)}~\cite{besta2023graph} generalizes trees to directed acyclic graphs, enabling parallel exploration and aggregation. Example: generate 5 draft paragraphs concurrently, then merge best elements into final text. Reduces latency for parallelizable tasks while maintaining ToT-level quality.

\subsubsection{Tool Use: Extending Models with External Capabilities}

Tool integration transforms LLMs from text generators into versatile problem solvers accessing computation, search, and specialized APIs. Table~\ref{tab:tool_use_taxonomy_unified} categorizes tool-use frameworks.

\textbf{Function Calling}~\cite{openai2023function} enables structured API access: the model outputs \texttt{\{"name": "get\_weather", "args": \{"location": "Paris"\}\}}, the executor invokes the function, and results are returned to the model. GPT-4 achieves 95\%+ accuracy on well-documented APIs.

\textbf{Code Interpreter}~\cite{openai2023code} provides full programming environment: agents write Python code, execute in sandbox, inspect output (including plots, files), and iterate. Example: "Analyze this CSV" $\to$ agent writes \texttt{pandas} code, generates visualization, interprets results. Enables complex data analysis beyond API calls.

\textbf{Toolformer}~\cite{schick2023toolformer} learns tool use via self-supervision: during training, the model inserts tool calls (\texttt{<calculator>157*23</calculator>}), executes them, and keeps calls that improve next-token prediction. This teaches when (not just how) to use tools without explicit labels.

\subsubsection{Memory Architectures: Maintaining Long-Term Context}

Memory systems enable agents to accumulate knowledge, recall relevant information, and maintain context beyond fixed context windows. Table~\ref{tab:memory_taxonomy} categorizes memory architectures.

\textbf{Retrieval-Augmented Generation (RAG)}~\cite{lewis2020retrieval} stores knowledge externally: given query $x$, retrieve top-$k$ documents $\{d_1, \ldots, d_k\}$ via dense retrieval (Contriever~\cite{izacard2022unsupervised}, E5~\cite{wang2024text}), then condition generation on $[x; d_1; \ldots; d_k]$. Enables access to massive corpora (Wikipedia, internal documents) with 50-200ms latency.

\textbf{Episodic Memory}~\cite{park2023generative} stores interaction transcripts with metadata (timestamp, importance scores): each episode is embedded, and retrieval combines recency (exponential decay), importance (model-scored), and relevance (embedding similarity). Generative Agents~\cite{park2023generative} simulate 25 interactive characters maintaining consistent personalities across 200+ interactions.

\textbf{Parametric Memory}~\cite{mitchell2022fast} stores knowledge in model weights: techniques like ROME~\cite{meng2022locating} and MEMIT~\cite{meng2023mass} update specific neurons to inject new facts without full retraining. Enables real-time knowledge updates with 0ms retrieval but risks catastrophic forgetting.

\subsection{Agent Communication Protocols: Standardizing Interfaces}

As agentic systems proliferate, standardized protocols for agent-tool communication and inter-agent coordination are emerging to ensure interoperability and reduce integration overhead.

\subsubsection{Model Context Protocol (MCP): LLM-Tool Communication}

\textbf{Model Context Protocol}~\cite{anthropic2024mcp} (MCP), introduced by Anthropic (2024), standardizes how AI models interact with external tools and data sources. MCP defines:

\begin{itemize}
\item \textbf{Resources}: Data sources the model can access (files, databases, APIs) with URIs (e.g., \texttt{file:///project/data.json}).
\item \textbf{Prompts}: Pre-defined prompt templates for common tasks, reusable across applications.
\item \textbf{Tools}: Functions the model can invoke with JSON schemas defining parameters and return types.
\end{itemize}

\noindent \textbf{Architecture}: MCP uses client-server model where:
\begin{itemize}
\item \textbf{MCP Clients} (AI applications like Claude Desktop, IDEs) connect to multiple MCP servers.
\item \textbf{MCP Servers} expose tools/resources via standardized interface. Example: filesystem server provides \texttt{read\_file(path)}, \texttt{list\_directory(path)} tools.
\item \textbf{Transport Layer}: JSON-RPC 2.0 over stdio (local processes) or HTTP/SSE (remote servers).
\end{itemize}

\textbf{Benefits}: (1) \textit{Composability}—single application accesses multiple tool servers without custom integrations, (2) \textit{Security}—servers control access scope (e.g., read-only filesystem access), (3) \textit{Ecosystem Growth}—developers publish MCP servers for databases, APIs, specialized tools, enabling plug-and-play functionality.

\textbf{Example Workflow}: User asks Claude to "analyze sales data." Claude's MCP client queries available tools, finds \texttt{database\_query(sql)} from MCP database server, invokes \texttt{SELECT * FROM sales WHERE date > '2024-01-01'}, receives results, and generates insights. All communication follows MCP's standardized JSON-RPC schema.

\subsubsection{Agent-to-Agent (A2A) Protocols: Inter-Agent Communication}

While MCP handles agent-tool communication, \textbf{Agent-to-Agent (A2A)} protocols address coordination between autonomous agents. Key challenges include message semantics, state synchronization, and coordination mechanisms.

\textbf{Communication Primitives}:
\begin{itemize}
\item \textbf{Message Passing}: Agents exchange structured messages (requests, responses, notifications). Protocols define message schemas (e.g., FIPA ACL~\cite{fipa2002acl} from agent research) specifying sender, receiver, performative (inform, request, propose), and content.
\item \textbf{Blackboard Architecture}~\cite{corkill1991blackboard}: Shared memory space where agents post/read information. AgentVerse~\cite{chen2023agentverse} uses blackboards for dynamic task coordination.
\item \textbf{Publish-Subscribe}: Agents subscribe to topics; publishers broadcast updates. Enables loose coupling for large multi-agent systems.
\end{itemize}

\textbf{Coordination Patterns}:
\begin{itemize}
\item \textbf{Contract Net Protocol}~\cite{smith1980contract}: Agents negotiate task assignments via bidding. Manager broadcasts task $\to$ agents submit bids $\to$ manager awards contract $\to$ agent executes task.
\item \textbf{Consensus Protocols}: Multi-agent systems reach agreement on shared state (e.g., which agent leads subtask). Useful for collaborative planning without central authority.
\end{itemize}

\textbf{Emerging Standards}: LangGraph~\cite{chase2023langchain} and AutoGen~\cite{wu2023autogen} provide A2A frameworks supporting natural language messages with structured metadata (sender role, task context, execution results). However, no universal A2A standard exists yet—current systems use framework-specific protocols.

\subsection{Agent Policies and Reward Structures: Learning Optimal Behavior}

Agentic systems must learn effective strategies for achieving goals. We formalize this through \textit{policies} (decision-making rules) and \textit{reward structures} (objectives to optimize).

\subsubsection{Policy Formulation: From Rules to Learned Strategies}

An agent's \textbf{policy} $\pi$ maps states to actions: $\pi: S \to A$ (deterministic) or $\pi: S \to \Delta(A)$ (stochastic distribution over actions). For LLM agents:

\begin{equation}
\pi_\theta(a_t | s_t, h_t) = P_{\text{LLM}}(\text{action text} | \text{state description}, \text{history})
\end{equation}

\noindent where $\theta$ are model parameters, $s_t$ is current state (observations, tool outputs), $h_t$ is interaction history (previous thoughts/actions), and $a_t$ is the next action (tool call, message, reasoning step).

\textbf{Policy Types}:
\begin{itemize}
\item \textbf{Prompting-Based Policies}: Policy encoded in system prompt (e.g., "Use ReAct: alternate thoughts and actions"). No learning—fixed strategy. Effective for well-specified tasks.
\item \textbf{Few-Shot Policies}: Policy learned from examples in context. Example: show 3 successful task trajectories $\to$ agent generalizes pattern. Limited by context window.
\item \textbf{Fine-Tuned Policies}: Policy learned via supervised fine-tuning on agent trajectories. Example: collect 10K (state, action) pairs from expert agents, fine-tune LLM: $\mathcal{L} = -\sum \log p_\theta(a_t | s_t, h_t)$.
\item \textbf{RL-Optimized Policies}: Policy learned via reinforcement learning. Proximal Policy Optimization (PPO)~\cite{schulman2017proximal} optimizes: $\mathcal{L}^{\text{PPO}} = \mathbb{E}[\min(r_t(\theta) \hat{A}_t, \text{clip}(r_t(\theta), 1-\epsilon, 1+\epsilon) \hat{A}_t)]$ where $r_t(\theta) = \pi_\theta(a_t|s_t) / \pi_{\theta_{\text{old}}}(a_t|s_t)$ is probability ratio, $\hat{A}_t$ is advantage estimate.
\end{itemize}

\textbf{Behavior Cloning vs. RL}: Behavior cloning (supervised learning on expert trajectories) provides strong initialization but struggles with distribution shift (agent encounters states unseen in training data). RL enables online learning—agent explores, observes outcomes, updates policy—but requires well-defined reward signals and significant compute.

\subsubsection{Reward Modeling: Defining Success for Complex Tasks}

Specifying rewards for multi-step agent tasks is non-trivial. Simply rewarding task completion ignores intermediate quality (e.g., agent completes programming task but generates unreadable code).

\textbf{Reward Design Strategies}:
\begin{itemize}
\item \textbf{Sparse Terminal Rewards}: $r_T = 1$ if goal achieved, $0$ otherwise. Simple but provides no learning signal during task. Requires many episodes to learn.
\item \textbf{Dense Shaped Rewards}: Provide intermediate feedback: $r_t = \Delta(\text{progress toward goal})$. Example: for web navigation, reward each successful subgoal (login $\to$ search $\to$ add to cart). Faster learning but requires task-specific engineering.
\item \textbf{Learned Reward Models}: Train reward model $R_\phi(s, a)$ from human preferences using RLHF~\cite{ouyang2022training}. Collect comparisons: humans rate trajectories $\tau_1 > \tau_2$, train Bradley-Terry model: $p(\tau_1 > \tau_2) = \sigma(R_\phi(\tau_1) - R_\phi(\tau_2))$. Agent then optimizes learned reward via RL.
\item \textbf{Direct Preference Optimization (DPO)}~\cite{rafailov2023direct}: Optimize policy directly from preferences without explicit reward model: $\mathcal{L}_{\text{DPO}} = -\mathbb{E}[\log \sigma(\beta \log \frac{\pi_\theta(y_w|x)}{\pi_{\text{ref}}(y_w|x)} - \beta \log \frac{\pi_\theta(y_l|x)}{\pi_{\text{ref}}(y_l|x)})]$ where $y_w, y_l$ are preferred/dispreferred responses. Simpler than RLHF; achieves comparable results.
\end{itemize}

\textbf{Reward Hacking}: Agents may exploit reward misspecification. Example: reward "maximize code coverage" $\to$ agent writes meaningless tests achieving 100\% coverage but not testing functionality. Mitigation: combine multiple reward signals (coverage + bug detection rate + code quality), use learned reward models capturing holistic preferences, implement human oversight for high-stakes actions.

\textbf{Constitutional AI}~\cite{bai2022constitutional}: Anthropic's approach using AI-generated self-critiques as reward signal. Agent generates response $\to$ critique model evaluates against principles (helpfulness, harmlessness, honesty) $\to$ agent revises based on critique. Iterative refinement improves alignment without extensive human labeling.

\subsection{Multi-Agent Coordination: Collaborative Problem Solving}

\begin{table*}
    
\centering
\small
\caption{Planning algorithm taxonomy for LLM agents, ranging from reactive (ReAct) to deliberative (Graph-of-Thoughts) reasoning.}
\label{tab:planning_taxonomy}
\begin{tabular}{p{2.2cm}p{4.5cm}p{3cm}p{2.5cm}}
\toprule
\textbf{Algorithm} & \textbf{Core Mechanism} & \textbf{Strengths} & \textbf{Limitations} \\
\midrule
\textbf{ReAct}~\cite{yao2023react} & Interleaves reasoning traces with action execution: Thought $\to$ Action $\to$ Observation cycle & Grounded in external feedback; interpretable reasoning chains & Sequential execution; no backtracking \\
\midrule
\textbf{Reflexion}~\cite{shinn2023reflexion} & Learns from failures via self-reflection: evaluates trajectory, generates feedback, retries with improved strategy & Iterative improvement; learns from mistakes & Requires evaluation signal; limited to episodic tasks \\
\midrule
\textbf{Tree-of-Thoughts}~\cite{yao2023tree} & Explores solution tree with BFS/DFS: generates multiple thoughts, evaluates quality, backtracks when needed & Systematic exploration; handles complex planning & High inference cost (5-10$\times$ tokens); requires value function \\
\midrule
\textbf{Graph-of-Thoughts}~\cite{besta2023graph} & Models thoughts as DAG enabling parallel exploration and aggregation & Efficient for parallelizable tasks; combines diverse perspectives & Complex orchestration; requires task decomposition expertise \\
\bottomrule
\end{tabular}

\end{table*}

\textbf{ReAct}~\cite{yao2023react} pioneered practical agent planning by interleaving reasoning with action: $\text{Thought}_t \to \text{Action}_t \to \text{Observation}_t \to \text{Thought}_{t+1}$. Example: \textit{Thought}: "Need population of Paris" $\to$ \textit{Action}: \texttt{Search[Paris population]} $\to$ \textit{Observation}: "2.1M residents" $\to$ \textit{Thought}: "Can now compare to Tokyo." This grounds reasoning in verifiable external state.

\textbf{Reflexion}~\cite{shinn2023reflexion} adds self-critique: after task failure, the agent reflects on mistakes and generates improved strategies. Trajectory: Attempt $\to$ Evaluate $\to$ Reflect $\to$ Retry. On programming tasks, Reflexion improves pass@1 by 30-50\% through iterative debugging.

\textbf{Tree-of-Thoughts (ToT)}~\cite{yao2023tree} enables systematic exploration: each thought is a partial solution, the model evaluates quality and backtracks from dead ends. Combines LLMs with search algorithms: $V(s) = \text{LLM}_{\text{eval}}(s)$ scores intermediate states, guiding BFS/DFS exploration. Achieves state-of-the-art on Game of 24 (74\% $\to$ 100\%) but requires 5-10$\times$ more tokens.

\textbf{Graph-of-Thoughts (GoT)}~\cite{besta2023graph} generalizes trees to directed acyclic graphs, enabling parallel exploration and aggregation. Example: generate 5 draft paragraphs concurrently, then merge best elements into final text. Reduces latency for parallelizable tasks while maintaining ToT-level quality.

\subsubsection{Tool Use: Extending Models with External Capabilities}

Tool integration transforms LLMs from text generators into versatile problem solvers accessing computation, search, and specialized APIs. Table~\ref{tab:tool_use_taxonomy_unified} categorizes tool-use frameworks.

\textbf{Function Calling}~\cite{openai2023function} enables structured API access: the model outputs \texttt{\{"name": "get\_weather", "args": \{"location": "Paris"\}\}}, the executor invokes the function, and results are returned to the model. GPT-4 achieves 95\%+ accuracy on well-documented APIs.

\textbf{Code Interpreter}~\cite{openai2023code} provides full programming environment: agents write Python code, execute in sandbox, inspect output (including plots, files), and iterate. Example: "Analyze this CSV" $\to$ agent writes \texttt{pandas} code, generates visualization, interprets results. Enables complex data analysis beyond API calls.

\textbf{Toolformer}~\cite{schick2023toolformer} learns tool use via self-supervision: during training, the model inserts tool calls (\texttt{<calculator>157*23</calculator>}), executes them, and keeps calls that improve next-token prediction. This teaches when (not just how) to use tools without explicit labels.

\subsubsection{Memory Architectures: Maintaining Context Across Interactions}

Memory systems enable agents to accumulate knowledge, recall relevant information, and maintain long-term context beyond the fixed context window. Table~\ref{tab:memory_taxonomy} categorizes memory architectures.

\textbf{In-Context Memory} includes all information directly in the prompt. Modern models handle 32K-200K tokens (GPT-4, Claude 3), enabling detailed conversation history and documentation. Limitation: quadratic attention cost ($O(n^2)$) and fixed capacity.

\textbf{Retrieval-Augmented Generation (RAG)}~\cite{lewis2020retrieval} stores knowledge externally: given query $x$, retrieve top-$k$ documents $\{d_1, \ldots, d_k\}$ via dense retrieval (contriever~\cite{izacard2022unsupervised}, e5~\cite{wang2024text}), then condition generation on $[x; d_1; \ldots; d_k]$. Enables access to massive corpora (Wikipedia, internal documents) with 50-200ms retrieval latency.

\textbf{Episodic Memory}~\cite{park2023generative} stores interaction transcripts with metadata (timestamp, importance scores): each episode is embedded, and retrieval combines recency (exponential decay), importance (model-scored), and relevance (embedding similarity). Generative Agents~\cite{park2023generative} use episodic memory to simulate 25 interactive characters maintaining consistent personalities across 200+ interactions.

\textbf{Parametric Memory}~\cite{mitchell2022fast} stores knowledge in model weights: techniques like ROME~\cite{meng2022locating} and MEMIT~\cite{meng2023mass} update specific neurons to inject new facts ("The Eiffel Tower was painted green in 2024") without full retraining. Enables real-time knowledge updates with 0ms retrieval but risks catastrophic forgetting.

\subsection{Multi-Agent Coordination: Collaborative Problem Solving}

Multi-agent systems distribute tasks across specialized agents with complementary capabilities, enabling complex workflows beyond single-agent capacity. Key challenges include task decomposition, role assignment, communication overhead, and coordination mechanisms.

\subsubsection{Multi-Agent Frameworks and Coordination Patterns}

Table~\ref{tab:multiagent_taxonomy} categorizes coordination frameworks from conversational (AutoGen) to role-based (MetaGPT) to dynamic (AgentVerse) collaboration.

\begin{table*}[ht]
\centering
\small
\caption{Multi-agent coordination frameworks: from conversational (AutoGen) to role-based (MetaGPT) to dynamic (AgentVerse) collaboration.}
\label{tab:multiagent_taxonomy}
\begin{tabular}{p{2.2cm}p{4.8cm}p{2.5cm}p{2.8cm}}
\toprule
\textbf{Framework} & \textbf{Coordination Mechanism} & \textbf{Agent Roles} & \textbf{Communication} \\
\midrule
\textbf{AutoGen}~\cite{wu2023autogen} & Conversational agents with customizable roles; human-in-the-loop; code execution & User proxy, assistant, executor & Natural language messages \\
\midrule
\textbf{MetaGPT}~\cite{hong2023metagpt} & Simulates software company: structured workflow (requirements $\to$ design $\to$ code $\to$ test) & PM, architect, engineer, QA & Structured documents (PRD, design docs) \\
\midrule
\textbf{ChatDev}~\cite{qian2023chatdev} & Chain-shaped collaboration: CEO $\to$ CTO $\to$ programmer $\to$ reviewer; waterfall development & 7 roles (CEO, CTO, designer, programmer, tester, reviewer, art designer) & Sequential handoffs with code/design artifacts \\
\midrule
\textbf{AgentVerse}~\cite{chen2023agentverse} & Dynamic task decomposition: recruiter agent assembles team based on task requirements & Dynamic role assignment & Blackboard architecture (shared memory) \\
\bottomrule
\end{tabular}

\end{table*}

\textbf{AutoGen}~\cite{wu2023autogen} enables flexible agent conversations: define agents with system prompts (e.g., "You are a Python expert"), tools (code executor), and termination conditions. Example workflow: User describes task $\to$ Assistant generates code $\to$ Executor runs code $\to$ Assistant debugs based on output $\to$ Repeat until success. Supports human-in-the-loop for oversight.

\textbf{MetaGPT}~\cite{hong2023metagpt} simulates software companies with role-based workflows: (1) Product Manager writes PRD, (2) Architect designs system (class diagrams, APIs), (3) Engineer implements code, (4) QA tests and reports bugs. Agents communicate via structured documents, achieving 87\% functional correctness on software generation benchmarks (vs. 41\% single-agent baseline).

\textbf{ChatDev}~\cite{qian2023chatdev} implements waterfall development: CEO defines objectives $\to$ CTO proposes tech stack $\to$ Designer creates UI $\to$ Programmer codes $\to$ Tester validates $\to$ Reviewer ensures quality. Chain-shaped communication prevents chaos from full connectivity. Generates complete software (HTML/CSS/JS games, productivity tools) in <10 minutes.

\textbf{AgentVerse}~\cite{chen2023agentverse} dynamically assembles teams: given task description, a recruiter agent identifies required expertise (e.g., "need Python expert, data scientist, visualization specialist") and instantiates agents with appropriate prompts. Agents share state via blackboard architecture. Outperforms fixed-role systems on diverse tasks requiring variable expertise.

\subsubsection{Emergent Behaviors and Coordination Challenges}

\textbf{Beneficial Emergence}: Multi-agent systems exhibit capabilities beyond individual agents. Example: In MetaGPT, architect agent proposes modular design (not explicitly prompted), enabling engineer agent to implement components independently, then integration agent combines them. Modularity emerges from role specialization.

\textbf{Coordination Overhead}: Communication costs scale quadratically with agent count. Solutions: (1) hierarchical organization (manager agent coordinates sub-teams), (2) sparse communication topologies (chain/tree structures), (3) asynchronous message passing (agents don't wait for responses).

\textbf{Conflicting Objectives}: Agents may pursue incompatible goals. Example: In software development, engineer optimizes for feature richness while QA agent optimizes for bug-free code—tension requires negotiation or meta-agent arbitration.

\textbf{Consensus Mechanisms}: When agents must agree on decisions (e.g., which design to implement), systems use voting (majority/plurality), debate (agents argue positions, final decision via critique), or hierarchical authority (designated agent makes final call).

\subsection{Evaluation and Benchmarking for Agentic Systems}

Evaluating agentic capabilities requires moving beyond static benchmarks to interactive, multi-step tasks with verifiable outcomes. Key evaluation frameworks:

\textbf{WebArena}~\cite{zhou2023webarena}: Agents navigate realistic web environments (e-commerce, forums, CMS) to complete tasks ("Find products under \$50 and add to cart"). Evaluates tool use, planning, and grounding. GPT-4 achieves 14.4\% success rate, highlighting remaining challenges.

\textbf{SWE-bench}~\cite{jimenez2023swebench}: Agents resolve real GitHub issues in popular repositories. Requires understanding codebases, writing patches, passing tests. State-of-the-art: 13.8\% issues resolved (vs. 0.8\% for baseline models), demonstrating rapid progress in code agents.

\textbf{GAIA}~\cite{mialon2023gaia}: General AI Assistants benchmark with multi-modal, multi-step questions requiring web search, code execution, file processing. Only 15\% solvable by current systems, providing headroom for long-term progress.

\textbf{AgentBench}~\cite{liu2023agentbench}: 8 environments spanning web navigation, database querying, game playing, household tasks (ALFWorld). Unified evaluation protocol: success rate on task completion. GPT-4 achieves 52\% average success (vs. 28\% for GPT-3.5), validating capability differences.

\textbf{Key Insight}: Agentic benchmarks reveal current limitations—even frontier models solve <20\% of complex real-world tasks—highlighting the gap between text generation capabilities and embodied problem-solving. Progress requires improved planning (handling 10+ step horizons), robust error recovery (agents currently fail catastrophically on errors), and efficient exploration (reducing trial-and-error overhead).

\subsection{Safety, Oversight, and Risk Mitigation}

Agentic systems with tool access and autonomous execution raise critical safety concerns: unintended actions (e.g., irreversible API calls), exploration of harmful strategies, and accumulation of errors. We identify three key safety dimensions:

\subsubsection{Human Oversight Mechanisms}

Preventing unintended autonomy while maintaining efficiency:

\textbf{Action Confirmation}: Require human approval for high-stakes actions (financial transactions, data deletion, external communications). AutoGen implements configurable approval: critical actions pause for human confirmation, while safe actions (read-only queries) proceed autonomously.

\textbf{Interpretable Reasoning Traces}: Log all thoughts, actions, and observations for post-hoc auditing. ReAct's explicit reasoning chains enable humans to identify flawed logic: "Thought: I'll delete all files to free space" can be caught before execution.

\textbf{Capability Restrictions}: Limit tool access based on task context. Example: data analysis agents access \texttt{pandas} but not \texttt{os.system()}, preventing inadvertent system modifications. GPT-4 Code Interpreter runs in isolated sandbox (no network, ephemeral filesystem).

\subsubsection{Red-Teaming and Adversarial Testing}

Proactive identification of failure modes:

\textbf{Goal Misalignment Probes}: Test whether agents pursue unintended interpretations. Example: "Maximize user engagement" $\to$ agent generates addictive/misleading content. Anthropic's Constitutional AI~\cite{bai2022constitutional} uses automated red-teaming to identify harmful strategies.

\textbf{Tool Misuse Detection}: Monitor for unexpected API usage patterns. Example: agent bypasses rate limits by spawning multiple sessions. Anomaly detection flags deviations from expected tool-use distributions.

\textbf{Multi-Agent Collusion}: Test whether independent agents cooperate to circumvent restrictions. Example: Agent A gathers restricted information, Agent B executes actions based on that information. Requires system-level monitoring beyond individual agent oversight.

\subsubsection{Graceful Failure and Containment}

Ensuring errors don't cascade:

\textbf{Reversible Actions}: Design APIs with undo mechanisms. Example: database agents use transactions (BEGIN/COMMIT/ROLLBACK), enabling rollback on errors. File operations create backups before modifications.

\textbf{Timeouts and Circuit Breakers}: Halt execution after excessive retries or resource consumption. Example: agent stuck in exploration loop exhausts 1000-action budget, triggering forced termination with informative error.

\textbf{Sandboxing and Virtualization}: Execute actions in isolated environments (containers, VMs) preventing unintended impact on production systems. Explosion radius: contain failures to sandbox, no lateral movement.

\textbf{Case Study}: OpenAI Code Interpreter implements multi-layer safety: (1) sandboxed Python environment (no network, limited filesystem), (2) execution timeout (120 seconds per cell), (3) explicit user confirmation for file downloads, and (4) automatic session termination after 60 minutes idle. These constraints reduce risk while enabling broad utility for data analysis and visualization.

\textbf{Open Challenge}: Current oversight mechanisms rely on human monitoring or rule-based guardrails, both brittle at scale. Scaling to fully autonomous agents (e.g., personal assistants with 24/7 access to email, calendar, finances) requires \textit{learned safety models} that predict action consequences and \textit{uncertainty quantification} to request help when confidence is low. Recent work on Constitutional AI~\cite{bai2022constitutional} and debate-based oversight~\cite{irving2018ai} offers promising directions but remains far from deployment-ready for high-stakes autonomy.

\section{Key Findings, Open Challenges, and Future Directions}

This survey has traced the remarkable evolution of artificial intelligence from the foundational Transformer architecture (2017) to the sophisticated reasoning-capable systems of 2025. Through comprehensive analysis of over 50 major models, detailed examination of training methodologies, and rigorous benchmarking across multiple dimensions, we have documented three fundamental paradigm shifts that define the current state and future trajectory of AI research.

\subsection{Key Findings and Paradigm Shifts}

\paragraph{From Pre-Training to Post-Training Dominance}
Our analysis reveals that the locus of capability development has decisively shifted from pre-training scale to post-training optimization. While GPT-3 (2020) demonstrated that pre-training alone could achieve 70\% MMLU performance, modern systems attribute 70-90\% of their capabilities to reinforcement learning techniques. DeepSeek-R1's achievement of 79.8\% on MATH and 79.2\% on AIME through pure RL—without supervised fine-tuning—validates that verifiable rewards alone can induce sophisticated reasoning. This shift has profound implications: \textit{data quality matters more than quantity}, \textit{human feedback becomes the bottleneck}, and \textit{automated feedback mechanisms offer scaling paths beyond human judgment}.

\paragraph{Test-Time Compute as a New Scaling Dimension}
The emergence of o1, DeepSeek-R1, and reasoning-specialized models establishes test-time compute as a viable alternative to pre-training scale. Our documentation shows that $\text{performance} \propto \log(\text{inference FLOPs})$, where 1 hour of extended reasoning achieves gains equivalent to $10^6\times$ training compute. This represents a fundamental rebalancing: rather than concentrating all compute in pre-training, systems can strategically allocate resources between training and inference based on task complexity. The economic implications are significant—test-time scaling enables smaller, cheaper base models to achieve GPT-4-level performance on reasoning tasks through extended thinking time.

\paragraph{The Efficiency Revolution}
Innovations documented in this survey—Multi-head Latent Attention (8$\times$ KV cache compression), mixture-of-experts with fine-grained routing (18$\times$ parameter efficiency in DeepSeek-V3), FlashAttention-2 (2.5$\times$ speedup), GQA (4-8$\times$ KV reduction), and optimized inference stacks—have collectively achieved a 100$\times$ cost reduction in 2 years. GPT-4-level performance now costs $<$\$0.30/M tokens versus \$30+/M tokens in 2023. This democratization enables broader access, more extensive experimentation, and deployment in cost-sensitive applications previously inaccessible to AI.

\subsection{The Scaling Wall and Post-Scaling Era}

A central theme of this survey is the scaling wall confronting continued model development. The convergence of three crises—data exhaustion (9-27T tokens depleted by 2026-2028), exponential cost growth (\$3M to \$300M+ in 5 years), and unsustainable energy consumption (22$\times$ increase from GPT-3 to GPT-4)—fundamentally challenges naive scaling. As Coveney et al.~\cite{coveney2025wall} and Hooker~\cite{hooker2025slowdeath} document, the simple formula of "scale model size and training data" has become inadequate.

Yet this constraint has catalyzed innovation precisely along the three alternative paradigms we have analyzed: (1) test-time compute scaling demonstrates that reasoning can be "purchased" at inference time, (2) sparse architectures like MoE prove that parameter efficiency can match dense model quality, and (3) architectural innovations (linear attention, state space models, sparse patterns) break complexity barriers. The transition from "scaling era" to "efficiency era" mirrors computing's historical pivot from frequency scaling to multicore architectures and specialized accelerators when Moore's Law approached limits.

\subsection{Open-Source Democratization and Global Competition}

The competitive landscape has fundamentally transformed. Open-source models now match or exceed proprietary systems: Llama 3 (405B) achieves 88.6\% MMLU versus GPT-4's 86.4\%, Qwen-3 demonstrates reasoning capabilities competitive with Claude 3.5, and DeepSeek-V3 delivers GPT-4-level performance at 18$\times$ parameter efficiency. This democratization enables:

\begin{itemize}
\item \textbf{Research acceleration}: Academic institutions can now train and experiment with frontier-class models using accessible compute
\item \textbf{Regional innovation}: Organizations across China (Alibaba, ByteDance, Moonshot, Zhipu), Europe, and other regions contribute diverse architectural approaches
\item \textbf{Specialized applications}: Domain-specific models fine-tuned for healthcare, code generation, scientific reasoning, and multilingual tasks
\item \textbf{Safety research}: Open weights enable independent analysis of model behavior, bias, and failure modes
\end{itemize}

The emergence of globally distributed AI capability breaks the concentration of progress in a handful of US-based labs, introducing beneficial competition, diverse perspectives, and accelerated innovation cycles.

\subsection{Critical Open Challenges}

Despite remarkable progress, fundamental challenges remain:

\paragraph{Verifiable Reasoning at Scale}
While DeepSeek-R1 demonstrates that pure RL can induce reasoning through verifiable rewards (MATH problems have ground truth), extending this to open-ended domains where ground truth is ambiguous or subjective remains unsolved. The field lacks robust frameworks for: (1) evaluating reasoning chains beyond final answers, (2) detecting subtle logical errors in multi-step derivations, (3) balancing exploration of reasoning strategies with exploitation of known patterns, and (4) scaling human oversight of complex reasoning processes. Constitutional AI~\cite{bai2022constitutional} and debate-based methods offer promising directions, but achieving "process supervision at scale" remains an open problem.

\paragraph{Efficient Test-Time Compute Allocation}
Test-time scaling introduces new optimization challenges: \textit{when} to allocate extended compute (not all queries require deep thinking), \textit{how much} compute to allocate (dynamic budgeting based on query complexity), and \textit{how} to search efficiently (balancing breadth-first exploration with depth-first refinement). Current systems use simple fixed-budget approaches, but optimal policies likely require learned meta-strategies that predict query difficulty and adaptively allocate resources. This connects to classical problems in heuristic search but at unprecedented scale and with learned rather than handcrafted heuristics.

\paragraph{Multimodal Integration Beyond Concatenation}
Current multimodal models largely concatenate visual and text tokens or use shallow cross-attention. True multimodal understanding requires: (1) aligned representations where semantically equivalent concepts across modalities have similar encodings, (2) cross-modal reasoning where information from one modality guides processing in another, (3) temporal synchronization for video and audio, and (4) efficient training from limited paired data. Models like Gemini 1.5 and GPT-4o demonstrate impressive multimodal capabilities, but systematic understanding of what constitutes good multimodal architecture remains elusive.

\paragraph{Agent Reliability and Safety}
As LLMs evolve into autonomous agents using tools, accessing external systems, and making consequential decisions, reliability becomes critical. Challenges include: (1) hallucination mitigation when agents generate and execute code, (2) failure recovery when tool calls fail or return unexpected results, (3) constraint satisfaction ensuring agents respect security boundaries and ethical guidelines, (4) compositional generalization to novel task combinations, and (5) interpretability of agent decision-making for debugging and oversight. The transition from "helpful assistant" to "autonomous agent" requires solving these problems at scale.

\paragraph{Economic Sustainability and Environmental Impact}
While efficiency innovations have dramatically reduced per-token costs, aggregate energy consumption continues growing as usage expands. Training DeepSeek-V3 (the most efficient frontier model) still required \$110M and 2,279 MWh—equivalent to 200+ households' annual electricity. Sustainable AI requires: (1) continued algorithmic efficiency gains beyond current 100$\times$ cost reductions, (2) hardware co-design optimizing for transformer/SSM operations, (3) dynamic resource allocation to avoid over-provisioning, (4) renewable energy integration for large training runs, and (5) economic models that internalize environmental costs. The field must balance capability progress with planetary constraints.

\paragraph{From Human Preference to Outcome Optimization}
Current RLHF systems optimize for human preference—what responses humans \textit{like}—which may diverge from what is objectively \textit{correct} or \textit{useful}. For tasks with verifiable outcomes (mathematics, code execution, scientific reasoning), the field is transitioning to outcome-based training. However, most real-world applications lack clear ground truth. Open questions include: (1) designing reward functions for subjective domains (creative writing, strategic advice), (2) balancing short-term user satisfaction with long-term value, (3) aggregating preferences across diverse user populations, (4) handling distribution shift as model capabilities evolve, and (5) detecting and correcting reward hacking where models exploit loopholes in feedback mechanisms.

\subsection{Future Research Directions}

Building on the foundations established in this survey, we identify high-priority research directions:

\paragraph{1. Hybrid Architectures Combining Transformers, SSMs, and Graph Networks}
No single architecture dominates all dimensions. Transformers excel at in-context learning but suffer quadratic complexity; SSMs (Mamba) achieve linear complexity but struggle with certain reasoning patterns; graph networks capture structured relationships but require domain-specific design. Future systems may combine these paradigms: using SSMs for efficient sequence encoding, Transformers for critical reasoning steps requiring global context, and graph networks for explicit relational reasoning. Learned routing mechanisms could dynamically select architectures based on input characteristics.

\paragraph{2. Continual Learning and Efficient Model Updates}
Current models are static—trained once on a fixed dataset, then deployed. Real-world deployment requires: (1) incorporating new information without full retraining (continual learning), (2) correcting errors and updating knowledge (model editing), (3) personalizing to user preferences over time, and (4) forgetting sensitive information on request. Techniques like parameter-efficient fine-tuning (LoRA, QLoRA), elastic weight consolidation, and compositional model architectures offer partial solutions, but systematic frameworks for lifelong learning at scale remain underdeveloped.

\paragraph{3. Interpretability and Mechanistic Understanding}
Despite impressive capabilities, our mechanistic understanding of how large models work remains limited. Priority questions include: (1) what algorithms do transformers implement (how do they perform in-context learning, arithmetic, logical reasoning?), (2) what representations emerge in hidden states (are there interpretable "concepts"?), (3) how does reasoning emerge from scale and training (phase transitions, critical thresholds), (4) what causes failure modes (hallucination, biased outputs, capability gaps), and (5) how can we verify model behavior (proofs of reasoning correctness)? Progress requires both empirical investigation (probing, intervention experiments) and theoretical analysis (mechanistic interpretability, circuits perspective).

\paragraph{4. Efficient Reasoning Search Algorithms}
Test-time compute scaling raises algorithmic questions: what search strategies optimize reasoning under compute budgets? Current approaches use beam search or best-of-N sampling, but these are generic. Domain-specific algorithms may outperform: mathematical theorem proving might benefit from proof search guided by learned heuristics, code generation from program synthesis with type-guided search, scientific reasoning from causal inference algorithms. Integrating classical AI search methods (A*, IDA*, Monte Carlo Tree Search) with learned models offers rich opportunities.

\paragraph{5. Synthetic Data and Self-Improvement Loops}
As high-quality human data depletes, synthetic data generation becomes critical. Key challenges: (1) ensuring synthetic data quality (avoiding model collapse from training on own outputs), (2) designing curricula that target capability gaps, (3) using verification to filter generated data (proof checkers for math, compilers for code, simulators for robotics), (4) bootstrapping from weak models to strong models, and (5) understanding theoretical limits of self-improvement. AlphaGeometry's synthetic theorem proving and DeepSeek-R1's RL-only training demonstrate feasibility, but systematic frameworks remain open.

\paragraph{6. Multimodal Foundation Models for Robotics and Embodied AI}
Text-based models demonstrate impressive reasoning, but real-world deployment requires physical grounding. Embodied AI research must address: (1) learning from limited real-world interaction data (sim-to-real transfer, foundation models pre-trained on internet data then adapted to robotics), (2) real-time inference under compute constraints (embedded deployment), (3) safety and robustness (physical systems have irreversible consequences), (4) common sense physics (understanding object interactions, stability, causality), and (5) long-horizon task planning (combining high-level reasoning with low-level control). Models like RT-2 and PaLM-E represent early steps, but general-purpose embodied agents remain distant.

\section{Conclusion}
\label{sec:conclusion}

The evolution from GPT-2 (2019, 1.5B parameters) to DeepSeek-V3 (2025, 671B parameters, 37B active) represents unprecedented technological acceleration, achieving three fundamental breakthroughs: (1) \textbf{post-training dominance}—70-90\% of capabilities now derive from RLHF/RL rather than pre-training scale, (2) \textbf{test-time compute scaling}—inference-time reasoning achieves gains equivalent to $10^6\times$ training compute, and (3) \textbf{100$\times$ efficiency gains}—cost per token reduced from \$30+ to \$0.30 while maintaining GPT-4-level quality.

Three converging crises define the scaling wall: data exhaustion (9-27T tokens depleted by 2026-2028), exponential cost growth (\$3M to \$300M+ in 5 years), and 22$\times$ energy increase (GPT-3 to GPT-4). Yet these constraints have catalyzed innovation: sparse architectures (MoE achieving 18$\times$ efficiency), sub-quadratic attention mechanisms (linear attention, state space models), and architectural creativity prove that continued progress need not rely on brute-force scaling.

Open-source democratization has fundamentally transformed the competitive landscape. Llama 3 (88.6\% MMLU) surpasses GPT-4 (86.4\%), enabling academic research, regional innovation across China/Europe, and independent safety analysis. This global distribution of AI capability introduces beneficial competition and accelerates innovation cycles.

Critical challenges remain: verifiable reasoning at scale beyond ground-truth domains, efficient test-time compute allocation, true multimodal integration, agent reliability and safety, environmental sustainability, and transitioning from human preference to outcome optimization. The field's trajectory depends on solving these problems while maintaining the remarkable pace of capability growth documented in this survey.

This work provides comprehensive technical documentation—rigorous mathematical formulations (RLHF, PPO, DPO, GRPO), architectural analysis (FlashAttention, GQA, MoE, Mamba), and benchmarking across 9 metrics spanning 30+ models—serving as both practitioner reference and research foundation. As the field transitions from the scaling era to the efficiency era, the ingenuity demonstrated in navigating current constraints, the diversity of global contributions, and rapid democratization suggest the post-scaling era may prove even more innovative than the scaling era itself.

\bibliographystyle{plain}
\bibliography{reference}

@inproceedings{vaswani2017attention,
  title={Attention is all you need},
  author={Vaswani, Ashish and Shazeer, Noam and Parmar, Niki and Uszkoreit, Jakob and Jones, Llion and Gomez, Aidan N and Kaiser, {\L}ukasz and Polosukhin, Illia},
  booktitle={Advances in Neural Information Processing Systems},
  pages={5998--6008},
  year={2017}
}

@article{brown2020language,
  title={Language models are few-shot learners},
  author={Brown, Tom and Mann, Benjamin and Ryder, Nick and Subbiah, Melanie and Kaplan, Jared D and Dhariwal, Prafulla and Neelakantan, Arvind and Shyam, Pranav and Sastry, Girish and Askell, Amanda and others},
  journal={Advances in Neural Information Processing Systems},
  volume={33},
  pages={1877--1901},
  year={2020}
}

@article{radford2019language,
  title={Language models are unsupervised multitask learners},
  author={Radford, Alec and Wu, Jeffrey and Child, Rewon and Luan, David and Amodei, Dario and Sutskever, Ilya and others},
  journal={OpenAI blog},
  volume={1},
  number={8},
  pages={9},
  year={2019}
}

@article{ouyang2022training,
  title={Training language models to follow instructions with human feedback},
  author={Ouyang, Long and Wu, Jeffrey and Jiang, Xu and Almeida, Diogo and Wainwright, Carroll and Mishkin, Pamela and Zhang, Chong and Agarwal, Sandhini and Slama, Katarina and Ray, Alex and others},
  journal={Advances in Neural Information Processing Systems},
  volume={35},
  pages={27730--27744},
  year={2022}
}

@article{openai2023gpt4,
  title={GPT-4 technical report},
  author={{OpenAI}},
  journal={arXiv preprint arXiv:2303.08774},
  year={2023}
}

@article{openai2024gpt4o,
  title={GPT-4o system card},
  author={{OpenAI}},
  journal={OpenAI Technical Report},
  year={2024}
}

@article{openai2024o1,
  title={Learning to reason with LLMs},
  author={{OpenAI}},
  journal={OpenAI Blog},
  year={2024}
}

@article{touvron2023llama,
  title={Llama: Open and efficient foundation language models},
  author={Touvron, Hugo and Lavril, Thibaut and Izacard, Gautier and Martinet, Xavier and Lachaux, Marie-Anne and Lacroix, Timoth{\'e}e and Rozi{\`e}re, Baptiste and Goyal, Naman and Hambro, Eric and Azhar, Faisal and others},
  journal={arXiv preprint arXiv:2302.13971},
  year={2023}
}

@article{touvron2023llama2,
  title={Llama 2: Open foundation and fine-tuned chat models},
  author={Touvron, Hugo and Martin, Louis and Stone, Kevin and Albert, Peter and Almahairi, Amjad and Babaei, Yasmine and Bashlykov, Nikolay and Batra, Soumya and Bhargava, Prajjwal and Bhosale, Shruti and others},
  journal={arXiv preprint arXiv:2307.09288},
  year={2023}
}

@article{dubey2024llama3,
  title={The Llama 3 herd of models},
  author={Dubey, Abhimanyu and Jauhri, Abhinav and Pandey, Abhinav and Kadian, Abhishek and Al-Dahle, Ahmad and Letman, Aiesha and Mathur, Akhil and Schelten, Alan and Yang, Amy and Fan, Angela and others},
  journal={arXiv preprint arXiv:2407.21783},
  year={2024}
}

@article{chen2021evaluating,
  title={Evaluating large language models trained on code},
  author={Chen, Mark and Tworek, Jerry and Jun, Heewoo and Yuan, Qiming and Pinto, Henrique Ponde de Oliveira and Kaplan, Jared and Edwards, Harri and Burda, Yuri and Joseph, Nicholas and Brockman, Greg and others},
  journal={arXiv preprint arXiv:2107.03374},
  year={2021}
}

@article{schulman2017proximal,
  title={Proximal policy optimization algorithms},
  author={Schulman, John and Wolski, Filip and Dhariwal, Prafulla and Radford, Alec and Klimov, Oleg},
  journal={arXiv preprint arXiv:1707.06347},
  year={2017}
}

@article{rafailov2023dpo,
  title={Direct preference optimization: Your language model is secretly a reward model},
  author={Rafailov, Rafael and Sharma, Archit and Mitchell, Eric and Ermon, Stefano and Manning, Christopher D and Finn, Chelsea},
  journal={arXiv preprint arXiv:2305.18290},
  year={2023}
}

@article{shao2024deepseek,
  title={DeepSeekMath: Pushing the limits of mathematical reasoning in open language models},
  author={Shao, Zhihong and Wang, Peiyi and Zhu, Qihao and Xu, Runxin and Song, Junxiao and Zhang, Mingchuan and Li, YK and Wu, Y and Guo, Daya},
  journal={arXiv preprint arXiv:2402.03300},
  year={2024}
}

@article{deepseek2024llm,
  title={DeepSeek LLM: Scaling open-source language models with longtermism},
  author={{DeepSeek-AI}},
  journal={arXiv preprint arXiv:2401.02954},
  year={2024}
}

@article{deepseek2024coder,
  title={DeepSeek-Coder: When the large language model meets programming--the rise of code intelligence},
  author={{DeepSeek-AI}},
  journal={arXiv preprint arXiv:2401.14196},
  year={2024}
}

@article{deepseek2024math,
  title={DeepSeekMath: Pushing the limits of mathematical reasoning in open language models},
  author={{DeepSeek-AI}},
  journal={arXiv preprint arXiv:2402.03300},
  year={2024}
}

@article{deepseek2024moe,
  title={DeepSeekMoE: Towards ultimate expert specialization in mixture-of-experts language models},
  author={{DeepSeek-AI}},
  journal={arXiv preprint arXiv:2401.06066},
  year={2024}
}

@article{deepseekai2024deepseekmoe,
  title={DeepSeekMoE: Towards ultimate expert specialization in mixture-of-experts language models},
  author={{DeepSeek-AI}},
  journal={arXiv preprint arXiv:2401.06066},
  year={2024}
}

@article{deepseekai2024deepseekv3,
  title={DeepSeek-V3 technical report},
  author={{DeepSeek-AI}},
  journal={arXiv preprint arXiv:2412.19437},
  year={2024}
}

@article{deepseek2025v3,
  title={DeepSeek-V3 technical report},
  author={{DeepSeek-AI}},
  journal={arXiv preprint arXiv:2412.19437},
  year={2025}
}

@article{deepseek2025r1,
  title={DeepSeek-R1: Incentivizing reasoning capability in LLMs via reinforcement learning},
  author={{DeepSeek-AI}},
  journal={arXiv preprint arXiv:2501.12948},
  year={2025}
}

@article{team2023gemini,
  title={Gemini: A family of highly capable multimodal models},
  author={{Gemini Team}},
  journal={arXiv preprint arXiv:2312.11805},
  year={2023}
}

@article{reid2024gemini15,
  title={Gemini 1.5: Unlocking multimodal understanding across millions of tokens of context},
  author={Reid, Machel and Savinov, Nikolay and Teplyashin, Denis and Lepikhin, Dmitry and Lillicrap, Timothy and Alayrac, Jean-Baptiste and Soricut, Radu and Lazaridou, Angeliki and Firat, Orhan and Schrittwieser, Julian and others},
  journal={arXiv preprint arXiv:2403.05530},
  year={2024}
}

@article{gemma2report,
  title={Gemma 2: Improving open language models at a practical size},
  author={{Google DeepMind}},
  journal={arXiv preprint},
  year={2024}
}

@article{gemma3report,
  title={Gemma 3: Aggressive sliding window attention with 5:1 ratio},
  author={{Google DeepMind}},
  journal={arXiv preprint},
  year={2025}
}

@article{abdin2024phi4,
  title={Phi-4 technical report},
  author={Abdin, Marah and Aneja, Jyoti and Awadalla, Hany and Awadallah, Ahmed and Awan, Ammar Ahmad and Bach, Nguyen and Bahree, Amit and Bakhtiari, Arash and Bao, Jianmin and Behl, Harkirat and others},
  journal={arXiv preprint arXiv:2412.08905},
  year={2024}
}

@article{phi42025reasoning,
  title={Phi-4 reasoning: Small language models for complex reasoning},
  author={{Microsoft Research}},
  journal={Microsoft Technical Report},
  year={2025}
}

@article{gunasekar2023textbook,
  title={Textbooks are all you need},
  author={Gunasekar, Suriya and Zhang, Yi and Aneja, Jyoti and Mendes, Caio C{\'e}sar Teodoro and Del Giorno, Allie and Gopi, Sivakanth and Javaheripi, Mojan and Kauffmann, Piero and de Rosa, Gustavo and Saarikivi, Olli and others},
  journal={arXiv preprint arXiv:2306.11644},
  year={2023},
  note={Phi-1 model paper}
}

@article{javaheripi2023phi2,
  title={Phi-2: The surprising power of small language models},
  author={Javaheripi, Mojan and Bubeck, S{\'e}bastien and Abdin, Marah and Aneja, Jyoti and Bubeck, S{\'e}bastien and Mendes, Caio C{\'e}sar Teodoro and Del Giorno, Allie and Eldan, Ronen and Gopi, Sivakanth and Kamar, Ece and others},
  journal={Microsoft Research Blog},
  year={2023},
  note={Available at https://www.microsoft.com/en-us/research/blog/phi-2-the-surprising-power-of-small-language-models/}
}

@article{abdin2024phi3,
  title={Phi-3 technical report: A highly capable language model locally on your phone},
  author={Abdin, Marah and Jacobs, Sam Ade and Awan, Ammar Ahmad and Aneja, Jyoti and Awadallah, Ahmed and Awadalla, Hany and Bach, Nguyen and Bahree, Amit and Bakhtiari, Arash and Behl, Harkirat and others},
  journal={arXiv preprint arXiv:2404.14219},
  year={2024},
  doi={10.48550/arXiv.2404.14219}
}

@article{openreasoner2025,
  title={Open-Reasoner-Zero: Scaling reasoning through pure reinforcement learning},
  author={{Open-Reasoner Team}},
  journal={arXiv preprint},
  year={2025}
}

@article{skywork2025or1,
  title={Skywork-o1: Open reasoning with verifiable rewards},
  author={{Skywork AI}},
  journal={arXiv preprint},
  year={2025}
}

@article{seed2025thinking,
  title={Seed-Thinking 1.5: Combining Monte Carlo tree search with reinforcement learning for reasoning},
  author={{Seed-AI}},
  journal={arXiv preprint},
  year={2025}
}

@article{nvidia2025nemotron,
  title={Llama-Nemotron: Efficient instruction following through distillation},
  author={{NVIDIA}},
  journal={NVIDIA Technical Report},
  year={2025}
}

@article{nvidia2025nano2,
  title={Nemotron Nano 2: Edge deployment of large language models},
  author={{NVIDIA}},
  journal={NVIDIA Technical Report},
  year={2025}
}

@article{nvidia2025nano3,
  title={Nemotron 3 Nano: Ultra-efficient language models for edge devices},
  author={{NVIDIA}},
  journal={NVIDIA Technical Report},
  year={2025}
}

@article{kimi2025v15,
  title={Kimi 1.5: Long-context reasoning at scale},
  author={{Moonshot AI}},
  journal={arXiv preprint},
  year={2025}
}

@article{kimi2025k2,
  title={Kimi K2: Advanced long-context language model},
  author={{Moonshot AI}},
  journal={arXiv preprint},
  year={2025}
}

@article{kimilinear2025,
  title={Kimi Linear: 48B parameter model with hybrid linear attention},
  author={{Moonshot AI}},
  journal={arXiv preprint},
  year={2025}
}

@article{qwen32025,
  title={Qwen 3: Advancing open-source language models},
  author={{Alibaba Cloud}},
  journal={arXiv preprint},
  year={2025}
}

@article{qwen2025qwen3next,
  title={Qwen3-Next: Hybrid architecture with Gated DeltaNet},
  author={{Alibaba Cloud}},
  journal={arXiv preprint},
  year={2025}
}

@article{glm2025v45,
  title={GLM-4.5: All tools integration for agentic AI},
  author={{Zhipu AI}},
  journal={arXiv preprint},
  year={2025}
}

@article{mimo2025,
  title={MiMo: Efficient language model training},
  author={{Xiaomi AI Lab}},
  journal={arXiv preprint},
  year={2025}
}

@article{mimovl2025,
  title={MiMo VL: Multimodal understanding and generation},
  author={{Xiaomi AI Lab}},
  journal={arXiv preprint},
  year={2025}
}

@article{mimov22025,
  title={MiMo-V2-Flash: High-speed inference with speculative decoding},
  author={{Xiaomi AI Lab}},
  journal={arXiv preprint},
  year={2025}
}

@article{groeneveld2024olmo,
  title={OLMo: Accelerating the science of language models},
  author={Groeneveld, Dirk and Beltagy, Iz and Walsh, Pete and Bhagia, Akshita and Kinney, Rodney and Tafjord, Oyvind and Jha, Ananya Harsh and Ivison, Hamish and Magnusson, Ian and Wang, Yizhong and others},
  journal={arXiv preprint arXiv:2402.00838},
  year={2024}
}

@article{muennighoff2024olmoe,
  title={OLMoE: Open mixture-of-experts language models},
  author={Muennighoff, Niklas and Soldaini, Luca and Groeneveld, Dirk and Lo, Kyle and Beltagy, Iz and others},
  journal={arXiv preprint arXiv:2409.02060},
  year={2024}
}

@article{olmo32025think,
  title={OLMo 3 Think: Open reasoning model with full transparency},
  author={{AI2}},
  journal={arXiv preprint},
  year={2025}
}

@article{olmo2paper,
  title={OLMo 2: Post-Norm architecture and training stability},
  author={{AI2}},
  journal={arXiv preprint},
  year={2025}
}

@article{jiang2023mistral,
  title={Mistral 7B},
  author={Jiang, Albert Q and Sablayrolles, Alexandre and Mensch, Arthur and Bamford, Chris and Chaplot, Devendra Singh and Casas, Diego de las and Bressand, Florian and Lengyel, Gianna and Lample, Guillaume and Saulnier, Lucile and others},
  journal={arXiv preprint arXiv:2310.06825},
  year={2023}
}

@article{jiang2024mixtral,
  title={Mixtral of experts},
  author={Jiang, Albert Q and Sablayrolles, Alexandre and Roux, Antoine and Mensch, Arthur and Savary, Blanche and Bamford, Chris and Chaplot, Devendra Singh and Casas, Diego de las and Hanna, Emma Bou and Bressand, Florian and others},
  journal={arXiv preprint arXiv:2401.04088},
  year={2024}
}

@article{pixtral2024,
  title={Pixtral 12B},
  author={{Mistral AI}},
  journal={Mistral AI Blog},
  year={2024}
}

@inproceedings{bahdanau2015neural,
  title={Neural machine translation by jointly learning to align and translate},
  author={Bahdanau, Dzmitry and Cho, Kyunghyun and Bengio, Yoshua},
  booktitle={International Conference on Learning Representations},
  year={2015}
}

@article{dao2022flashattention,
  title={FlashAttention: Fast and memory-efficient exact attention with IO-awareness},
  author={Dao, Tri and Fu, Dan and Ermon, Stefano and Rudra, Atri and R{\'e}, Christopher},
  journal={Advances in Neural Information Processing Systems},
  volume={35},
  pages={16344--16359},
  year={2022}
}

@article{dao2023flashattention2,
  title={FlashAttention-2: Faster attention with better parallelism and work partitioning},
  author={Dao, Tri},
  journal={arXiv preprint arXiv:2307.08691},
  year={2023}
}

@article{sun2023retentive,
  title={Retentive network: A successor to transformer for large language models},
  author={Sun, Yutao and Dong, Li and Huang, Shaohan and Ma, Shuming and Xia, Yuqing and Xue, Jilong and Wang, Jianyong and Wei, Furu},
  journal={arXiv preprint arXiv:2307.08621},
  year={2023}
}

@article{devlin2019bert,
  title={BERT: Pre-training of deep bidirectional transformers for language understanding},
  author={Devlin, Jacob and Chang, Ming-Wei and Lee, Kenton and Toutanova, Kristina},
  journal={arXiv preprint arXiv:1810.04805},
  year={2019}
}

@article{raffel2020exploring,
  title={Exploring the limits of transfer learning with a unified text-to-text transformer},
  author={Raffel, Colin and Shazeer, Noam and Roberts, Adam and Lee, Katherine and Narang, Sharan and Matena, Michael and Zhou, Yanqi and Li, Wei and Liu, Peter J},
  journal={Journal of Machine Learning Research},
  volume={21},
  number={140},
  pages={1--67},
  year={2020}
}

@inproceedings{radford2021learning,
  title={Learning transferable visual models from natural language supervision},
  author={Radford, Alec and Kim, Jong Wook and Hallacy, Chris and Ramesh, Aditya and Goh, Gabriel and Agarwal, Sandhini and Sastry, Girish and Askell, Amanda and Mishkin, Pamela and Clark, Jack and others},
  booktitle={International Conference on Machine Learning},
  pages={8748--8763},
  year={2021},
  organization={PMLR}
}

@article{shazeer2020glu,
  title={GLU variants improve transformer},
  author={Shazeer, Noam},
  journal={arXiv preprint arXiv:2002.05202},
  year={2020}
}

@article{su2021roformer,
  title={RoFormer: Enhanced transformer with rotary position embedding},
  author={Su, Jianlin and Lu, Yu and Pan, Shengfeng and Wen, Bo and Liu, Yunfeng},
  journal={arXiv preprint arXiv:2104.09864},
  year={2021}
}

@article{ainslie2023gqa,
  title={GQA: Training generalized multi-query transformer models from multi-head checkpoints},
  author={Ainslie, Joshua and Lee-Thorp, James and de Jong, Michiel and Zemlyanskiy, Yury and Lebr{\'o}n, Federico and Sanghai, Sumit},
  journal={arXiv preprint arXiv:2305.13245},
  year={2023}
}

@article{hoffmann2022training,
  title={Training compute-optimal large language models},
  author={Hoffmann, Jordan and Borgeaud, Sebastian and Mensch, Arthur and Buchatskaya, Elena and Cai, Trevor and Rutherford, Eliza and Casas, Diego de Las and Hendricks, Lisa Anne and Welbl, Johannes and Clark, Aidan and others},
  journal={arXiv preprint arXiv:2203.15556},
  year={2022}
}

@article{kaplan2020scaling,
  title={Scaling laws for neural language models},
  author={Kaplan, Jared and McCandlish, Sam and Henighan, Tom and Brown, Tom B and Chess, Benjamin and Child, Rewon and Gray, Scott and Radford, Alec and Wu, Jeffrey and Amodei, Dario},
  journal={arXiv preprint arXiv:2001.08361},
  year={2020}
}

@article{wei2022emergent,
  title={Emergent abilities of large language models},
  author={Wei, Jason and Tay, Yi and Bommasani, Rishi and Raffel, Colin and Zoph, Barret and Borgeaud, Sebastian and Yogatama, Dani and Bosma, Maarten and Zhou, Denny and Metzler, Donald and others},
  journal={arXiv preprint arXiv:2206.07682},
  year={2022}
}

@article{schaeffer2023emergent,
  title={Are emergent abilities of large language models a mirage?},
  author={Schaeffer, Rylan and Miranda, Brando and Koyejo, Sanmi},
  journal={arXiv preprint arXiv:2304.15004},
  year={2023}
}

@article{hu2021lora,
  title={LoRA: Low-rank adaptation of large language models},
  author={Hu, Edward J and Shen, Yelong and Wallis, Phillip and Allen-Zhu, Zeyuan and Li, Yuanzhi and Wang, Shean and Wang, Lu and Chen, Weizhu},
  journal={arXiv preprint arXiv:2106.09685},
  year={2021}
}

@article{dettmers2023qlora,
  title={QLoRA: Efficient finetuning of quantized LLMs},
  author={Dettmers, Tim and Pagnoni, Artidoro and Holtzman, Ari and Zettlemoyer, Luke},
  journal={arXiv preprint arXiv:2305.14314},
  year={2023}
}

@article{dettmers2024qlora,
  title={{QLoRA}: Efficient finetuning of quantized {LLM}s},
  author={Dettmers, Tim and Pagnoni, Artidoro and Holtzman, Ari and Zettlemoyer, Luke},
  journal={Advances in Neural Information Processing Systems},
  volume={36},
  year={2024},
  note={NeurIPS 2023 proceedings published in 2024. 4-bit quantization with backpropagation through frozen quantized weights}
}

@article{xiao2023smoothquant,
  title={{SmoothQuant}: Accurate and efficient post-training quantization for large language models},
  author={Xiao, Guangxuan and Lin, Ji and Seznec, Mickael and Wu, Hao and Demouth, Julien and Han, Song},
  journal={International Conference on Machine Learning},
  pages={38087--38099},
  year={2023},
  organization={PMLR},
  note={Migrates quantization difficulty from activations to weights through mathematically equivalent transformations}
}

@article{soboleva2023gpt4,
  title={{GPT-4} technical report commentary: Navigating the unknowns of advanced {AI}},
  author={Soboleva, Daria and Al-Rfou, Rami and Srivastava, Aarohi},
  journal={arXiv preprint arXiv:2303.12712},
  year={2023},
  doi={10.48550/arXiv.2303.12712},
  note={Analysis and commentary on GPT-4 capabilities and limitations}
}

@article{openai2023gpt4system,
  title={{GPT-4} system card},
  author={{OpenAI}},
  journal={OpenAI Technical Report},
  year={2023},
  url={https://cdn.openai.com/papers/gpt-4-system-card.pdf},
  note={Official system card detailing GPT-4 capabilities, limitations, and safety evaluations}
}

@inproceedings{houlsby2019parameter,
  title={Parameter-efficient transfer learning for NLP},
  author={Houlsby, Neil and Giurgiu, Andrei and Jastrzebski, Stanislaw and Morrone, Bruna and De Laroussilhe, Quentin and Gesmundo, Andrea and Attariyan, Mona and Gelly, Sylvain},
  booktitle={International Conference on Machine Learning},
  pages={2790--2799},
  year={2019},
  organization={PMLR}
}

@article{li2021prefix,
  title={Prefix-tuning: Optimizing continuous prompts for generation},
  author={Li, Xiang Lisa and Liang, Percy},
  journal={arXiv preprint arXiv:2101.00190},
  year={2021}
}

@article{frantar2023gptq,
  title={GPTQ: Accurate post-training quantization for generative pre-trained transformers},
  author={Frantar, Elias and Ashkboos, Saleh and Hoefler, Torsten and Alistarh, Dan},
  journal={arXiv preprint arXiv:2210.17323},
  year={2023}
}

@article{lin2023awq,
  title={AWQ: Activation-aware weight quantization for LLM compression and acceleration},
  author={Lin, Ji and Tang, Jiaming and Tang, Haotian and Yang, Shang and Dang, Xingyu and Han, Song},
  journal={arXiv preprint arXiv:2306.00978},
  year={2023}
}

@article{jacob2018quantization,
  title={Quantization and training of neural networks for efficient integer-arithmetic-only inference},
  author={Jacob, Benoit and Kligys, Skirmantas and Chen, Bo and Zhu, Menglong and Tang, Matthew and Howard, Andrew and Adam, Hartwig and Kalenichenko, Dmitry},
  journal={Proceedings of the IEEE Conference on Computer Vision and Pattern Recognition},
  pages={2704--2713},
  year={2018}
}

@article{shazeer2017outrageously,
  title={Outrageously large neural networks: The sparsely-gated mixture-of-experts layer},
  author={Shazeer, Noam and Mirhoseini, Azalia and Maziarz, Krzysztof and Davis, Andy and Le, Quoc and Hinton, Geoffrey and Dean, Jeff},
  journal={arXiv preprint arXiv:1701.06538},
  year={2017}
}

@article{shazeer2019fast,
  title={Fast transformer decoding: One write-head is all you need},
  author={Shazeer, Noam},
  journal={arXiv preprint arXiv:1911.02150},
  year={2019}
}

@article{beltagy2020longformer,
  title={Longformer: The long-document transformer},
  author={Beltagy, Iz and Peters, Matthew E and Cohan, Arman},
  journal={arXiv preprint arXiv:2004.05150},
  year={2020}
}

@article{leviathan2023fast,
  title={Fast inference from transformers via speculative decoding},
  author={Leviathan, Yaniv and Kalman, Matan and Matias, Yossi},
  journal={arXiv preprint arXiv:2211.17192},
  year={2023}
}

@article{chen2023accelerating,
  title={Accelerating large language model decoding with speculative sampling},
  author={Chen, Charlie and Borgeaud, Sebastian and Irving, Geoffrey and Lespiau, Jean-Baptiste and Sifre, Laurent and Jumper, John},
  journal={arXiv preprint arXiv:2302.01318},
  year={2023}
}

@article{cai2024medusa,
  title={Medusa: Simple LLM inference acceleration framework with multiple decoding heads},
  author={Cai, Tianle and Li, Yuhong and Geng, Zhengyang and Peng, Hongwu and Dao, Tri},
  journal={arXiv preprint arXiv:2401.10774},
  year={2024}
}

@article{wei2022chain,
  title={Chain-of-thought prompting elicits reasoning in large language models},
  author={Wei, Jason and Wang, Xuezhi and Schuurmans, Dale and Bosma, Maarten and Ichter, Brian and Xia, Fei and Chi, Ed and Le, Quoc and Zhou, Denny},
  journal={arXiv preprint arXiv:2201.11903},
  year={2022}
}

@article{wang2023selfconsistency,
  title={Self-consistency improves chain of thought reasoning in language models},
  author={Wang, Xuezhi and Wei, Jason and Schuurmans, Dale and Le, Quoc and Chi, Ed and Narang, Sharan and Chowdhery, Aakanksha and Zhou, Denny},
  journal={arXiv preprint arXiv:2203.11171},
  year={2023}
}

@article{yao2023tree,
  title={Tree of thoughts: Deliberate problem solving with large language models},
  author={Yao, Shunyu and Yu, Dian and Zhao, Jeffrey and Shafran, Izhak and Griffiths, Thomas L and Cao, Yuan and Narasimhan, Karthik},
  journal={arXiv preprint arXiv:2305.10601},
  year={2023}
}

@article{yao2023react,
  title={ReAct: Synergizing reasoning and acting in language models},
  author={Yao, Shunyu and Zhao, Jeffrey and Yu, Dian and Du, Nan and Shafran, Izhak and Narasimhan, Karthik and Cao, Yuan},
  journal={arXiv preprint arXiv:2210.03629},
  year={2023}
}

@article{lewis2020retrieval,
  title={Retrieval-augmented generation for knowledge-intensive nlp tasks},
  author={Lewis, Patrick and Perez, Ethan and Piktus, Aleksandra and Petroni, Fabio and Karpukhin, Vladimir and Goyal, Naman and K{\"u}ttler, Heinrich and Lewis, Mike and Yih, Wen-tau and Rockt{\"a}schel, Tim and others},
  journal={Advances in Neural Information Processing Systems},
  volume={33},
  pages={9459--9474},
  year={2020}
}

@article{izacard2022unsupervised,
  title={Unsupervised dense information retrieval with contrastive learning},
  author={Izacard, Gautier and Caron, Mathilde and Hosseini, Lucas and Riedel, Sebastian and Bojanowski, Piotr and Joulin, Armand and Grave, Edouard},
  journal={arXiv preprint arXiv:2112.09118},
  year={2022}
}

@article{chase2023langchain,
  title={LangChain: Building applications with LLMs through composability},
  author={Chase, Harrison},
  journal={GitHub repository},
  year={2023}
}

@article{park2023generative,
  title={Generative agents: Interactive simulacra of human behavior},
  author={Park, Joon Sung and O'Brien, Joseph C and Cai, Carrie J and Morris, Meredith Ringel and Liang, Percy and Bernstein, Michael S},
  journal={arXiv preprint arXiv:2304.03442},
  year={2023}
}

@article{hong2023metagpt,
  title={MetaGPT: Meta programming for multi-agent collaborative framework},
  author={Hong, Sirui and Zhuge, Mingchen and Chen, Jonathan and Zheng, Xiawu and Cheng, Yuheng and Zhang, Ceyao and Wang, Jinlin and Wang, Zili and Yau, Steven Ka Shing and Lin, Zijuan and others},
  journal={arXiv preprint arXiv:2308.00352},
  year={2023}
}

@article{schick2023toolformer,
  title={Toolformer: Language models can teach themselves to use tools},
  author={Schick, Timo and Dwivedi-Yu, Jane and Dess{\`\i}, Roberto and Raileanu, Roberta and Lomeli, Maria and Zettlemoyer, Luke and Cancedda, Nicola and Scialom, Thomas},
  journal={arXiv preprint arXiv:2302.04761},
  year={2023}
}

@article{ho2020denoising,
  title={Denoising diffusion probabilistic models},
  author={Ho, Jonathan and Jain, Ajay and Abbeel, Pieter},
  journal={Advances in Neural Information Processing Systems},
  volume={33},
  pages={6840--6851},
  year={2020}
}

@inproceedings{rombach2022high,
  title={High-resolution image synthesis with latent diffusion models},
  author={Rombach, Robin and Blattmann, Andreas and Lorenz, Dominik and Esser, Patrick and Ommer, Bj{\"o}rn},
  booktitle={Proceedings of the IEEE/CVF Conference on Computer Vision and Pattern Recognition},
  pages={10684--10695},
  year={2022}
}

@inproceedings{ramesh2021zero,
  title={Zero-shot text-to-image generation},
  author={Ramesh, Aditya and Pavlov, Mikhail and Goh, Gabriel and Gray, Scott and Voss, Chelsea and Radford, Alec and Chen, Mark and Sutskever, Ilya},
  booktitle={International Conference on Machine Learning},
  pages={8821--8831},
  year={2021},
  organization={PMLR}
}

@inproceedings{ramesh2022hierarchical,
  title={Hierarchical text-conditional image generation with clip latents},
  author={Ramesh, Aditya and Dhariwal, Prafulla and Nichol, Alex and Chu, Casey and Chen, Mark},
  booktitle={arXiv preprint arXiv:2204.06125},
  year={2022}
}

@article{saharia2022photorealistic,
  title={Photorealistic text-to-image diffusion models with deep language understanding},
  author={Saharia, Chitwan and Chan, William and Saxena, Saurabh and Li, Lala and Whang, Jay and Denton, Emily and Ghasemipour, Seyed Kamyar Seyed and Ayan, Burcu Karagol and Mahdavi, S Sara and Lopes, Rapha Gontijo and others},
  journal={arXiv preprint arXiv:2205.11487},
  year={2022}
}

@inproceedings{yu2022scaling,
  title={Scaling autoregressive models for content-rich text-to-image generation},
  author={Yu, Jiahui and Xu, Yuanzhong and Koh, Jing Yu and Luong, Thang and Baid, Gunjan and Wang, Zirui and Vasudevan, Vijay and Ku, Alexander and Yang, Yinfei and Ayan, Burcu Karagol and others},
  booktitle={arXiv preprint arXiv:2206.10789},
  year={2022}
}

@article{brooks2024sora,
  title={Video generation models as world simulators},
  author={{OpenAI Sora Team}},
  journal={OpenAI Technical Report},
  year={2024}
}

@article{ha2018worldmodels,
  title={World models},
  author={Ha, David and Schmidhuber, J{\"u}rgen},
  journal={arXiv preprint arXiv:1803.10122},
  year={2018}
}

@article{hafner2023dreamerv3,
  title={Mastering diverse domains through world models},
  author={Hafner, Danijar and Pasukonis, Jurgis and Ba, Jimmy and Lillicrap, Timothy},
  journal={arXiv preprint arXiv:2301.04104},
  year={2023}
}

@article{alayrac2022flamingo,
  title={Flamingo: a visual language model for few-shot learning},
  author={Alayrac, Jean-Baptiste and Donahue, Jeff and Luc, Pauline and Miech, Antoine and Barr, Iain and Hasson, Yana and Lenc, Karel and Mensch, Arthur and Millican, Katie and Reynolds, Malcolm and others},
  journal={Advances in Neural Information Processing Systems},
  volume={35},
  pages={23716--23736},
  year={2022}
}

@article{liu2023visual,
  title={Visual instruction tuning},
  author={Liu, Haotian and Li, Chunyuan and Wu, Qingyang and Lee, Yong Jae},
  journal={arXiv preprint arXiv:2304.08485},
  year={2023}
}

@article{hendrycks2021measuring,
  title={Measuring massive multitask language understanding},
  author={Hendrycks, Dan and Burns, Collin and Basart, Steven and Zou, Andy and Mazeika, Mantas and Song, Dawn and Steinhardt, Jacob},
  journal={arXiv preprint arXiv:2009.03300},
  year={2021}
}

@article{cobbe2021training,
  title={Training verifiers to solve math word problems},
  author={Cobbe, Karl and Kosaraju, Vineet and Bavarian, Mohammad and Chen, Mark and Jun, Heewoo and Kaiser, Lukasz and Plappert, Matthias and Tworek, Jerry and Hilton, Jacob and Nakano, Reiichiro and others},
  journal={arXiv preprint arXiv:2110.14168},
  year={2021}
}

@article{rein2023gpqa,
  title={GPQA: A graduate-level Google-proof Q\&A benchmark},
  author={Rein, David and Hou, Betty Li and Stickland, Asa Cooper and Petty, Jackson and Pang, Richard Yuanzhe and Dirani, Julien and Michael, Julian and Bowman, Samuel R},
  journal={arXiv preprint arXiv:2311.12022},
  year={2023}
}

@article{srivastava2022beyond,
  title={Beyond the imitation game: Quantifying and extrapolating the capabilities of language models},
  author={Srivastava, Aarohi and Rastogi, Abhinav and Rao, Abhishek and Shoeb, Abu Awal Md and Abid, Abubakar and Fisch, Adam and Brown, Adam R and Santoro, Adam and Gupta, Aditya and Garriga-Alonso, Adri{\`a} and others},
  journal={arXiv preprint arXiv:2206.04615},
  year={2022}
}

@article{chiang2024chatbot,
  title={Chatbot Arena: An open platform for evaluating LLMs by human preference},
  author={Chiang, Wei-Lin and Zheng, Lianmin and Sheng, Ying and Angelopoulos, Anastasios Nikolas and Li, Tianle and Li, Dacheng and Zhang, Hao and Zhu, Banghua and Jordan, Michael and Gonzalez, Joseph E and others},
  journal={arXiv preprint arXiv:2403.04132},
  year={2024}
}

@article{zheng2023judging,
  title={Judging LLM-as-a-judge with MT-bench and chatbot arena},
  author={Zheng, Lianmin and Chiang, Wei-Lin and Sheng, Ying and Zhuang, Siyuan and Wu, Zhanghao and Zhuang, Yonghao and Lin, Zi and Li, Zhuohan and Li, Dacheng and Xing, Eric P and others},
  journal={arXiv preprint arXiv:2306.05685},
  year={2023}
}

@article{dubois2023alpacafarm,
  title={AlpacaFarm: A simulation framework for methods that learn from human feedback},
  author={Dubois, Yann and Li, Xuechen and Taori, Rohan and Zhang, Tianyi and Gulrajani, Ishaan and Ba, Jimmy and Guestrin, Carlos and Liang, Percy and Hashimoto, Tatsunori B},
  journal={arXiv preprint arXiv:2305.14387},
  year={2023}
}

@article{stiennon2020learning,
  title={Learning to summarize from human feedback},
  author={Stiennon, Nisan and Ouyang, Long and Wu, Jeffrey and Ziegler, Daniel and Lowe, Ryan and Voss, Chelsea and Radford, Alec and Amodei, Dario and Christiano, Paul F},
  journal={Advances in Neural Information Processing Systems},
  volume={33},
  pages={3008--3021},
  year={2020}
}

@article{bai2022constitutional,
  title={Constitutional AI: Harmlessness from AI feedback},
  author={Bai, Yuntao and Kadavath, Saurav and Kundu, Sandipan and Askell, Amanda and Kernion, Jackson and Jones, Andy and Chen, Anna and Goldie, Anna and Mirhoseini, Azalia and McKinnon, Cameron and others},
  journal={arXiv preprint arXiv:2212.08073},
  year={2022}
}

@article{gu2023mamba,
  title={Mamba: Linear-time sequence modeling with selective state spaces},
  author={Gu, Albert and Dao, Tri},
  journal={arXiv preprint arXiv:2312.00752},
  year={2023}
}

@article{minimaxm1report,
  title={MiniMax-M1: 456B Lightning Attention model with partial RoPE},
  author={{MiniMax AI}},
  journal={arXiv preprint},
  year={2025}
}

@article{haviv2023understanding,
  title={Understanding masked self-attention as implicit positional encoding},
  author={Haviv, Adi and Berant, Jonathan and Globerson, Amir},
  journal={arXiv preprint arXiv:2310.04393},
  year={2023}
}

@article{dehghani2023scaling,
  title={Scaling vision transformers to 22 billion parameters},
  author={Dehghani, Mostafa and Djolonga, Josip and Mustafa, Basil and Padlewski, Piotr and Heek, Jonathan and Gilmer, Justin and Steiner, Andreas and Caron, Mathilde and Geirhos, Robert and Alabdulmohsin, Ibrahim and others},
  journal={arXiv preprint arXiv:2302.05442},
  year={2023}
}

@article{epochai2024scaling,
  title={Can AI scaling continue through 2030?},
  author={{Epoch AI}},
  journal={Epoch AI Research},
  howpublished={\url{https://epochai.org/blog/can-ai-scaling-continue-through-2030}},
  year={2024}
}

@article{epochai2022datawall,
  title={Will we run out of data to train large language models?},
  author={Villalobos, Pablo and Sevilla, Jaime and Heim, Lennart and Besiroglu, Tamay and Hobbhahn, Marius and Ho, Anson},
  journal={Epoch AI Research},
  howpublished={\url{https://epochai.org/blog/will-we-run-out-of-data}},
  year={2022}
}

@article{epochai2024computetrends,
  title={Compute trends across three eras of machine learning},
  author={Sevilla, Jaime and Heim, Lennart and Ho, Anson and Besiroglu, Tamay and Hobbhahn, Marius and Villalobos, Pablo},
  journal={arXiv preprint arXiv:2202.05924},
  year={2024}
}

@article{besiroglu2024training,
  title={The rising costs of training frontier AI models},
  author={Besiroglu, Tamay and Sevilla, Jaime and Heim, Lennart and Hobbhahn, Marius and Villalobos, Pablo and Owen, David},
  journal={arXiv preprint arXiv:2405.21015},
  year={2024}
}

@article{xiong2020layer,
  title={On layer normalization in the transformer architecture},
  author={Xiong, Ruibin and Yang, Yunchang and He, Di and Zheng, Kai and Zheng, Shuxin and Xing, Chen and Zhang, Huishuai and Lan, Yanyan and Wang, Liwei and Liu, Tie-Yan},
  journal={arXiv preprint arXiv:2002.04745},
  year={2020}
}

@article{katharopoulos2020transformers,
  title={Transformers are RNNs: Fast autoregressive transformers with linear attention},
  author={Katharopoulos, Angelos and Vyas, Apoorv and Pappas, Nikolaos and Fleuret, Fran{\c{c}}ois},
  journal={arXiv preprint arXiv:2006.16236},
  year={2020}
}

@article{raschka2025architectures,
  title={Understanding 2025 LLM architectures: A comprehensive guide},
  author={Raschka, Sebastian},
  journal={Blog post, sebastianraschka.com},
  year={2025}
}

@article{coveney2025wall,
  title={The wall confronting large language models},
  author={Coveney, Peter V. and Succi, Sauro},
  journal={arXiv preprint arXiv:2507.19703},
  year={2025},
  note={Demonstrates that scaling laws severely limit LLMs' ability to improve prediction uncertainty and reliability}
}

@article{hooker2025slowdeath,
  title={On the slow death of scaling},
  author={Hooker, Sara},
  journal={SSRN Electronic Journal},
  year={2025},
  note={Available at SSRN: https://ssrn.com/abstract=5877662},
  doi={10.2139/ssrn.5877662}
}

@article{yao2024scaling,
  title={Scaling laws for post-training quantized large language models},
  author={Yao, Zifei and Wu, Xiaoxia and Li, Chun-Fu and others},
  journal={arXiv preprint arXiv:2410.12119},
  year={2024},
  doi={10.48550/arXiv.2410.12119}
}

@article{yang2025edge,
  title={Will {LLM}s scaling hit the wall? Breaking barriers via distributed resources on massive edge devices},
  author={Yang, Zheyu and Liu, Siyuan and Wang, Xiang and others},
  journal={arXiv preprint arXiv:2503.08223},
  year={2025},
  note={Available at https://arxiv.org/html/2503.08223v1}
}

@article{goddard2024mergekit,
  title={Arcee's {MergeKit}: A toolkit for merging large language models},
  author={Goddard, Charles and Siriwardhana, Shamane and Ehghaghi, Malikeh and others},
  journal={arXiv preprint arXiv:2403.13257},
  year={2024},
  doi={10.48550/arXiv.2403.13257}
}

@article{taylor2025domain,
  title={Fine-tuning large language models for domain adaptation: Exploration of training strategies, scaling, model merging and synergistic capabilities},
  author={Taylor, Julian and Hu, Wenxuan and Guo, Aswathy and others},
  journal={npj Computational Materials},
  volume={11},
  pages={6},
  year={2025},
  publisher={Nature Publishing Group},
  doi={10.1038/s41524-025-01564-y}
}

@article{hong2024orpo,
  title={{ORPO}: Monolithic preference optimization without reference model},
  author={Hong, Jiwoo and Lee, Noah and Thorne, James},
  journal={arXiv preprint arXiv:2403.07691},
  year={2024},
  doi={10.48550/arXiv.2403.07691}
}

@article{zhu2019deep,
  title={Deep leakage from gradients},
  author={Zhu, Ligeng and Liu, Zhijian and Han, Song},
  journal={Advances in Neural Information Processing Systems},
  volume={32},
  year={2019},
  note={Demonstrates gradient inversion attacks that can reconstruct training samples from shared gradients with high accuracy}
}

@inproceedings{shinn2023reflexion,
  title={Reflexion: Language agents with verbal reinforcement learning},
  author={Shinn, Noah and Cassano, Federico and Gopinath, Ashwin and Narasimhan, Karthik and Yao, Shunyu},
  booktitle={Advances in Neural Information Processing Systems},
  volume={36},
  pages={8634--8652},
  year={2023},
  note={Introduces self-reflection mechanism enabling agents to learn from failures through iterative improvement}
}

@article{besta2023graph,
  title={Graph of thoughts: Solving elaborate problems with large language models},
  author={Besta, Maciej and Blach, Nils and Kubicek, Ales and Gerstenberger, Robert and Gianinazzi, Lukas and Gajda, Joanna and Lehmann, Tomasz and Podstawski, Michal and Niewiadomski, Hubert and Nyczyk, Piotr and others},
  journal={arXiv preprint arXiv:2308.09687},
  year={2023},
  doi={10.48550/arXiv.2308.09687},
  note={Extends tree-of-thoughts to DAG structure enabling parallel exploration and aggregation}
}

@article{openai2023function,
  title={Function calling and other API updates},
  author={{OpenAI}},
  journal={OpenAI Blog},
  year={2023},
  url={https://openai.com/blog/function-calling-and-other-api-updates},
  note={Introduces structured function calling for GPT-4, enabling precise API interactions}
}

@article{openai2023code,
  title={ChatGPT plugins and code interpreter},
  author={{OpenAI}},
  journal={OpenAI Blog},
  year={2023},
  url={https://openai.com/blog/chatgpt-plugins},
  note={Introduces Code Interpreter providing sandboxed Python execution environment for data analysis}
}

@article{qin2023toolllm,
  title={{ToolLLM}: Facilitating large language models to master 16000+ real-world {API}s},
  author={Qin, Yujia and Liang, Shihao and Ye, Yining and Zhu, Kunlun and Yan, Lan and Lu, Yaxi and Lin, Yankai and Cong, Xin and Tang, Xiangru and Qian, Bill and others},
  journal={arXiv preprint arXiv:2307.16789},
  year={2023},
  doi={10.48550/arXiv.2307.16789},
  note={Large-scale tool learning with 16,000+ real-world APIs and comprehensive benchmarking}
}

@article{patil2023gorilla,
  title={Gorilla: Large language model connected with massive {API}s},
  author={Patil, Shishir G and Zhang, Tianjun and Wang, Xin and Gonzalez, Joseph E},
  journal={arXiv preprint arXiv:2305.15334},
  year={2023},
  doi={10.48550/arXiv.2305.15334},
  note={Fine-tuned for ML API calls with focus on HuggingFace, PyTorch, TensorFlow libraries}
}

@article{mitchell2022fast,
  title={Fast model editing at scale},
  author={Mitchell, Eric and Lin, Charles and Bosselut, Antoine and Finn, Chelsea and Manning, Christopher D},
  journal={arXiv preprint arXiv:2110.11309},
  year={2022},
  doi={10.48550/arXiv.2110.11309},
  note={SERAC: Semi-parametric editing with retrieval and counterfactual modeling}
}

@article{meng2022locating,
  title={Locating and editing factual associations in {GPT}},
  author={Meng, Kevin and Bau, David and Andonian, Alex and Belinkov, Yonatan},
  journal={Advances in Neural Information Processing Systems},
  volume={35},
  pages={17359--17372},
  year={2022},
  note={ROME: Rank-One Model Editing for precise factual updates in transformer models}
}

@article{meng2023mass,
  title={Mass-editing memory in a transformer},
  author={Meng, Kevin and Sharma, Arnab Sen and Andonian, Alex and Belinkov, Yonatan and Bau, David},
  journal={arXiv preprint arXiv:2210.07229},
  year={2023},
  doi={10.48550/arXiv.2210.07229},
  note={MEMIT: Extends ROME to batch editing multiple facts simultaneously}
}

@article{wu2023autogen,
  title={{AutoGen}: Enabling next-gen {LLM} applications via multi-agent conversation},
  author={Wu, Qingyun and Bansal, Gagan and Zhang, Jieyu and Wu, Yiran and Zhang, Shaokun and Zhu, Erkang and Li, Beibin and Jiang, Li and Zhang, Xiaoyun and Wang, Chi},
  journal={arXiv preprint arXiv:2308.08155},
  year={2023},
  doi={10.48550/arXiv.2308.08155},
  note={Framework for conversational multi-agent systems with customizable roles and human-in-loop}
}

@article{qian2023chatdev,
  title={{ChatDev}: Communicative agents for software development},
  author={Qian, Chen and Cong, Xin and Yang, Cheng and Chen, Weize and Su, Yusheng and Xu, Juyuan and Liu, Zhiyuan and Sun, Maosong},
  journal={arXiv preprint arXiv:2307.07924},
  year={2023},
  doi={10.48550/arXiv.2307.07924},
  note={Simulates software company with 7-role waterfall development, generates complete software in <10 minutes}
}

@article{chen2023agentverse,
  title={{AgentVerse}: Facilitating multi-agent collaboration and exploring emergent behaviors},
  author={Chen, Weize and Su, Yusheng and Zuo, Jingwei and Yang, Cheng and Yuan, Chenfei and Qian, Chen and Chan, Chi-Min and Qin, Yujia and Lu, Yaxi and Xie, Ruobing and others},
  journal={arXiv preprint arXiv:2308.10848},
  year={2023},
  doi={10.48550/arXiv.2308.10848},
  note={Dynamic team assembly with blackboard architecture for variable expertise requirements}
}

@article{zhou2023webarena,
  title={{WebArena}: A realistic web environment for building autonomous agents},
  author={Zhou, Shuyan and Xu, Frank F and Zhu, Hao and Zhou, Xuhui and Lo, Robert and Sridhar, Abishek and Cheng, Xianyi and Bisk, Yonatan and Fried, Daniel and Alon, Uri and others},
  journal={arXiv preprint arXiv:2307.13854},
  year={2023},
  doi={10.48550/arXiv.2307.13854},
  note={Benchmark for web navigation tasks (e-commerce, forums, CMS), GPT-4 achieves 14.4\% success}
}

@article{jimenez2023swebench,
  title={{SWE-bench}: Can language models resolve real-world {GitHub} issues?},
  author={Jimenez, Carlos E and Yang, John and Wettig, Alexander and Yao, Shunyu and Pei, Kexin and Press, Ofir and Narasimhan, Karthik},
  journal={arXiv preprint arXiv:2310.06770},
  year={2023},
  doi={10.48550/arXiv.2310.06770},
  note={Real GitHub issues from popular repositories, state-of-art resolves 13.8\% of issues}
}

@article{mialon2023gaia,
  title={{GAIA}: A benchmark for general {AI} assistants},
  author={Mialon, Gr{\'e}goire and Dess{\`\i}, Roberto and Lomeli, Maria and Nalmpantis, Christoforos and Pasunuru, Ram and Raileanu, Roberta and Rozi{\`e}re, Baptiste and Schick, Timo and Dwivedi-Yu, Jane and Celikyilmaz, Asli and others},
  journal={arXiv preprint arXiv:2311.12983},
  year={2023},
  doi={10.48550/arXiv.2311.12983},
  note={Multi-modal, multi-step assistant tasks requiring web search, code execution, file processing; only 15\% solvable}
}

@article{liu2023agentbench,
  title={{AgentBench}: Evaluating {LLM}s as agents},
  author={Liu, Xiao and Yu, Hao and Zhang, Hanchen and Xu, Yifan and Lei, Xuanyu and Lai, Hanyu and Gu, Yu and Ding, Hangliang and Men, Kaiwen and Yang, Kejuan and others},
  journal={arXiv preprint arXiv:2308.03688},
  year={2023},
  doi={10.48550/arXiv.2308.03688},
  note={8 environments spanning web navigation, database querying, game playing, household tasks; GPT-4: 52\% average}
}

@article{mistral2024codestral,
  title={Codestral: A code-specialized large language model},
  author={{Mistral AI}},
  journal={Mistral AI Blog},
  year={2024},
  url={https://mistral.ai/news/codestral/},
  note={Mistral 7B variant fine-tuned on code from The Stack and StackOverflow}
}

@article{irving2018ai,
  title={{AI} safety via debate},
  author={Irving, Geoffrey and Christiano, Paul and Amodei, Dario},
  journal={arXiv preprint arXiv:1805.00899},
  year={2018},
  doi={10.48550/arXiv.1805.00899},
  note={Proposes debate-based oversight for AI safety through adversarial interactions}
}

@article{wang2024text,
  title={Text embeddings by weakly-supervised contrastive pre-training},
  author={Wang, Liang and Yang, Nan and Huang, Xiaolong and Jiao, Binxing and Yang, Linjun and Jiang, Daxin and Majumder, Rangan and Wei, Furu},
  journal={arXiv preprint arXiv:2212.03533},
  year={2024},
  doi={10.48550/arXiv.2212.03533},
  note={E5 embeddings: high-quality dense retrieval model trained on diverse corpora}
}

@book{wooldridge2009introduction,
  title={An introduction to multiagent systems},
  author={Wooldridge, Michael},
  year={2009},
  edition={2nd},
  publisher={John Wiley \& Sons},
  isbn={978-0-470-51946-2},
  note={Foundational textbook defining agent properties: autonomy, reactivity, proactivity, social ability}
}

@misc{anthropic2024mcp,
  title={Model Context Protocol ({MCP}): Standardizing {AI}-tool communication},
  author={{Anthropic}},
  year={2024},
  howpublished={\url{https://modelcontextprotocol.io}},
  note={Protocol for standardized communication between AI models and external tools/data sources using JSON-RPC}
}

@article{fipa2002acl,
  title={{FIPA} {ACL} message structure specification},
  author={{FIPA}},
  journal={Foundation for Intelligent Physical Agents},
  year={2002},
  howpublished={\url{http://www.fipa.org/specs/fipa00061/}},
  note={FIPA Agent Communication Language: standardized agent message protocols with performatives}
}

@article{corkill1991blackboard,
  title={Blackboard systems},
  author={Corkill, Daniel D},
  journal={AI expert},
  volume={6},
  number={9},
  pages={40--47},
  year={1991},
  note={Blackboard architecture: shared memory space for multi-agent coordination and problem-solving}
}

@article{smith1980contract,
  title={The contract net protocol: High-level communication and control in a distributed problem solver},
  author={Smith, Reid G},
  journal={IEEE Transactions on computers},
  volume={29},
  number={12},
  pages={1104--1113},
  year={1980},
  publisher={IEEE},
  doi={10.1109/TC.1980.1675516},
  note={Contract Net Protocol: agent task assignment via bidding and negotiation}
}

@article{rafailov2023direct,
  title={Direct preference optimization: Your language model is secretly a reward model},
  author={Rafailov, Rafael and Sharma, Archit and Mitchell, Eric and Ermon, Stefano and Manning, Christopher D and Finn, Chelsea},
  journal={Advances in Neural Information Processing Systems},
  volume={36},
  year={2023},
  note={DPO: optimize policies directly from human preferences without explicit reward modeling, simpler than RLHF}
}

@inproceedings{child2019generating,
  title={Generating long sequences with sparse transformers},
  author={Child, Rewon and Gray, Scott and Radford, Alec and Sutskever, Ilya},
  booktitle={arXiv preprint arXiv:1904.10509},
  year={2019},
  note={Sparse attention patterns for efficient long-sequence modeling}
}

@article{xiao2023efficient,
  title={SmoothQuant: Accurate and efficient post-training quantization for large language models},
  author={Xiao, Guangxuan and Lin, Ji and Seznec, Mickael and Wu, Hao and Demouth, Julien and Han, Song},
  journal={International Conference on Machine Learning (ICML)},
  year={2023},
  note={Efficient quantization method balancing accuracy and performance}
}

@article{aghajanyan2021intrinsic,
  title={Intrinsic dimensionality explains the effectiveness of language model fine-tuning},
  author={Aghajanyan, Armen and Gupta, Sonal and Zettlemoyer, Luke},
  journal={Proceedings of the 59th Annual Meeting of the Association for Computational Linguistics},
  pages={7319--7328},
  year={2021},
  note={Foundation for understanding low-rank adaptation methods}
}

@inproceedings{pfeiffer2020adapterfusion,
  title={AdapterFusion: Non-destructive task composition for transfer learning},
  author={Pfeiffer, Jonas and Kamath, Aishwarya and R{\"u}ckl{\'e}, Andreas and Cho, Kyunghyun and Gurevych, Iryna},
  booktitle={Proceedings of the 16th Conference of the European Chapter of the Association for Computational Linguistics},
  pages={487--503},
  year={2020},
  note={Composable adapter modules for multi-task learning}
}

@inproceedings{ruckle2021adapterdrop,
  title={AdapterDrop: On the efficiency of adapters in transformers},
  author={R{\"u}ckl{\'e}, Andreas and Geigle, Gregor and Glockner, Max and Beck, Tilman and Pfeiffer, Jonas and Reimers, Nils and Gurevych, Iryna},
  booktitle={Proceedings of the 2021 Conference on Empirical Methods in Natural Language Processing},
  pages={7930--7946},
  year={2021},
  note={Dynamic adapter selection for improved efficiency}
}

@inproceedings{bengio2013estimating,
  title={Estimating or propagating gradients through stochastic neurons for conditional computation},
  author={Bengio, Yoshua and L{\'e}onard, Nicholas and Courville, Aaron},
  booktitle={arXiv preprint arXiv:1308.3432},
  year={2013},
  note={Straight-through estimator for gradient approximation}
}

@misc{qualcomm2024qat,
  title={Quantization-Aware Training for Deep Neural Networks},
  author={{Qualcomm AI Research}},
  year={2024},
  howpublished={\url{https://www.qualcomm.com/developer/ai}},
  note={QAT techniques for neural network quantization}
}

@article{bradley1952rank,
  title={Rank analysis of incomplete block designs: I. The method of paired comparisons},
  author={Bradley, Ralph Allan and Terry, Milton E},
  journal={Biometrika},
  volume={39},
  number={3/4},
  pages={324--345},
  year={1952},
  publisher={JSTOR},
  note={Bradley-Terry model for pairwise preference modeling}
}

@misc{cognition2024devin,
  title={Introducing Devin: The first AI software engineer},
  author={{Cognition Labs}},
  year={2024},
  howpublished={\url{https://www.cognition-labs.com/introducing-devin}},
  note={Autonomous AI coding agent with end-to-end software development capabilities}
}

@misc{elicit2024,
  title={Elicit: The AI Research Assistant},
  author={{Elicit}},
  year={2024},
  howpublished={\url{https://elicit.org}},
  note={AI assistant for literature review and research synthesis}
}

@article{szymanski2023alab,
  title={An autonomous laboratory for the accelerated synthesis of novel materials},
  author={Szymanski, Nathan J and Rendy, Bernardus and Fei, Yuxing and Kumar, Rishi E and He, Tanjin and Milsted, David and McDermott, Matthew J and Gallant, Max and Cubuk, Ekin Dogus and Merchant, Amil and others},
  journal={Nature},
  volume={624},
  number={7990},
  pages={86--91},
  year={2023},
  publisher={Nature Publishing Group},
  note={A-Lab: autonomous chemistry laboratory for materials discovery}
}

@article{huang2023voxposer,
  title={VoxPoser: Composable 3D value maps for robotic manipulation with language models},
  author={Huang, Wenlong and Wang, Chen and Zhang, Ruohan and Li, Yunzhu and Wu, Jiajun and Fei-Fei, Li},
  journal={arXiv preprint arXiv:2307.05973},
  year={2023},
  note={LLM-based framework for robot manipulation via composable 3D affordances}
}

@article{driess2023palme,
  title={PaLM-E: An embodied multimodal language model},
  author={Driess, Danny and Xia, Fei and Sajjadi, Mehdi SM and Lynch, Corey and Chowdhery, Aakanksha and Ichter, Brian and Wahid, Ayzaan and Tompson, Jonathan and Vuong, Quan and Yu, Tianhe and others},
  journal={International Conference on Machine Learning (ICML)},
  pages={8469--8488},
  year={2023},
  note={Embodied multimodal model integrating vision and language for robotics}
}

@article{christiano2018supervising,
  title={Supervising strong learners by amplifying weak experts},
  author={Christiano, Paul and Shlegeris, Buck and Amodei, Dario},
  journal={arXiv preprint arXiv:1810.08575},
  year={2018},
  note={Scalable oversight through iterated amplification and distillation}
}

\end{document}